\documentclass[preprint,review]{elsarticle}

\usepackage{geometry}
\usepackage{graphicx}
\usepackage{savesym}
\usepackage{amsmath}
\savesymbol{iint}
\usepackage{txfonts}
\restoresymbol{TXF}{iint}
\usepackage{url}
\usepackage{multirow} 

\usepackage{bm} 
\usepackage{upgreek} 
\usepackage{tikz} 
\tikzstyle{arrow} = [thick,->,>=stealth]
\usepackage{enumitem} 
\usepackage[ruled]{algorithm2e}

\usepackage{algorithmicx}
\usepackage{algpseudocode}

\usepackage{array}
\usepackage{subcaption}

\usepackage{booktabs} 
\DeclareMathOperator*{\argmax}{arg\,max}
\DeclareMathOperator{\tr}{tr}
\DeclareMathOperator{\diag}{diag}
\usepackage{makecell}
\usepackage{threeparttable}
\usepackage[labelsep=period,figurename=Fig.]{caption}

\journal{Pattern Recognition}
\begin{document}
\begin{frontmatter}

\title{Orthogonal Least Squares Based Fast Feature Selection for Linear Classification}

\author[mymainaddress1]{Sikai Zhang}
\ead{matthew.szhang91@gmail.com}
\address[mymainaddress1]{Department of Mechanical Engineering, The University of Sheffield, Sheffield, United Kingdom}
\author[mymainaddress2]{Zi-Qiang Lang\corref{mycorrespondingauthor}}
\cortext[mycorrespondingauthor]{Corresponding author}
\ead{z.lang@sheffield.ac.uk}
\address[mymainaddress2]{Department of Automatic Control and Systems Engineering, The University of Sheffield, Sheffield, United Kingdom}

\begin{abstract}
An Orthogonal Least Squares (OLS) based feature selection method is proposed for both binomial and multinomial classification.
The novel Squared Orthogonal Correlation Coefficient (SOCC) is defined based on Error Reduction Ratio (ERR) in OLS and used as the feature ranking criterion.
The equivalence between the canonical correlation coefficient, Fisher's criterion, and the sum of the SOCCs is revealed, which unveils the statistical implication of ERR in OLS for the first time.
It is also shown that the OLS based feature selection method has speed advantages when applied for greedy search.
The proposed method is comprehensively compared with the mutual information based feature selection methods and the embedded methods using both synthetic and real world datasets.
The results show that the proposed method is always in the top 5 among the 12 candidate methods.
Besides, the proposed method can be directly applied to continuous features without discretisation, which is another significant advantage over mutual information based methods.
\end{abstract}

\end{frontmatter}

\section{Introduction}
The aim of the feature selection for classification is to select an optimal subset of features given the candidate features, which are numerical or categorical, and the response, which is categorical.
The feature selection methods can be divided into three types: filter, wrapper, and embedded methods \cite{guyon2003introduction,abualigah2017text,abualigah2018new}.
The filter methods rank the individual candidate features based on certain statistics, such as the correlation coefficient and mutual information \cite{peng2005feature,abualigah2019feature,abualigah2020multi}.
The wrapper methods train classifiers by ranking the subsets of candidate features based on their classification performance.
The embedded methods, e.g. LASSO \cite{tibshirani1996regression} and CART \cite{breiman1984classification}, select optimal features during the training process of a specific classifier.

Comparing with the other two methods, a filter method is not based on a specific type of classifiers, so a filter method is more suitable to be used in the early stage where the type of classifiers has not been decided.
To rank the features by a filter method, it is desired that the features in the subset have the high relevance to the response, while the low redundancy between themselves.
A straightforward way is to optimise the objective function constructed by the difference or the quotient between the relevance and the redundancy.
For example, the well-known minimal-Redundancy-Maximal-Relevance (mRMR) method adopts this idea, in which the relevance and redundancy are quantified by the mutual information \cite{ding2005minimum}.
The second idea is to control the redundancy by orthogonalising the candidate features, and to find the maximum relevance between the orthogonalised features and the response.
The second idea has been used in the term selection of time series models by Orthogonal Least Squares (OLS), where the relevance is defined by the Error Reduction Ratio (ERR) \cite{chen1989orthogonal}.
These two ideas basically evaluate the relevance between the single feature and the response, and the relevance is analysed separately with the redundancy.
The third idea uses the overall relevance between the subset features and the response.
The definition of the overall relevance has taken the redundancy into consideration, e.g. the multiple correlation coefficient and the canonical correlation coefficient \cite{cooley1971multivariate}.
The first idea is extensively used in mutual information based filter methods \cite{brown2012conditional}.
However, the idea does not take the feature interaction into consideration, which makes the filter methods have a well-known drawback compared to the wrapper and embedded methods \cite{li2017recent}.
The interaction\footnote{The feature interaction can also be understood in the way of the conditional redundancy \cite{brown2012conditional}.} between features exists when a feature has to be combined with one or more other features to represent the response \cite{zeng2015novel}.
Without considering the feature interaction, the feature selection methods will fail to select the features having low relevance individually but high relevance together.
In this paper, the proposed feature selection method is based on the third idea with the feature interaction issue addressed using an approach similar to the wrapper methods in order to simultaneously handle the feature relevance, feature redundancy, and feature interaction.
This can, in conjunction with the second idea, achieve a faster computation speed.
In addition, compared to the mutual information based methods, which can only work with the discrete or categorical features \cite{li2017feature,sharmin2019simultaneous}, the proposed method is applicable to both numerical (including discrete and continuous) and categorical features.
The proposed method is also closely related to two recent advanced topics in the feature selection field, which are the multi-label feature selection \cite{kashef2019label,zhang2019distinguishing,hu2020multi} and multivariate time series feature selection \cite{dau2019ucr,ircio2020mutual}, respectively.
The fundamental issue that is addressed under the two research topics is to develop methods that can deal with the features and response that have to be represented in a matrix form.
The proposed method can naturally deal with the feature and response represented by a matrix without a need to introduce any additional techniques.

Basically, the contributions of the present study are in two aspects. 
First, the study reveals, for the first time, the relationships between the OLS and some well-known statistics including multiple correlation coefficient, canonical correlation coefficient, and Fisher’s criterion. 
Second, via utilising these relationships, a novel feature selection method for classification is developed. 
The novel method can deal with both numerical (including continuous and discrete) and categorical features, has a much faster computation speed when used with a greedy search, and can simultaneously address issues associated with feature relevance, feature redundancy, and feature interaction.
The computational efficiency and general applicability show the proposed method has potential to be widely applied to address feature selection issues in classification problems.

The rest of the paper is organised as follows.
The related work about OLS is introduced in {Section 2}.
In {Section 3}, based on OLS, the definition of the SOCCs is given.
The relationships of the SOCCs with multiple correlation coefficient and canonical correlation coefficient are analysed.
Then, via these relationships, an OLS based feature selection method is developed for binomial classification ({Section 4}) and multinomial classification ({Section 5}), respectively.
After that, the speed advantage of the method in the greedy search is analysed for both binomial and multinomial classification problems.
The relationship of the SOCC with Linear Discriminant Analysis (LDA) is also studied in {Section 5} to demonstrate the statistical implication of the proposed SOCC in classification.
In {Section 6}, a detailed example is provided to illustrate the procedure of the proposed method, and its relationship with Canonical Correlation Analysis (CCA) and linear discriminant analysis.
Moreover, a comprehensive comparison of the proposed method with the mutual information based methods and the embedded methods is carried out on both synthetic and real world datasets.
Finally, conclusions are summarised in {Section 7}.

\section{Related work} \label{sec:rw}
An important basis of the new method proposed in the present study is the OLS.
The OLS and associated representative works are briefly summarised in Table \ref{tbl:review}.
OLS was firstly developed by Korenberg for the fast parameter estimation \cite{korenberg1988identifying} and term selection \cite{korenberg1989robust} of the polynomial Nonlinear AutoRegressive with eXogenous input (NARX) model.
The criterion used for term selection is the Mean-Square Error Reduction (MSER).
Given a linear regression model
\begin{equation}
    \mathbf{y} = \left({\mathbf{1}, \mathbf{X}}\right)\begin{pmatrix}
    \beta_0 \\
    \bm{\upbeta}
    \end{pmatrix} + \mathbf{e}\text{,}
\end{equation}
where the response vector is
\begin{equation}
    \mathbf{y} = \left({y_1,\ldots,y_N}\right)^\top\text{,}
\end{equation}
the design matrix of $n$ independent variables with a constant term is
\begin{equation}
    \left({\mathbf{1}, \mathbf{X}}\right) = \left({\mathbf{1}, \mathbf{x}_1, \ldots, \mathbf{x}_n}\right) = \begin{pmatrix}
         1 & x_{1,1} & \hdots & x_{1,n}\\
         \vdots & \vdots & \ddots & \vdots \\
         1 & x_{N,1} & \hdots & x_{N,n}
    \end{pmatrix}\text{,}
\end{equation}
the unknown parameter vector is
\begin{equation}
    \begin{pmatrix}
    \beta_0 \\
    \bm{\upbeta}
    \end{pmatrix} = \left({\beta_0,\beta_1,\ldots,\beta_n}\right)^\top\text{,}
\end{equation}
the error term is
\begin{equation}
    \mathbf{e} = \left({e_1,\ldots,e_N}\right)^\top\text{,}
\end{equation}
the MSER is defined as
\begin{equation} \label{eq:mser}
    \text{MSER}_i = \hat{g}_i^{*2}{\mathbf{w}}_i^{*\top}{{\mathbf{w}}_i^*}\quad\quad i = 0,\ldots,n\text{.}
\end{equation}
In \eqref{eq:mser}, $\mathbf{w}_i$ is the $i^\text{th}$ column of matrix $\mathbf{W}^* = ({{\mathbf{w}}_0^*}, {{\mathbf{w}}_1^*}, \ldots ,{{\mathbf{w}}_n^*})$ determined from the unnormalised reduced QR decomposition of the design matrix $({\mathbf{1}, \mathbf{X}})$, such that
\begin{equation}
    \left({\mathbf{1}, \mathbf{X}}\right) = \mathbf{W}^*\mathbf{A}^*
\end{equation}
where $\mathbf{A}^*$ is the upper triangle matrix from the QR decomposition.
In \eqref{eq:mser}, $\hat{g}_i^{*}$ is obtained by solving the normal equation
\begin{equation}
    \mathbf{W}^{*\top}\mathbf{W}^*\begin{pmatrix}
    \hat{g}_0^{*}\\
    \vdots\\
    \hat{g}_n^{*}
    \end{pmatrix} = \mathbf{W}^{*\top}\mathbf{y}\text{.}
\end{equation}
Chen et al. \cite{chen1989orthogonal} extended OLS to the polynomial Nonlinear AutoRegressive Moving Average with eXogenous input (NARMAX) model, where the term ranking criterion is changed to the ERR, which is the MSER normalised by the inner product of the measured response, i.e.
\begin{equation} \label{eq:err}
    \text{ERR}_i = \frac{{{\hat{g}_i^{*2}{\mathbf{w}}_i^{*\top}{{\mathbf{w}}_i^*}} }}{{{{\mathbf{y}}^{\top}}{\mathbf{y}}}}\quad\quad i = 0,\ldots,n\text{.}
\end{equation}
Later, OLS became well-known in the nonlinear system identification field and the researchers explored the application of OLS in machine learning.
For example, Chen et al. \cite{chen1991orthogonal} applied the ERR to choose the centres of Radial Basis Functions (RBFs) for training RBF neural networks.
Wei and Billings \cite{wei2006feature} extended the ERR to the unsupervised feature selection, where the ERR is applied to evaluate the explanation capability of the selected features to all candidate features.
Recently, Solares et al. \cite{solares2019novel} applied the OLS method to the selection of the terms in the logistic-NARX model, which maps the continuous response of the NARX model to the binary response by a logistic function.
The point-Biserial Correlation Coefficient between the Orthogonalised term and measured response was adopted as the term ranking criterion (OBCC).
However, none of these previous studies has revealed any relationship between the OLS and well-known statistics such as canonical correlation coefficient and Fisher’s criterion.  
Probably because of this, the OLS has never be used as a general feature selection method to solve binomial or multinomial classification problems.

In the present study, a novel feature ranking criterion referred to as Squared Orthogonal Correlation Coefficient (SOCC) is defined using ERR associated with the OLS approach to a standard linear regression problem. 
For the first time, the SOCC reveals the statistic implication of ERR in OLS and unveils a significant relationship between the ERR and classic statistics such as canonical correlation coefficient and Fisher’s criterion. 
This, consequently, enables the development of an effective Canonical Correlation Analysis (CCA) based fast feature selection approach for both binomial and multinomial classifications.

\begin{table}[ht]
\caption{A summary of OLS and associated representative works.}
  \centering
  \scalebox{0.9}{
    \begin{tabular}{ l l l l }
    \toprule
      Author & Year & Criterion & Task \\
    \hline
     Korenberg \cite{korenberg1989robust} & 1989 & MSER & Term selection for NARX model \\
     Chen et al. \cite{chen1989orthogonal} & 1989 & ERR & Term selection for NARMAX model \\
     Chen et al. \cite{chen1991orthogonal}      & 1991 & ERR  & RBF centre selection  \\   
    Wei and Billings \cite{wei2006feature} & 2006 & ERR  & Unsupervised feature selection \\
    Solares et al. \cite{solares2019novel}   & 2019 & OBCC  & Term selection for the logistic-NARX model \\
    Zhang and Lang (This paper) & 2020 & SOCC & Supervised feature selection for classification \\
    \bottomrule
    \end{tabular}}
  \label{tbl:review}
\end{table}

\section{Squared orthogonal correlation coefficients} \label{sec:socc}
\subsection{Definition}
The vectors $\mathbf{y}$ and $\mathbf{w}_i^*$ used in the traditional ERR \eqref{eq:err} normally have non-zero mean values, which makes the connection between ERR and other classic statistics lost.
To overcome this problem, the SOCCs between $\mathbf{X}$ and $\mathbf{y}$ are defined as
\begin{equation} \label{eq:h1}
    h_i = \frac{{\hat{g}_i^{2}{\mathbf{w}_{\text{C}i}^{\top}{\mathbf{w}_{\text{C}i}}} }}{{{{\mathbf{y}_\text{C}}^{\top}}{\mathbf{y}_\text{C}}}}\quad\quad i = 1,\ldots,n\text{,}
\end{equation}
In \eqref{eq:h1}, $\mathbf{y}_\text{C}$ is the centred $\mathbf{y}$, $\mathbf{w}_{\text{C}i}$ is the $i^\text{th}$ column of matrix $\mathbf{W}_\text{C} = ({\mathbf{w}_{\text{C}1}}, \ldots ,{\mathbf{w}_{_\text{C}n}})$ determined from the unnormalised reduced QR decomposition of the centred $\mathbf{X}$, denoted as $\mathbf{X}_\text{C}$, such that
\begin{equation} 
    \mathbf{X}_\text{C} = \mathbf{W}_\text{C}\mathbf{A}
\end{equation}
where $\mathbf{A}$ is the upper triangle matrix of the QR decomposition.
In \eqref{eq:h1}, the parameter $\hat{g}_i$ is obtained from the normal equation
\begin{equation}
    \mathbf{W}_\text{C}^{\top}\mathbf{W}_\text{C}\begin{pmatrix}
    \hat{g}_1\\
    \vdots\\
    \hat{g}_n
    \end{pmatrix} = \mathbf{W}_\text{C}^{\top}\mathbf{y}_\text{C}\text{.}
\end{equation}
As $\mathbf{W}_\text{C}$ is orthogonal, the inner product $\mathbf{W}_\text{C}^{\top}\mathbf{W}_\text{C}$ is the diagonal matrix $\diag{({\mathbf{w}_{\text{C}1}^{\top}\mathbf{w}_{\text{C}1},\ldots,\mathbf{w}_{\text{C}n}^{\top}\mathbf{w}_{\text{C}n}})}$.
Thus, the computation of the parameter vector $\hat{g}_i$ can be simplified as
\begin{equation}
    \hat{g}_i = \frac{\mathbf{w}^{\top}_{\text{C}i}\mathbf{y}_\text{C}}{\mathbf{w}^{\top}_{\text{C}i}\mathbf{w}_{\text{C}i}}\text{.}
\end{equation}
Correspondingly, the SOCCs \eqref{eq:h1} can be rewritten as
\begin{equation} \label{eq:cerrwy}
    h_i = \frac{\mathbf{y}_\text{C}^\top{\mathbf{w}_{\text{C}i}{\mathbf{w}^{\top}_{\text{C}i}}}\mathbf{y}_\text{C}}{{\mathbf{w}_{\text{C}i}^{\top}{\mathbf{w}_{\text{C}i}}}\mathbf{y}_\text{C}^\top\mathbf{y}_\text{C}}\text{,}\quad\quad i = 1,\ldots,n\text{.}
\end{equation}
The SOCCs have a close relationship with the Pearson correlation coefficient, multiple correlation coefficient and the canonical correlation coefficients.

\subsection{Relationship with Pearson correlation coefficient}
The sample Pearson correlation coefficient between $\bm{\upgamma} \in \mathbb{R}^n$ and $\bm{\upomega} \in \mathbb{R}^n$ is given by
\begin{equation}
    r(\bm{\upgamma}, \bm{\upomega}) = \frac{\bm{\upgamma}_{\text{C}}^\top\bm{\upomega}_{\text{C}}}{\sqrt{\bm{\upgamma}_{\text{C}}^\top\bm{\upgamma}_{\text{C}}}\sqrt{\bm{\upomega}_{\text{C}}^\top\bm{\upomega}_{\text{C}}}}\text{,}
\end{equation}
where $\bm{\upgamma}_{\text{C}} \in \mathbb{R}^n$ is the vector $\bm{\upgamma}$ centred by its sample mean and $\bm{\upomega}_{\text{C}} \in \mathbb{R}^n$ is the vector $\bm{\upomega}$ centred by its sample mean.
Obviously, the SOCCs $h_i$ in \eqref{eq:cerrwy} is the squared sample Pearson correlation coefficient between $\mathbf{y}$ and $\mathbf{w}_{\text{C}i}$, i.e.
\begin{equation}
    r^2(\mathbf{y}, \mathbf{w}_{\text{C}i}) = h_i\quad\quad i = 1,\ldots,n\text{.}
\end{equation}

\subsection{Relationship with multiple correlation coefficient}
The multiple correlation coefficient is the measure of linear association between two or more independent variables and a dependent variable.
If the $n$ columns in the design matrix $\mathbf{X}$ are the samples of $n$ independent variables and the response vector $\mathbf{y}$ is the samples of a dependent variable, the association between $\mathbf{X}$ and $\mathbf{y}$ can be measured by the multiple correlation coefficient $R({\mathbf{X},\mathbf{y}})$ or $R({\mathbf{y},\mathbf{X}})$.
The multiple correlation analysis of $\mathbf{X}$ and $\mathbf{y}$ is to find a projection direction, so that the Pearson correlation coefficient between $\mathbf{y}_\text{C}$ and the projected $\mathbf{X}_\text{C}$ is maximised.
The optimal projection direction is the solution $\bm{\hat\upbeta}$ of the normal equation \cite{cohen2014applied}
\begin{equation} \label{eq:body_beta1}
    \left( {\mathbf{X}_\text{C}^{\top}\mathbf{X}_\text{C}} \right)\bm{\hat\upbeta} = \mathbf{X}_\text{C}^{\top}\mathbf{y}_\text{C}\text{.} 
\end{equation}
Then, the multiple correlation coefficient $R({\mathbf{X},\mathbf{y}})$ or $R({\mathbf{y},\mathbf{X}})$ is defined as \cite{cohen2014applied}
\begin{equation} \label{eq:mcc}
    R({\mathbf{X},\mathbf{y}}) = R({\mathbf{y},\mathbf{X}}) = r({\hat{\mathbf{y}}_\text{C},\mathbf{y}_\text{C}}) = \frac{\hat{\mathbf{y}}^\top_\text{C}\mathbf{y}_\text{C}}{\sqrt{\hat{\mathbf{y}}^\top_\text{C}\hat{\mathbf{y}}_\text{C}}\sqrt{\mathbf{y}_\text{C}^\top\mathbf{y}_\text{C}}}\text{,}
\end{equation}
where
\begin{equation}
    \hat{\mathbf{y}}_\text{C} = \mathbf{X}_\text{C}\bm{\hat\upbeta}\text{.}
\end{equation}
The relationship between the SOCCs and the multiple correlation coefficient is (see \ref{s:appendix1} for proof)
\begin{equation} \label{eq:scerr}
    R^2({\mathbf{X},\mathbf{y}}) = \sum_{i = 1}^n h_i\text{.}
\end{equation}
In other words, the sum of the SOCCs between $\mathbf{X}$ and $\mathbf{y}$ is equal to the squared multiple correlation coefficient between $\mathbf{X}$ and $\mathbf{y}$.

\subsection{Relationship with canonical correlation coefficient}
The canonical correlation coefficient is the measure of linear association between two or more independent variables and two or more dependent variables.
Given a response matrix as
\begin{equation}
    \mathbf{Y} = \left({\mathbf{y}_1, \ldots, \mathbf{y}_m}\right) = \begin{pmatrix}
     y_{1,1} & \hdots & y_{1,m}\\
     \vdots & \ddots & \vdots \\
     y_{N,1} & \hdots & y_{N,m}
\end{pmatrix}\text{,}
\end{equation}
if the $n$ columns in the design matrix $\mathbf{X}$ are the samples of $n$ independent variables and the $m$ columns in the response matrix $\mathbf{Y}$ are the samples of $m$ dependent variables, the association between $\mathbf{X}$ and $\mathbf{Y}$ can be measured by the canonical correlation coefficient $R({\mathbf{X},\mathbf{Y}})$.
The Canonical Correlation Analysis (CCA) for $\mathbf{X}$ and $\mathbf{Y}$ is to find a pair of the projection directions $\mathbf{a}$ and $\mathbf{b}$, so that the Pearson correlation coefficient between $\mathbf{X}_\text{C}\mathbf{a}$ and $\mathbf{Y}_\text{C}\mathbf{b}$ is maximised, that is
\begin{equation}
    \argmax_{\mathbf{a},\mathbf{b}} r({\mathbf{X}_\text{C}\mathbf{a},\mathbf{Y}_\text{C}\mathbf{b}})\text{,}
\end{equation}
where 
\begin{equation} 
\begin{split}
    \mathbf{Y}_\text{C} &= \begin{pmatrix}
    \mathbf{y}_{\text{C}1},\ldots,\mathbf{y}_{\text{C}m}
    \end{pmatrix} \\
    &= \begin{pmatrix}
    y_{1,1} - \bar{y}_1 & \hdots & y_{1,m} - \bar{y}_m \\
    \vdots & \ddots & \vdots \\
    y_{N,1} - \bar{y}_1 & \hdots & y_{N,m} - \bar{y}_m
    \end{pmatrix}\text{,}
\end{split}
\end{equation}
and $\bar{y}_i$ is the sample mean of $\mathbf{y}_i$.
The canonical correlation coefficient between $\mathbf{X}$ and $\mathbf{Y}$ can be computed by
\begin{equation}
    R({\mathbf{X},\mathbf{Y}}) = r({\mathbf{X}_\text{C}\mathbf{a},\mathbf{Y}_\text{C}\mathbf{b}}) = \frac{\mathbf{a}^\top\mathbf{R}_{\mathbf{X},\mathbf{Y}}\mathbf{b}}{\sqrt{\mathbf{a}^\top\mathbf{R}_{\mathbf{X},\mathbf{X}}\mathbf{a}}\sqrt{\mathbf{b}^\top\mathbf{R}_{\mathbf{Y},\mathbf{Y}}\mathbf{b}}}\text{,}
\end{equation}
where the correlation matrices are given by
\begin{equation}
    \mathbf{R}_{\mathbf{X},\mathbf{Y}} = \begin{pmatrix}
    r({\mathbf{x}_1,\mathbf{y}_1}) & \dots & r({\mathbf{x}_1,\mathbf{y}_{m}}) \\
    \vdots & \ddots & \vdots \\
    r({\mathbf{x}_n,\mathbf{y}_1}) & \dots & r({\mathbf{x}_n,\mathbf{y}_{m}})
    \end{pmatrix} \;
    \mathbf{R}_{\mathbf{X},\mathbf{X}} = \begin{pmatrix}
    r({\mathbf{x}_1,\mathbf{x}_1}) & \dots & r({\mathbf{x}_1,\mathbf{x}_{n}}) \\
    \vdots & \ddots & \vdots \\
    r({\mathbf{x}_n,\mathbf{x}_1}) & \dots & r({\mathbf{x}_n,\mathbf{x}_{n}})
    \end{pmatrix} \;
    \mathbf{R}_{\mathbf{Y},\mathbf{Y}} = \begin{pmatrix}
    r({\mathbf{y}_1,\mathbf{y}_1}) & \dots & r({\mathbf{y}_1,\mathbf{y}_{m}}) \\
    \vdots & \ddots & \vdots \\
    r({\mathbf{y}_m,\mathbf{y}_1}) & \dots & r({\mathbf{y}_m,\mathbf{y}_{m}})
    \end{pmatrix} \text{.}
\end{equation}
The multiple correlation coefficient $R({\mathbf{X},\mathbf{y}})$ is a special case of the canonical correlation coefficient $R({\mathbf{X},\mathbf{Y}})$, when $\mathbf{Y}$ is a column vector $\mathbf{y}$.
The CCA can be transformed to the eigenvalue problem given by \cite[p.~173]{cooley1971multivariate}
\begin{subequations}\label{eq:ccc0}
\begin{align} 
    \mathbf{R}_{\mathbf{X},\mathbf{X}}^{-1} \mathbf{R}_{\mathbf{X},\mathbf{Y}} \mathbf{R}_{\mathbf{Y},\mathbf{Y}}^{-1} \mathbf{R}_{\mathbf{Y},\mathbf{X}} \mathbf{a} &= R^2({\mathbf{X},\mathbf{Y}}) \mathbf{a} \label{eq:ccca}\\
    \mathbf{R}_{\mathbf{Y},\mathbf{Y}}^{-1} \mathbf{R}_{\mathbf{Y},\mathbf{X}} \mathbf{R}_{\mathbf{X},\mathbf{X}}^{-1} \mathbf{R}_{\mathbf{X},\mathbf{Y}} \mathbf{b} &= R^2({\mathbf{X},\mathbf{Y}}) \mathbf{b} \label{eq:cccb} \text{.}
\end{align}
\end{subequations}
The two projection directions $\mathbf{a}$ and $\mathbf{b}$ are the eigenvectors, and the eigenvalue is the square of the canonical correlation coefficient.
If $\mathbf{X}_\text{C}$ and $\mathbf{Y}_\text{C}$ have full column rank, the number of the non-zero solutions of \eqref{eq:ccc0} is not more than $n \wedge m$, where the operator $\wedge$ returns the minimum of two values on both sides.
Thus, in contrast with the multiple correlation coefficient which only has one value, there are $n \wedge m$ canonical correlation coefficients (which may contain zeros) for $\mathbf{X}$ and $\mathbf{Y}$, which are denoted as $R_1({\mathbf{X},\mathbf{Y}}),\ldots,R_{n \wedge m}({\mathbf{X},\mathbf{Y}})$.

To connect the SOCCs with the canonical correlation coefficients, the SOCCs between $\mathbf{X}$ and $\mathbf{Y}$ are defined as
\begin{equation} \label{eq:hij}
    h_{i,j} = \frac{\mathbf{v}_{\text{C}j}^\top{\mathbf{w}_{\text{C}i}{\mathbf{w}_{\text{C}i}^\top}}\mathbf{v}_{\text{C}j}}
    {{\mathbf{w}_{\text{C}i}^{\top}{\mathbf{w}_{\text{C}i}}}\mathbf{v}_{\text{C}j}^\top\mathbf{v}_{\text{C}j}}\text{,}
\end{equation}
where $\mathbf{w}_{\text{C}i}$ is the $i^\text{th}$ column of matrix $\mathbf{W}_\text{C} = ({\mathbf{w}_{\text{C}1}}, \ldots ,{\mathbf{w}_{_\text{C}n}})$ determined from the unnormalised reduced QR decomposition of $\mathbf{X}_\text{C}$, and $\mathbf{v}_{\text{C}j}$ is the $i^\text{th}$ column of matrix $\mathbf{V}_\text{C} = ({\mathbf{v}_{\text{C}1}}, \ldots ,{\mathbf{v}_{_\text{C}m}})$ determined from the unnormalised reduced QR decomposition of $\mathbf{Y}_\text{C}$, i.e.
\begin{equation}
\begin{split}
    \mathbf{X}_\text{C} &= \mathbf{W}_\text{C}\mathbf{A} \\
    \mathbf{Y}_\text{C} &= \mathbf{V}_\text{C}\mathbf{B}\text{.}
\end{split}
\end{equation}
The matrices $\mathbf{A}$ and $\mathbf{B}$ are the upper triangle matrices of the QR decomposition.
The relationship is (see \ref{s:appendix2} for proof)
\begin{equation} \label{eq:ccch}
    \sum_{k = 1}^{n \wedge m}R_k^2({\mathbf{X},\mathbf{Y}}) = \sum_{i = 1}^{n}\sum_{j = 1}^{m}h_{i,j}\text{,}
\end{equation}
which is a natural extension of \eqref{eq:scerr} to the case where the response vector $\mathbf{y}$ becomes the response matrix $\mathbf{Y}$.

\section{OLS based fast feature selection for binomial classification} \label{sec:bc}
If the $N$ instances of $\mathbf{X}$ belong to two classes and the $n$ variables in $\mathbf{X}$ represent $n$ features, the feature selection problem for the binomial classification is to find the $t$ features from the $n$ features of $\mathbf{X}$, which is optimal to classify the $N$ instances into the two classes.
The two classes can be assigned values 0 and 1 to form a dummy response vector $\mathbf{y}$ for the $N$ instances.
Thus, the goodness of the features for the classification can be evaluated by the multiple correlation coefficient between the features of interest and the dummy response vector.
In fact, the two classes can be assigned to any distinct values to form the response vector, which has no effect on the value of the multiple correlation coefficient.
However, to be consistent to the multinomial classification case, the dummy encoding is adopted.
The optimal $t$ features can be searched by comparing all $\binom{n}{t}$ feature combinations exhaustively, where
\begin{equation}
    \binom{n}{t} = \frac{n!}{t!(n-t)!}\text{.}
\end{equation}
In some cases, the \textit{exhaustive search} is too expensive in computation.
A realistic approach is to select only one feature in one step.
In each step, the previously selected features will not be changed.
For example, the three `optimal' features can be selected in three steps as shown in Table \ref{tbl:greedy}.
As each step selects the feature which maximises the multiple correlation, the search is referred to the \textit{greedy search} \cite{guyon2003introduction}.
The algorithm of the OLS based greedy feature selection for binomial classification can be summarised in 5 steps.

\begin{table}[ht]
\caption{An example for selecting three features from $n$ features by the greedy search for binomial classification, where $i = 1,\ldots,n$ for step 1, $i = 1,2,4,5,\ldots,n$ for step 2, $i = 1,2,4,6,7,\ldots,n$ for step 3.}
  \centering
  \scalebox{0.9}{
    \begin{tabular}{ c c l }
    \toprule
      & Multiple Correlation & Selected Feature \\
    \hline
    Step 1 & $R({\mathbf{x}_3,\mathbf{y}})\geq R({\mathbf{x}_i,\mathbf{y}})$ & $\mathbf{x}_3$ \\
    Step 2 & $R({(\mathbf{x}_3,\mathbf{x}_5),\mathbf{y}})\geq R({(\mathbf{x}_3,\mathbf{x}_i),\mathbf{y}})$ & $\mathbf{x}_3,\mathbf{x}_5$ \\
    Step 3 & $R({(\mathbf{x}_3,\mathbf{x}_5,\mathbf{x}_1),\mathbf{y}})\geq R({(\mathbf{x}_3,\mathbf{x}_5,\mathbf{x}_i),\mathbf{y}})$ & $\mathbf{x}_3,\mathbf{x}_5,\mathbf{x}_1$ \\
    \bottomrule
    \end{tabular}}
  \label{tbl:greedy}
\end{table}

\bigskip
\textbf{Input}: \\
$\mathbf{X}$: $N \times n$ matrix containing $N$ instances and $n$ features.\\
$\mathbf{y}$: $N \times 1$ vector formed by dummy encoding.\\
$t$: The number of features to be selected.\medskip

\textbf{Step 1.} Centre $\mathbf{X}$ into $\mathbf{X}_{\text{C}}$, centre $\mathbf{y}$ into $\mathbf{y}_{\text{C}}$, and let $p=0$. \medskip

\textbf{Step 2.} Divide $\mathbf{X}$ into $(\mathbf{X}_\text{s}, \mathbf{X}_\text{r})$, where the selected feature matrix is given by
\begin{equation}
    \mathbf{X}_\text{s} = \left(\mathbf{x}_{\text{s}1},\ldots,\mathbf{x}_{\text{s}p}\right)\text{,}
\end{equation}
and the remaining feature matrix is given by
\begin{equation}
    \mathbf{X}_\text{r} = \left(\mathbf{x}_{\text{r}1},\ldots,\mathbf{x}_{\text{r}q}\right)\text{,}
\end{equation}
where $p$ is the number of the selected features, and $q$ is the number of the remaining features.
Correspondingly, divide $\mathbf{X}_\text{C}$ into $(\mathbf{X}_\text{Cs}, \mathbf{X}_\text{Cr})$, where
\begin{equation}
\begin{split}
    \mathbf{X}_\text{Cs} &= \left(\mathbf{x}_{\text{Cs}1},\ldots,\mathbf{x}_{\text{Cs}p}\right) \\
    \mathbf{X}_\text{Cr} &= \left(\mathbf{x}_{\text{Cr}1},\ldots,\mathbf{x}_{\text{Cr}q}\right)\text{.}
\end{split}
\end{equation}\medskip

\textbf{Step 3.} If $p = 0$, let
\begin{equation}
\begin{split}
    \mathbf{W}_{\text{Cr}} &= \mathbf{X}_{\text{Cr}} \\
    \mathbf{w}_{\text{Cr}i} &= \mathbf{x}_{\text{Cr}i}\text{,}\quad i = 1,\ldots,q\text{.}
\end{split}
\end{equation}
Otherwise, first, orthogonalise $\mathbf{X}_{\text{Cs}}$ into $\mathbf{W}_{\text{Cs}}$, where
\begin{equation}
    \mathbf{W}_{\text{Cs}} = \left(\mathbf{w}_{\text{Cs}1},\ldots,\mathbf{w}_{\text{Cs}p}\right)\text{,}
\end{equation}
and $\mathbf{w}_{\text{Cs}i}^\top\mathbf{w}_{\text{Cs}j} = 0$ for $i \neq j$.
Then, orthogonalise each feature in $\mathbf{X}_{\text{C}r}$ to $\mathbf{W}_{\text{Cs}}$ to form the matrix $\mathbf{W}_{\text{Cr}}$, where
\begin{equation}
    \mathbf{W}_{\text{Cr}} = \left(\mathbf{w}_{\text{Cr}1},\ldots,\mathbf{w}_{\text{Cr}q}\right)\text{,}
\end{equation}
and $\mathbf{w}_{\text{Cr}i}$ is obtained through the classical Gram-Schmidt process, which is given by
\begin{equation} 
    \mathbf{w}_{\text{Cr}i} = \mathbf{x}_{\text{Cr}i} - \sum_{j = 1}^{p}\frac{\mathbf{x}_{\text{Cr}i}^\top\mathbf{w}_{\text{Cs}j}}
    {\mathbf{w}_{\text{Cs}j}^\top\mathbf{w}_{\text{Cs}j}}\mathbf{w}_{\text{Cs}j}\text{,}\quad i = 1,\ldots,q \text{.}
\end{equation}
It should be noticed that $\mathbf{w}_{\text{Cr}i}$ is orthogonal to $\mathbf{W}_{\text{Cs}}$ but not to $\mathbf{W}_{\text{Cr}}$, that is $\mathbf{w}_{\text{Cr}i}^\top\mathbf{w}_{\text{Cs}j} = 0$ but $\mathbf{w}_{\text{Cr}i}^\top\mathbf{w}_{\text{Cr}j} \neq 0$.\medskip

\textbf{Step 4.} Compute $r^2({\mathbf{w}_{\text{Cr}i},\mathbf{y}_\text{C}})$ by
\begin{equation}
    r^2({\mathbf{w}_{\text{Cr}i},\mathbf{y}_\text{C}}) = h_i\text{,}\quad i = 1,\ldots,q \text{,}
\end{equation}
where
\begin{equation}
    h_i = \frac{\mathbf{y}_{\text{C}}^\top{\mathbf{w}_{\text{Cr}i}{\mathbf{w}_{\text{Cr}i}^\top}}\mathbf{y}_{\text{C}}}
    {{\mathbf{w}_{\text{Cr}i}^{\top}{\mathbf{w}_{\text{Cr}i}}}\mathbf{y}_{\text{C}}^\top\mathbf{y}_{\text{C}}}\text{.}
\end{equation}\medskip

\textbf{Step 5.} Find an $i$ which maximises $r^2({\mathbf{w}_{\text{Cr}i},\mathbf{y}_\text{C}})$ such that
\begin{equation}
    i_\text{max} = \argmax_{i} r^2({\mathbf{w}_{\text{Cr}i},\mathbf{y}_\text{C}})\text{.}
\end{equation}
Then, remove $\mathbf{x}_{i_\text{max}}$ from $\mathbf{X}_\text{r}$, add it into $\mathbf{X}_\text{s}$, reduce $q$ by 1, and increase $p$ by 1.
After that, return to \textbf{Step 2} until $p=t$, when \textbf{Output} $\mathbf{X}_\text{s}$ to complete the feature selection. 

The pseudocode of the algorithm is given in Algorithm \ref{al:bin}.\bigskip

\begin{algorithm}[ht] 
    \KwIn{$\mathbf{X}$, $\mathbf{y}$, $t$}
    \KwOut{$\mathbf{X}_\text{s}$}
    Centre $\mathbf{X}$ and $\mathbf{y}$ to $\mathbf{X}_{\text{C}}$ and $\mathbf{y}_{\text{C}}$\;
    $p \leftarrow 0$\;
    \While{$p<t$}{
    Divide $\mathbf{X}_\text{C}$ into the selected centred features $\mathbf{X}_\text{Cs}$ and the remaining centred feature $\mathbf{X}_\text{Cr}$\;
    \eIf{$p = 0$}{
    $\mathbf{W}_{\text{Cr}} \leftarrow \mathbf{X}_{\text{Cr}}$, which is composed of $(\mathbf{w}_{\text{Cr}1},\ldots,\mathbf{w}_{\text{Cr}q})$\;
    }{
    Orthogonalise $\mathbf{X}_{\text{Cs}}$ to itself to form $\mathbf{W}_{\text{Cs}}$, which is composed of $(\mathbf{w}_{\text{Cs}1},\ldots,\mathbf{w}_{\text{Cs}p})$\;
    Orthogonalise $\mathbf{X}_{\text{Cr}}$ to $\mathbf{W}_{\text{Cs}}$ to form $\mathbf{W}_{\text{Cr}}$, which is composed of $(\mathbf{w}_{\text{Cr}1},\ldots,\mathbf{w}_{\text{Cr}q})$\;
    }
    Compute $r^2({\mathbf{w}_{\text{Cr}i},\mathbf{y}_\text{C}})$ by \eqref{eq:cerrwy}\;
    Find feature index $i_\text{max}$, such that $r^2({\mathbf{w}_{\text{Cr}i},\mathbf{y}_\text{C}})$ is maximum with $i \in \{1,\ldots,n\}$\;
    Select feature $\mathbf{x}_{i_\text{max}}$ into $\mathbf{X}_\text{s}$\;
    $p \leftarrow p+1$\;
    }
    \caption{Pseudocode of the OLS based feature selection for binomial classification.}\label{al:bin}
\end{algorithm}

The multiple correlation coefficient can be obtained either using the definition \eqref{eq:mcc} or the sum of the SOCCs \eqref{eq:scerr}.
In the greedy search, the OLS based feature selection method has the computational speed advantage over the definition based feature selection method.
The computation complexity of the two feature selection methods can be explicitly compared by the asymptotic upper bound notation $\mathcal{O}$ \cite[p.~47]{cormen2009introduction}.
At Step $k$ of the greedy search, the $k-1$ optimal features have been selected, and the rest of the $n-k+1$ features are the candidates of the $k^\text{th}$ optimal feature.
The candidate feature matrix is a $N \times k$ matrix composed of the $k-1$ selected features and a candidate feature.
According to the normal equation \eqref{eq:body_beta1}, the definition based feature selection method is dominated by computing the inner product of the $N \times k$ centred candidate feature matrix.
The computational complexity of the inner product of one centred candidate matrix is $\mathcal{O}(k^2 N)$.
There are $n-k+1$ candidate features, so the complexity for Step $k$ is
\begin{equation}
    (n-k+1)\mathcal{O}(k^2 N) = \mathcal{O}(k^2 n N)\text{.}
\end{equation}
Thus, the overall complexity for $t$ features selection is given by
\begin{equation}
    \sum_{k = 1}^{t}\mathcal{O}(k^2 n N) = \mathcal{O}\left(\sum_{k = 1}^{t}k^2 n N\right) = \mathcal{O}(t^3 n N).
\end{equation}
For OLS based feature selection, as the SOCCs of the selected features ($h_1$ to $h_{k-1}$) have been computed in Step 1 to Step $k-1$, only the SOCCs of the candidate feature ($h_k$) is required to compute.
Thus, OLS based feature selection is dominated by the classical Gram-Schmidt orthogonalisation process.
At Step $k$ of the greedy search, the computational complexity of the orthogonalisation of one candidate feature is $\mathcal{O}(k N)$.
There are $n-k+1$ candidate features, so the complexity for Step $k$ is
\begin{equation}
    (n-k+1)\mathcal{O}(k N) = \mathcal{O}(k n N)\text{.}
\end{equation}
Thus, the overall complexity for $t$ features selection is given by
\begin{equation}
    \sum_{k = 1}^{t}\mathcal{O}(k n N) = \mathcal{O}\left(\sum_{k = 1}^{t} k n N\right) = \mathcal{O}(t^2 n N).
\end{equation}
Consequently, compared to the definition based feature selection method, the OLS based feature selection method has a significant computational speed advantage in the greedy search.

\section{OLS based fast feature selection for multinomial classification} \label{sec:mc}
If the $N$ instances of $\mathbf{X}$ belong to $c$ classes, where $c \leq N$, and the $n$ columns in $\mathbf{X}$ represent $n$ features, the feature selection problem for the multinomial classification is to find the $t$ features from the $n$ features of $\mathbf{X}$, which is optimal to classify the $N$ instances into the $c$ classes.
Similar to the last section, the $c$ classes can be encoded to certain values to form a response variable.
The ordinal encoding is to assign $1,\ldots,c$ to the $c$ labels to form a vector $\mathbf{y}$.
Then, the multiple correlation coefficient between the features and $\mathbf{y}$ can be adopted to indicate the goodness of the features for the classification.
The $c$ labels can also be encoded to form a matrix $\mathbf{Y}$ using, e.g. $c$-label dummy encoding (or called one-hot encoding), $c-1$-label dummy encoding, effects encoding, and contrast encoding \cite[Chapter~5]{cohen2014applied}.
When the response is encoded as a matrix $\mathbf{Y}$, the canonical correlation coefficients between $\mathbf{X}$ and $\mathbf{Y}$ can be used as the feature selection criterion.
Similar to the last section, an example of the greed search for multinomial classification is illustrated in Table \ref{tbl:greedym}, where the response is encoded as an $N \times c-1$ matrix $\mathbf{Y}$ and the ranking criterion is the sum of the squared canonical correlation coefficients.

\begin{table}[ht]
\caption{An example for selecting three features from $n$ features by the greedy search for multinomial classification, where $i = 1,\ldots,n$ for step 1, $i = 1,2,4,5,\ldots,n$ for step 2, $i = 1,2,4,6,7,\ldots,n$ for step 3.}
  \centering
  \scalebox{0.9}{
    \begin{tabular}{ c c l }
    \toprule
      & \makecell{Ranking \\ Criterion} & \makecell{Selected \\ Feature} \\
    \hline
     1 & $\sum\limits_{k = 1}^{1\wedge c-1}R_k^2({\mathbf{x}_3,\mathbf{Y}}) \geq \sum\limits_{k = 1}^{1\wedge c-1}R_k^2({\mathbf{x}_i,\mathbf{Y}})$ & $\mathbf{x}_3$ \\
     2 & $\sum\limits_{k = 1}^{2\wedge c-1}R_k({(\mathbf{x}_3,\mathbf{x}_5),\mathbf{Y}})\geq \sum\limits_{k = 1}^{2\wedge c-1}R_k({(\mathbf{x}_3,\mathbf{x}_i),\mathbf{Y}})$ & $\mathbf{x}_3,\mathbf{x}_5$ \\
     3 & $\sum\limits_{k = 1}^{3\wedge c-1}R_k({(\mathbf{x}_3,\mathbf{x}_5,\mathbf{x}_1),\mathbf{Y}})\geq \sum\limits_{k = 1}^{3\wedge c-1}R_k({(\mathbf{x}_3,\mathbf{x}_5,\mathbf{x}_i),\mathbf{Y}})$ & $\mathbf{x}_3,\mathbf{x}_5,\mathbf{x}_1$ \\
    \bottomrule
    \end{tabular}}
  \label{tbl:greedym}
\end{table}

In the following, the relationship between CCA and a classical LDA classifier will first be demonstrated to reveal the implication of the canonical correlation coefficient in a classification problem.
Then, the algorithm of the OLS based feature selection for multinomial classification is developed where the sum of the squared canonical correlation coefficients will be used as the feature ranking criterion.
After that, a version of the algorithm that can be used to deal with categorical features is presented.

\subsection{Relationship with linear discriminant analysis} \label{ss:lda}
As the feature selection is used for the multinomial classification, it is reasonable to know the performance of the features in the Linear Discriminant Analysis (LDA), where the label encoding is not required.
For the convenience of LDA, the feature matrix $\mathbf{X}$ is decomposed into $\mathbf{X}^{(1)},\ldots,\mathbf{X}^{(c)}$, where the $N_i \times n$ matrix $\mathbf{X}^{(i)}$ represents the $N_i$ instances belonged to the $i^\text{th}$ class.
The within-class scatter matrix of the samples is
\begin{equation}
    \mathbf{S}_w = \sum_{i = 1}^c \left(\mathbf{X}^{(i)} - \mathbf{1}^{(i)}\bar{\mathbf{x}}^{(i)\top}\right)^\top\left(\mathbf{X}^{(i)} - \mathbf{1}^{(i)}\bar{\mathbf{x}}^{(i)\top}\right)\text{,}
\end{equation}
where $\bar{\mathbf{x}}^{(i)}$ is the sample mean of each feature in $\mathbf{X}^{(i)}$ given by
\begin{equation}
    \bar{\mathbf{x}}^{(i)} = \left({\bar{x}_1^{(i)},\ldots,\bar{x}_n^{(i)}}\right)^\top
\end{equation}
and $\mathbf{1}^{i}$ is $N_i \times 1$ vector of ones.
The between-class scatter matrix of the samples is
\begin{equation}
    \mathbf{S}_b = \sum_{i = 1}^c N_i \left({\bar{\mathbf{x}}^{(i)} - \bar{\mathbf{x}}}\right)\left({\bar{\mathbf{x}}^{(i)} - \bar{\mathbf{x}}}\right)^\top\text{,}
\end{equation}
where $\bar{\mathbf{x}}$ is the overall sample mean of each feature.
The aim of LDA is to find a projection direction $\mathbf{d}$ for $\mathbf{X}$, so that the ratio between the projected between-class scatter and the projected within-class scatter is maximised.
The ratio is called Fisher's criterion, which is given by
\begin{equation}
    J = \frac{\mathbf{d}^\top\mathbf{S}_b\mathbf{d}}{\mathbf{d}^\top\mathbf{S}_w\mathbf{d}}\text{.}
\end{equation}
The larger Fisher's criterion $J$ implies the better the separation of the $c$ classes.
The LDA can be transformed to the eigenvalue problem given by \cite[p.~246]{cooley1971multivariate}
\begin{equation} \label{eq:lda}
    \mathbf{S}_w^{-1}\mathbf{S}_b\mathbf{d} = J\mathbf{d}\text{,}
\end{equation}
where the eigenvector is the optimal projection direction $\mathbf{d}$ and the eigenvalue is the maximised Fisher's criterion $J$.

The relationship between LDA and CCA can be found when $\mathbf{Y}$ is formed by $c$ or $c-1$-label dummy encoding.
Under the two encoding schemes, the eigenvalue problem \eqref{eq:ccca} can be rewritten as \cite{sun2007class}
\begin{equation}
    \left({\mathbf{S}_b+\mathbf{S}_w}\right)^{-1}\mathbf{S}_b\mathbf{a} = R^2({\mathbf{X},\mathbf{Y}})\mathbf{a}\text{,}
\end{equation}
or in the form of
\begin{equation} \label{eq:ccalda}
    \mathbf{S}_w^{-1}\mathbf{S}_b\mathbf{a} = \frac{R^2({\mathbf{X},\mathbf{Y}})}{1-R^2({\mathbf{X},\mathbf{Y}})}\mathbf{a}\text{.}
\end{equation}
Comparing \eqref{eq:lda} and \eqref{eq:ccalda}, it is found that LDA and CCA are equivalent, and Fisher's criterion of LDA can be evaluated by
\begin{equation} \label{eq:jr}
    J = \frac{R^2({\mathbf{X},\mathbf{Y}})}{1-R^2({\mathbf{X},\mathbf{Y}})}\text{.}
\end{equation}
indicating that CCA has a close relationship with the Fisher's criterion.

To better reveal this relationship, the $c-1$-label dummy encoding and the canonical correlation coefficients are adopted, instead of the ordinal encoding and the multiple correlation coefficient.
The $c-1$-label dummy encoding constructs a $N \times c-1$ matrix $Y = (y_{i,j})$, where
\begin{equation}
    y_{i,j} = \left\{\begin{aligned}
        1 \quad\quad & i^\text{th} \text{ instance is belonged to } j^\text{th} \text{ class} \\
        0 \quad\quad & \text{otherwise.}
    \end{aligned}\right.
\end{equation}
Thus, the dummy response in the last section is a special case of $c-1$-label dummy encoding where $c = 2$.

\subsection{OLS based feature selection algorithm} \label{ss:ols}
The algorithm of OLS based feature selection for multinomial classification can be summarised in 5 steps.

\bigskip
\textbf{Input}: \\
$\mathbf{X}$: $N \times n$ matrix containing $N$ instances and $n$ features.\\
$\mathbf{Y}$: $N \times c-1$ matrix formed by $c-1$-label dummy encoding.\\
$t$: The number of features to be selected.\medskip

\textbf{Step 1.} Centre $\mathbf{Y}$ into $\mathbf{Y}_{\text{C}}$, orthogonalise $\mathbf{Y}_{\text{C}}$ into $\mathbf{V}_{\text{C}}$, centre $\mathbf{X}$ into $\mathbf{X}_{\text{C}}$, and let $p=0$. \medskip

\textbf{Step 2.} Divide $\mathbf{X}$ into $(\mathbf{X}_\text{s}, \mathbf{X}_\text{r})$, where the selected feature matrix is given by
\begin{equation}
    \mathbf{X}_\text{s} = \left(\mathbf{x}_{\text{s}1},\ldots,\mathbf{x}_{\text{s}p}\right)\text{,}
\end{equation}
and the remaining feature matrix is given by
\begin{equation}
    \mathbf{X}_\text{r} = \left(\mathbf{x}_{\text{r}1},\ldots,\mathbf{x}_{\text{r}q}\right)\text{,}
\end{equation}
where $p$ is the number of the selected features, and $q$ is the number of the remaining features.
Correspondingly, divide $\mathbf{X}_\text{C}$ into $(\mathbf{X}_\text{Cs}, \mathbf{X}_\text{Cr})$, where
\begin{equation}
\begin{split}
    \mathbf{X}_\text{Cs} &= \left(\mathbf{x}_{\text{Cs}1},\ldots,\mathbf{x}_{\text{Cs}p}\right) \\
    \mathbf{X}_\text{Cr} &= \left(\mathbf{x}_{\text{Cr}1},\ldots,\mathbf{x}_{\text{Cr}q}\right)\text{.}
\end{split}
\end{equation}\medskip

\textbf{Step 3.} If $p = 0$, let
\begin{equation}
\begin{split}
    \mathbf{W}_{\text{Cr}} &= \mathbf{X}_{\text{Cr}} \\
    \mathbf{w}_{\text{Cr}i} &= \mathbf{x}_{\text{Cr}i}\text{,}\quad i = 1,\ldots,q\text{.}
\end{split}
\end{equation}
Otherwise, first, orthogonalise $\mathbf{X}_{\text{Cs}}$ into $\mathbf{W}_{\text{Cs}}$, where
\begin{equation}
    \mathbf{W}_{\text{Cs}} = \left(\mathbf{w}_{\text{Cs}1},\ldots,\mathbf{w}_{\text{Cs}p}\right)\text{,}
\end{equation}
and $\mathbf{w}_{\text{Cs}i}^\top\mathbf{w}_{\text{Cs}j} = 0$ for $i \neq j$.
Then, orthogonalise each feature in $\mathbf{X}_{\text{C}r}$ to $\mathbf{W}_{\text{Cs}}$ to form the matrix $\mathbf{W}_{\text{Cr}}$, where
\begin{equation}
    \mathbf{W}_{\text{Cr}} = \left(\mathbf{w}_{\text{Cr}1},\ldots,\mathbf{w}_{\text{Cr}q}\right)\text{,}
\end{equation}
and $\mathbf{w}_{\text{Cr}i}$ is obtained through the classical Gram-Schmidt process, which is given by
\begin{equation} \label{eq:algorth}
    \mathbf{w}_{\text{Cr}i} = \mathbf{x}_{\text{Cr}i} - \sum_{j = 1}^{p}\frac{\mathbf{x}_{\text{Cr}i}^\top\mathbf{w}_{\text{Cs}j}}
    {\mathbf{w}_{\text{Cs}j}^\top\mathbf{w}_{\text{Cs}j}}\mathbf{w}_{\text{Cs}j}\text{,}\quad i = 1,\ldots,q \text{.}
\end{equation}
It should be noticed that $\mathbf{w}_{\text{Cr}i}$ is orthogonal to $\mathbf{W}_{\text{Cs}}$ but not to $\mathbf{W}_{\text{Cr}}$, that is $\mathbf{w}_{\text{Cr}i}^\top\mathbf{w}_{\text{Cs}j} = 0$ but $\mathbf{w}_{\text{Cr}i}^\top\mathbf{w}_{\text{Cr}j} \neq 0$.\medskip

\textbf{Step 4.} Compute $R^2({\mathbf{w}_{\text{Cr}i},\mathbf{V}_\text{C}})$ by
\begin{equation} \label{eq:wcriVc}
    R^2({\mathbf{w}_{\text{Cr}i},\mathbf{V}_\text{C}}) = \sum_{j = 1}^{c-1}h_{i,j}\text{,}\quad i = 1,\ldots,q \text{,}
\end{equation}
where
\begin{equation}
    h_{i,j} = \frac{\mathbf{v}_{\text{C}j}^\top{\mathbf{w}_{\text{Cr}i}{\mathbf{w}_{\text{Cr}i}^\top}}\mathbf{v}_{\text{C}j}}
    {{\mathbf{w}_{\text{Cr}i}^{\top}{\mathbf{w}_{\text{Cr}i}}}\mathbf{v}_{\text{C}j}^\top\mathbf{v}_{\text{C}j}}\text{.}
\end{equation}\medskip

\textbf{Step 5.} Find an $i$ which maximises $R^2({\mathbf{w}_{\text{Cr}i},\mathbf{V}_\text{C}})$ such that
\begin{equation}
    i_\text{max} = \argmax_{i} R^2({\mathbf{w}_{\text{Cr}i},\mathbf{V}_\text{C}})\text{.}
\end{equation}
Then, remove $\mathbf{x}_{i_\text{max}}$ from $\mathbf{X}_\text{r}$, add it into $\mathbf{X}_\text{s}$, reduce $q$ by 1, and increase $p$ by 1.
After that, return to \textbf{Step 2} until $p=t$, when \textbf{Output} $\mathbf{X}_\text{s}$ to complete the feature selection. 

The pseudocode of the algorithm is given in Algorithm \ref{al:multi}.\bigskip

\begin{algorithm}[ht] 
    \KwIn{$\mathbf{X}$, $\mathbf{Y}$, $t$}
    \KwOut{$\mathbf{X}_\text{s}$}
    Centre $\mathbf{X}$ and $\mathbf{Y}$ to $\mathbf{X}_{\text{C}}$ and $\mathbf{Y}_{\text{C}}$\;
    Orthogonalise $\mathbf{Y}_{\text{C}}$ to itself to form $\mathbf{V}_{\text{C}}$\;
    $p \leftarrow 0$\;
    \While{$p<t$}{
    Divide $\mathbf{X}_\text{C}$ into the selected centred features $\mathbf{X}_\text{Cs}$ and the remaining centred feature $\mathbf{X}_\text{Cr}$\;
    \eIf{$p = 0$}{
    $\mathbf{W}_{\text{Cr}} \leftarrow \mathbf{X}_{\text{Cr}}$, which is composed of $(\mathbf{w}_{\text{Cr}1},\ldots,\mathbf{w}_{\text{Cr}q})$\;
    }{
    Orthogonalise $\mathbf{X}_{\text{Cs}}$ to itself to form $\mathbf{W}_{\text{Cs}}$, which is composed of $(\mathbf{w}_{\text{Cs}1},\ldots,\mathbf{w}_{\text{Cs}p})$\;
    Orthogonalise $\mathbf{X}_{\text{Cr}}$ to $\mathbf{W}_{\text{Cs}}$ to form $\mathbf{W}_{\text{Cr}}$, which is composed of $(\mathbf{w}_{\text{Cr}1},\ldots,\mathbf{w}_{\text{Cr}q})$\;
    }
    Compute $R^2({\mathbf{w}_{\text{Cr}i},\mathbf{V}_\text{C}})$ by \eqref{eq:hij}\;
    Find feature index $i_\text{max}$, such that $R^2({\mathbf{w}_{\text{Cr}i},\mathbf{V}_\text{C}})$ is maximum with $i \in \{1,\ldots,n\}$\;
    Select feature $\mathbf{x}_{i_\text{max}}$ into $\mathbf{X}_\text{s}$\;
    $p \leftarrow p+1$\;
    }
    \caption{Pseudocode of the OLS based feature selection for multinomial classification.}\label{al:multi}
\end{algorithm}

The speed advantage of the OLS based feature selection method is reflected in \textbf{Step 4}.
To evaluate the goodness of the candidate feature $\mathbf{x}_{\text{r}i}$, CCA requires to compute the canonical correlation coefficient $R({(\mathbf{X}_{\text{s}},\mathbf{x}_{\text{r}i}),\mathbf{Y}})$, while OLS only needs to compute the multiple correlation coefficient $R({\mathbf{w}_{\text{Cr}i},\mathbf{V}_\text{C}})$, because
\begin{equation}
\begin{split}
    \sum_{k = 1}^{p+1\wedge c-1}R_k^2({(\mathbf{X}_{\text{s}},\mathbf{x}_{\text{r}i}),\mathbf{Y}}) &= \sum_{k = 1}^{{p+1\wedge c-1}}R_k^2({(\mathbf{W}_{\text{Cs}},\mathbf{w}_{\text{Cr}i}),\mathbf{V}_\text{C}}) \\
    &= \sum_{k = 1}^{{p\wedge c-1}}R_k^2({\mathbf{W}_{\text{Cs}},\mathbf{V}_\text{C}}) + R^2({\mathbf{w}_{\text{Cr}i},\mathbf{V}_\text{C}})\text{.}
\end{split}
\end{equation}
For each candidate feature $\mathbf{x}_{\text{r}i}$, $R({\mathbf{W}_{\text{Cs}},\mathbf{V}_\text{C}})$ is the same.
Thus, to find the maximal $R({(\mathbf{X}_{\text{s}},\mathbf{x}_{\text{r}i}),\mathbf{Y}})$, only $R({\mathbf{w}_{\text{Cr}i},\mathbf{V}_\text{C}})$ is required to compute.
In addition, although the multiple correlation coefficient $R({\mathbf{w}_{\text{C}i}, \mathbf{Y}})$, which is equal to $R({\mathbf{w}_{\text{C}i}, \mathbf{V}_{\text{C}}})$, can be computed through the definition \eqref{eq:mcc}, OLS provides a faster way of computation.
Equation \eqref{eq:mcc} requires to solve the normal equation which is dominated by the inner product of the $N \times c-1$ matrix $\mathbf{Y}_\text{C}$, whose computational complexity is $\mathcal{O}\left({c^2N}\right)$. 
For OLS, as the orthogonalisation of $\mathbf{Y}$ is only required once in \textbf{Step 1}, the dominant part of computation is \eqref{eq:wcriVc} whose computational complexity is only $\mathcal{O}\left(cN\right)$.

As the above introduction of the OLS based algorithm is basically conceptual, some speed optimisation steps have been omitted.
For example, in \textbf{Step 3}, $\mathbf{W}_{\textbf{Cs}}$ computed for selecting the $i^\text{th}$ optimal feature can be reused for selecting the $i+1^\text{th}$ optimal feature.
The further optimisation of the OLS  speed can be found in the original paper of OLS based model term selection \cite{chen1989orthogonal}.

\subsection{Dealing with categorical features} \label{ss:olsd}
When the features are categorical, the feature encoding is required for OLS based feature selection.
In the previous analysis, $n$ features are represented by $n$ column vectors in $\mathbf{X}$, but some encoding methods may encode the categorical features into matrices.
In these cases, the feature matrix is composed of $n$ submatrices, that is
\begin{equation}
    \mathbf{X} = \left(\mathbf{X}_1, \ldots, \mathbf{X}_n\right)\text{,} 
\end{equation}
where the matrix $\mathbf{X}_i$ is the encoded $i^\text{th}$ feature.
An OLS based feature selection algorithm similar to the algorithm in Section \ref{ss:ols} can be applied to the matrix encoded features.
The candidate orthogonal feature matrix in \textbf{Step 3} in Section \ref{ss:ols} is given by
\begin{equation}
    \mathbf{W}_{\text{Cr}} = \left(\mathbf{W}_{\text{Cr}1},\ldots,\mathbf{W}_{\text{Cr}q}\right)\text{,}
\end{equation}
where
\begin{equation}
    \mathbf{W}_{\text{Cr}i} = \left(\mathbf{w}^{[1]}_{\text{Cr}i},\ldots,\mathbf{w}^{[z_i]}_{\text{Cr}i}\right)\text{.}
\end{equation}
is a $N \times z_i$ matrix.
Besides being orthogonal to the selected orthogonal feature matrix $\mathbf{W}_{\text{Cs}}$, the submatrix $\mathbf{W}_{\text{Cr}i}$ should be column-wise orthogonal via an additional orthogonalisation process.
In \textbf{Step 4}, the sum of the squared canonical correlation coefficients can be computed by
\begin{equation}
    \sum_{k = 1}^{z_i\wedge c-1}R_k^2({\mathbf{W}_{\text{Cr}i},\mathbf{V}_\text{C}}) = \sum_{j = 1}^{c-1}\sum_{g = 1}^{z_i}h_{i,j}^{[g]}\text{,}\quad i = 1,\ldots,q \text{,}
\end{equation}
where
\begin{equation}
    h_{i,j}^{[g]} = \frac{\mathbf{v}_{\text{C}j}^\top{\mathbf{w}_{\text{Cr}i}^{[g]}{\mathbf{w}_{\text{Cr}i}^{[g]\top}}}\mathbf{v}_{\text{C}j}}
    {{\mathbf{w}_{\text{Cr}i}^{[g]\top}{\mathbf{w}^{[g]}_{\text{Cr}i}}}\mathbf{v}_{\text{C}j}^\top\mathbf{v}_{\text{C}j}}\text{.}
\end{equation}
Finally, the sum of the squared canonical correlation coefficients are used to rank the features for \textbf{Step 5}.

\section{Empirical study} \label{sec:cs}
In this section, firstly, a simple example is used to illustrate the procedure of the OLS based feature selection method when applied to the Fisher's iris data \cite{fisher1936use}.
The relationship between the SOCCs with canonical correlation coefficient and Fisher's criterion is demonstrated via this case study.
Then, the OLS based feature selection methods are compared with mutual information based filter methods and the embedded methods using both synthetic and real world datasets.
For the filter methods, the features are selected via greedy search with different ranking criteria.
The mutual information based feature selection methods in this comparison are summarised in \cite{brown2012conditional}.
The ranking criteria are the difference and quotient schemes (mRMRd and mRMRq), Mutual Information Maximisation (MIM), Joint Mutual Information (JMI), Conditional Mutual Information Maximisation (CMIM), Conditional Infomax Feature Extraction (CIFE), Interaction Capping (ICAP), and Double Input Symmetrical Relevance (DISR).
The embedded methods are LASSO and elastic net (Net), and the \textsc{Matlab} function \texttt{lasso} is adopted.

The OLS based method shows superiority in computation speed compared to the direct use of the definition of the canonical correlation coefficient.
For example, in the two real world datasets (Dexter and Gisette), the OLS method takes 401 ms for Dexter and 5109 ms for Gisette to select 20 features on a 2.6 GHz personal laptop, while the traditional definition based method takes 13200 ms and 134922 ms, respectively.
The empirical studies are implemented in \textsc{Matlab} R2021a, and the code will be published in GitHub\footnote{\url{https://github.com/MatthewSZhang/fs_ols}}.

\subsection{An illustration of the OLS based feature selection}
\begin{table}[ht]
\caption{Fisher's Iris Dataset.}
  \centering
  \scalebox{0.9}{
    \begin{tabular}{ c c c c c }
    \toprule
     \makecell{Sepal\\Length} & \makecell{Sepal\\Width} & \makecell{Petal\\Length} & \makecell{Petal\\Width} & Species \\
    \hline
    5.1	& 3.5 & 1.4 & 0.2 & setosa \\
    4.9 & 3 & 1.4 & 0.2 & setosa \\
    7 & 3.2 & 4.7 & 1.4 & versicolor \\
    6.4 & 3.2 & 4.5 & 1.5 & versicolor \\
    6.3 & 3.3 & 6 & 2.5 & virginica \\
    5.8 & 2.7 & 5.1 & 1.9 & virginica \\
    7.1 & 3 & 5.9 & 2.1 & virginica \\
    \bottomrule
    \end{tabular}}
  \label{tbl:fisher}
\end{table}

The Fisher's iris data are given in Table \ref{tbl:fisher}.
The 7 instances have 4 features and 3 classes, so $N = 7$, $n = 4$, and $c = 3$.
The objective of the feature selection is to find 3 optimal features for the 3 species classification.

The feature matrix is $\mathbf{X}$ and the $c-1$-label dummy encoded response is 
\begin{equation}
    \mathbf{Y} = \begin{pmatrix}
    1 & 1 & 0 & 0 & 0 & 0 & 0 \\
    0 & 0 & 1 & 1 & 0 & 0 & 0 \\
    \end{pmatrix}^\top\text{,}
\end{equation}
where $(1, 0)$ represents setosa, $(0, 1)$ represents versicolor, and $(0, 0)$ represents virginica.
Following the algorithm introduced in Section \ref{ss:ols}, the procedure of the OLS based feature selection method is shown below.

\textbf{Step 1.} First, centre $\mathbf{Y}$ into $\mathbf{Y}_\text{C}$.
Second, orthogonalise $\mathbf{Y}_\text{C}$ into $\mathbf{V}_\text{C}$.
Through the classical Gram-Schmidt process, use the first column of $\mathbf{Y}_\text{C}$ as $\mathbf{v}_{\text{C}1}$, then orthogonalise the second column to the first column.
Thus, the centred orthogonalised response matrix is given by
\begin{equation}
    \mathbf{V}_\text{C} = \begin{pmatrix}
    0.7143 & 0.7143 & -0.2857 & -0.2857 & -0.2857 & -0.2857 & -0.2857 \\
    0.0000 & 0.0000 & 0.6000 & 0.6000 & -0.4000 & -0.4000 & -0.4000 \\
    \end{pmatrix}^\top\text{.}
\end{equation}
Third, centre $\mathbf{X}$ into $\mathbf{X}_\text{C}$.

\textbf{Step 2.} As no feature has been selected, $\mathbf{X}_\text{s}$ is empty and $\mathbf{X}_\text{r}$ is the same as $\mathbf{X}$.
Correspondingly, $\mathbf{X}_\text{Cs}$ is empty and $\mathbf{X}_\text{Cr}$ is the same as $\mathbf{X}_\text{C}$.

\textbf{Step 3.} In this step, the centred features in $\mathbf{X}_\text{Cr}$ are required to be orthogonalised to $\mathbf{W}_\text{Cs}$.
As no feature has been selected, let $\mathbf{W}_\text{Cr}$ equal to $\mathbf{X}_\text{Cr}$.

\textbf{Step 4.} The multiple correlation coefficients between $\mathbf{w}_{\text{Cr}i}$ and $\mathbf{V}_\text{C}$ are 0.7628, 0.2264, 0.9779, and 0.9604.

\textbf{Step 5.} The third feature (i.e. petal length) has the highest multiple correlation.
Thus, the petal length is selected into $\mathbf{X}_\text{s}$, and the features contained in $\mathbf{X}_\text{r}$ in order are sepal length, sepal width, and petal width.

\textbf{Step 2.} According to the new $\mathbf{X}_\text{s}$ and $\mathbf{X}_\text{r}$, the centred matrix $\mathbf{X}_\text{C}$ is divided into $(\mathbf{X}_\text{Cs}, \mathbf{X}_\text{Cr})$.

\textbf{Step 3.} As only one feature is in $\mathbf{X}_\text{Cs}$, let the orthogonalised feature $\mathbf{W}_{\text{Cs}}$ equal to $\mathbf{X}_\text{Cs}$.
Through the classical Gram-Schmidt process, the features in $\mathbf{X}_\text{Cr}$ are orthogonalised to $\mathbf{W}_{\text{Cs}}$.

\textbf{Step 4.} The multiple correlation coefficients between $\mathbf{w}_{\text{Cr}i}$ and $\mathbf{V}_\text{C}$ are 0.4458, 0.0841, and 0.4644.

\textbf{Step 5.} The third feature (i.e. petal width) has the highest multiple correlation.
Thus, the features contained in $\mathbf{X}_\text{s}$ in order are petal length and petal width, and the features contained in $\mathbf{X}_\text{r}$ in order are sepal length and sepal width.

\textbf{Step 2.} According to the new $\mathbf{X}_\text{s}$ and $\mathbf{X}_\text{r}$, the centred matrix $\mathbf{X}_\text{C}$ is divided into $(\mathbf{X}_\text{Cs}, \mathbf{X}_\text{Cr})$.

\textbf{Step 3.} Keep the first column of $\mathbf{X}_\text{Cs}$ unchanged, and orthogonalise the second column to the first column through the classical Gram-Schmidt process.
Each feature in $\mathbf{X}_\text{Cr}$ is orthogonalised to $\mathbf{W}_{\text{Cs}}$, respectively, to obtain $\mathbf{W}_{\text{Cr}}$.

\textbf{Step 4.} The multiple correlation coefficients between $\mathbf{w}_{\text{Cr}i}$ and $\mathbf{V}_\text{C}$ are 0.0382 and 0.1108.

\textbf{Step 5.} The second feature (i.e. sepal width), which has the highest multiple correlation, is selected into $\mathbf{X}_\text{s}$.
Therefore, the 3 selected features are petal length, petal width, and sepal width.\bigskip

The squared canonical correlation coefficients between the 3 features and $\mathbf{Y}$ are given by
\begin{equation} \label{eq:iriscca}
\begin{split}
    R_1^2({(\mathbf{x}_3,\mathbf{x}_4,\mathbf{x}_2),\mathbf{Y}}) &= 0.9905 \\
    R_2^2({(\mathbf{x}_3,\mathbf{x}_4,\mathbf{x}_2),\mathbf{Y}}) &= 0.5626\text{.} 
\end{split}
\end{equation}
In LDA, the within-class scatter matrix is given by
\begin{equation}
    \mathbf{S}_w = \begin{pmatrix}
    0.5067  &  0.2367  &  0.2700\\
    0.2367  &  0.1917  &  0.1800\\
    0.2700  &  0.1800  &  0.3050\\
    \end{pmatrix}\text{,}
\end{equation}
and the between-class scatter matrix is given by
\begin{equation}
    \mathbf{S}_b = \begin{pmatrix}
   22.4305 &  10.1333  & -1.1886\\
   10.1333 &   4.6483  & -0.5800\\
   -1.1886 &  -0.5800  &  0.0893\\
    \end{pmatrix}\text{.}
\end{equation}
Through solving the eigenvalue problem \eqref{eq:lda}, the Fisher's criteria of LDA are given by
\begin{equation} \label{eq:irislda}
\begin{split}
    J_1 &= 104.1481 \\
    J_2 &= 1.2864\text{.} \\
\end{split}
\end{equation}
From \eqref{eq:iriscca} and \eqref{eq:irislda}, it can be verified that the relationship between the squared canonical correlation coefficients and Fisher's criterion of LDA is as described by \eqref{eq:jr}.
To verify the equality between the squared canonical correlation coefficients and SOCCs in  \eqref{eq:ccch}, the right hand side of \eqref{eq:ccch} is given by
\begin{equation}
\begin{split}
    R_1^2({(\mathbf{x}_3,\mathbf{x}_4,\mathbf{x}_2),\mathbf{Y}}) + R_2^2({(\mathbf{x}_3,\mathbf{x}_4,\mathbf{x}_2),\mathbf{Y}}) &= 0.9905 + 0.5626 \\
    & = 1.5531\text{,}
\end{split}
\end{equation}
and the left hand side of \eqref{eq:ccch} is given by the sum of the maxima in each iteration, that is $0.9779 + 0.4644 + 0.1108 = 1.5531$.

\subsection{Application to synthetic data for binomial classification}
In this case study, the proposed feature selection method for a binomial classification is investigated.
The $N \times n$ feature matrix is sampled from the multivariate normal distribution, which is given by $\mathbf{X} \sim \mathcal{M}(\bm{\upmu}, \bm{\Upsigma}_\mathcal{N})$, where the mean values in the $n \times 1$ vector $\bm{\upmu}$ are sampled from the normal distribution with mean 0 and standard deviation 0.1.
The $n \times n$ covariance matrix $\bm{\Upsigma}_\mathcal{N}$ is sampled from the Wishart distribution, which is given by $\bm{\Upsigma}_\mathcal{N} \sim \mathcal{W}(\bm{\Upsigma}_\mathcal{W},N)/N$, where $\bm{\Upsigma}_\mathcal{W}$ is a $n \times n$ diagonal matrix whose main diagonal is uniformly distributed on the interval $(0,1)$.
Let the number of the instances is $600$, i.e. $N=600$, and the number of the candidate features are $100$, i.e. $n = 100$.
The $5^\text{th}$, $10^\text{th}$, and $15^\text{th}$ features are used to construct the dummy response vector $\mathbf{y}$, which is sampled from the Bernoulli distribution (i.e. 1 trial binomial distribution) given by $\mathbf{y} \sim \mathcal{B}(\bm{\uppi})$, where the probability vector $\bm{\uppi} = (\pi_1,\ldots,\pi_N)^\top$ is generated by the binomial logistic regression model, that is
\begin{equation}
    \pi_i = \frac{1}{1 + \exp{\left(-(- 2x_{i,5} -3x_{i,10} + 4x_{i,15})\right)}}\text{,}\quad i = 1,\ldots,N \text{.}
\end{equation}
Given $\mathbf{X}$ and $\mathbf{y}$, the aim of the feature selection study is to find the 3 correct feature indices (i.e. 5, 10, and 15).

The proposed OLS based feature selection method is compared with the mutual information based filter methods and embedded methods.
For the mutual information based features selection methods, the continuous features are discretised into 4 categories by the mean values and the mean values $\pm$ the standard deviation.
For the OLS based feature selection, the continuous features are treated in two ways.
One (denoted by OLS) implements the algorithm in subsection \ref{ss:ols} and use continuous features directly.
Another one (denoted by OLSd) implements the algorithm in subsection \ref{ss:olsd}, where the continuous features are discretised into 4 categories by the mean values and the mean values $\pm$ the standard deviation, and then encoded into matrices by $c-1$ dummy encoding.

The simulation study is repeated 100 times to check how many times the feature selection methods choose the 3 correct features, and the results are given by Table \ref{tbl:bfs}.
In this comparison, two OLS based feature selection methods choose the right features 95 times and 88 times, respectively, in the 100 tests, which are higher than what can be achieved by the mutual information based feature selection methods and the embedded methods.

\begin{table}[ht]
\caption{A comparison with mutual information based feature selection methods in binomial classification.}
  \centering
  \scalebox{0.9}{
    \begin{tabular}{ c c c c c c }
    \toprule
    Method & \makecell{Times of selecting\\correct features} & Method & \makecell{Times of selecting\\correct features} & Method & \makecell{Times of selecting\\correct features}\\
    \hline
    OLS     & \textbf{95} & MIM & 73 & ICAP & 74\\
    OLSd    & 88 &  JMI & 79 & DISR & 79 \\
    mRMRd   & 76 &  CMIM & 74 & LASSO & 85\\
    mRMRq   & 77 &  CIFE & 76 & Net & 87\\
    \bottomrule
    \end{tabular}}
  \label{tbl:bfs}
\end{table}

\subsection{Application to synthetic data for multinomial classification}
In this case study, the feature selection for a 3-class multinomial classification is investigated.
The $N \times n$ feature matrix is generated in the same way as in the last subsection.
The number of the instances is $900$, i.e. $N=900$, and the number of the candidate features are $100$, i.e. $n = 100$.
We use the $5^\text{th}$, $10^\text{th}$, and $15^\text{th}$ features to construct the $N \times 3$ response matrix $\mathbf{Y}'$, which is $c$-label dummy encoded.
$\mathbf{Y}'$ is sampled from the categorical distribution (i.e. 1 trial multinomial distribution) given by $\mathbf{Y}' \sim \mathcal{C}(\bm{\Uppi})$, where the $N \times 3$ probability matrix $\bm{\Uppi} = (\pi_{i,j})$ is composed of the probability vector for each class, that is $\bm{\Uppi} = \left(\bm{\uppi}_1, \bm{\uppi}_2, \bm{\uppi}_3\right)$.
The probability vectors are generated by the multinomial logistic regression model \cite[p.~270]{hosmer2013applied}.
First, the probability ratios are given by
\begin{equation} \label{eq:ratiomlr}
\begin{split}
    \frac{\pi_{i,1}}{\pi_{i,3}} &= \exp{(- x_{i,5} -x_{i,10} + x_{i,15})} \\
    \frac{\pi_{i,2}}{\pi_{i,3}} &= \exp{(x_{i,5} -x_{i,10} - x_{i,15})}\text{,}\quad i = 1,\ldots,N\text{.}
\end{split}
\end{equation}
Second, the probability of $\bm{\uppi}_3$ is given by
\begin{equation} \label{eq:pi3}
    \pi_{i,3} = \frac{1}{1 + \frac{\pi_{i,1}}{\pi_{i,3}} + \frac{\pi_{i,2}}{\pi_{i,3}}}\text{,}\quad i = 1,\ldots,N\text{.}
\end{equation}
Finally, $\bm{\uppi}_1$ and $\bm{\uppi}_2$ can be computed by substituting \eqref{eq:pi3} into \eqref{eq:ratiomlr}.
To make the response matrix become $c-1$-label dummy encoded, the first column of $\mathbf{Y}'$ is removed to form $\mathbf{Y}$, which becomes a $N \times 2$ matrix containing only 0 and 1.
Given $\mathbf{X}$ and $\mathbf{Y}$, the aim of the feature selection simulation is to find the 3 correct feature indices (i.e. 5, 10, and 15).

The task is repeated 100 times, and the number of times when a correct feature selection is achieved is shown in Table \ref{tbl:mfs}.
Two OLS based methods still give the best results, especially OLS which uses the continuous features.

\begin{table}[ht]
\caption{A comparison with mutual information based feature selection methods in multinomial classification.}
  \centering
  \scalebox{0.9}{
    \begin{tabular}{ c c c c c c }
    \toprule
    Method & \makecell{Times of selecting\\correct features} & Method & \makecell{Times of selecting\\correct features} & Method & \makecell{Times of selecting\\correct features}\\
    \hline
    OLS     & \textbf{92} & MIM & 82 & ICAP & 82\\
    OLSd    & 84 &  JMI & 80 & DISR & 80 \\
    mRMRd   & 83 &  CMIM & 82 & LASSO & 79\\
    mRMRq   & 84 &  CIFE & 67 & Net & 79\\
    \bottomrule
    \end{tabular}}
  \label{tbl:mfs}
\end{table}

\subsection{Application to the datasets of NIPS feature selection challenge}
Two datasets from the NIPS feature selection challenge \footnote{\url{https://competitions.codalab.org/competitions/3931}} are used for the feature selection methods evaluation.
The detail of the datasets are illustrated in Table \ref{tbl:nips}.
Dexter dataset is from Reuters text categorisation task and Gisette dataset is from a handwriting recoginition task.
Both of the datasets have 2 classes.
The features of the datasets are composed of real features and artificial features (called probes).
As the probes do not carry information of the class labels, the desirable feature selection methods should avoid selecting them.
The datasets are divided into training, validation, and test data.
The labels of the test data are withheld by the data providers, and the performance on the test data are obtained by uploading the results to the challenge website.

\begin{table}[ht]
\caption{Summary of the NIPS feature selection challenge datasets.}
  \centering
  \scalebox{0.9}{
    \begin{tabular}{ c c c }
    \toprule
    Name & Feature (Real/Probe) & Train/Validation/Test \\
    \hline
    Dexter  & 20000 (9947/10053) & 300/300/2000 \\
    Gisette & 5000 (2500/2500)  & 6000/1000/6500 \\
    \bottomrule
    \end{tabular}}
  \label{tbl:nips}
\end{table}

The feature values in both datasets are quantised to 1000 levels, and the features are treated as continuous.
For the mutual information based methods, the 1000 levels are discretised into 10 equal width bins.
For OLSd, the discretised features are encoded into matrices by $c-1$ dummy encoding.
For OLS, the continuous features are used directly.

\begin{figure}[ht]
\centering
\begin{subfigure}[b]{0.4\textwidth}
\centering
\includegraphics[width=1\linewidth]{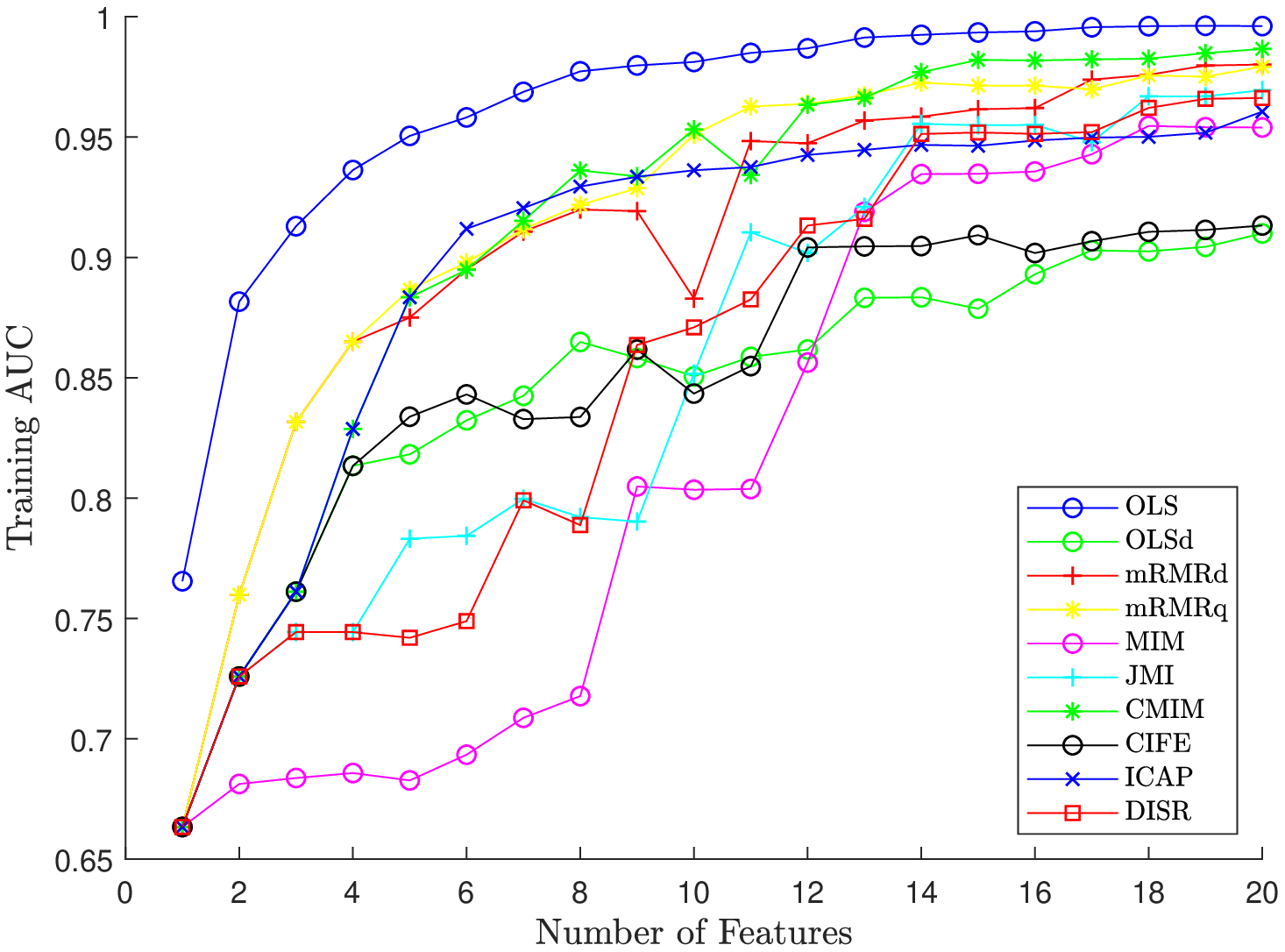}
\caption{}
\label{fig:dexter_a}
\end{subfigure}
\hspace{1cm}
\begin{subfigure}[b]{0.4\textwidth}
\centering
\includegraphics[width=1\linewidth]{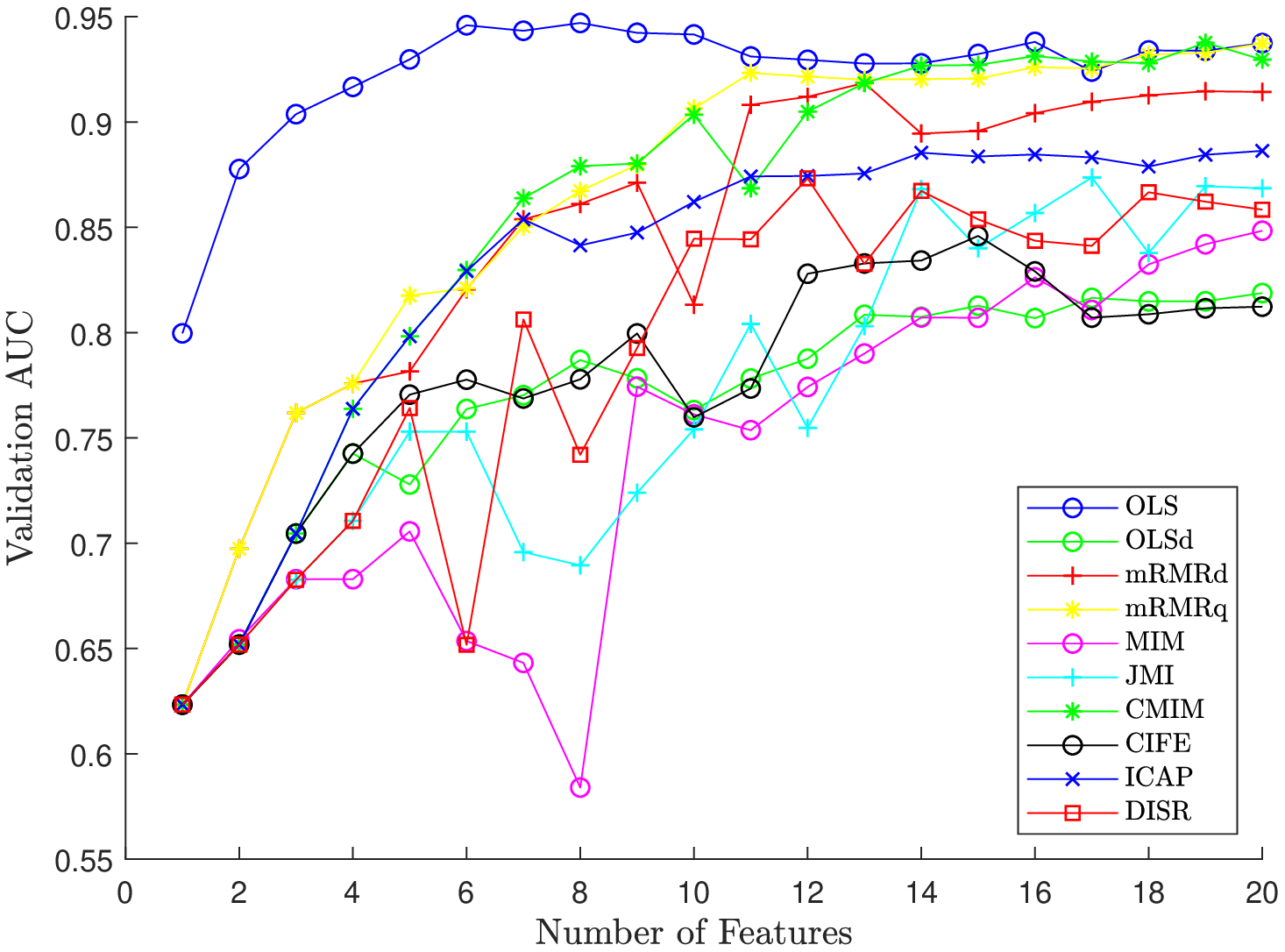}
\caption{}
\label{fig:dexter_b}
\end{subfigure}
\caption{AUC results of the feature selection methods on (a) training and (b) validation Dexter dataset.}
\label{fig:dexter}
\end{figure}

\begin{figure}[ht]
\centering
\begin{subfigure}[b]{0.4\textwidth}
\centering
\includegraphics[width=1\linewidth]{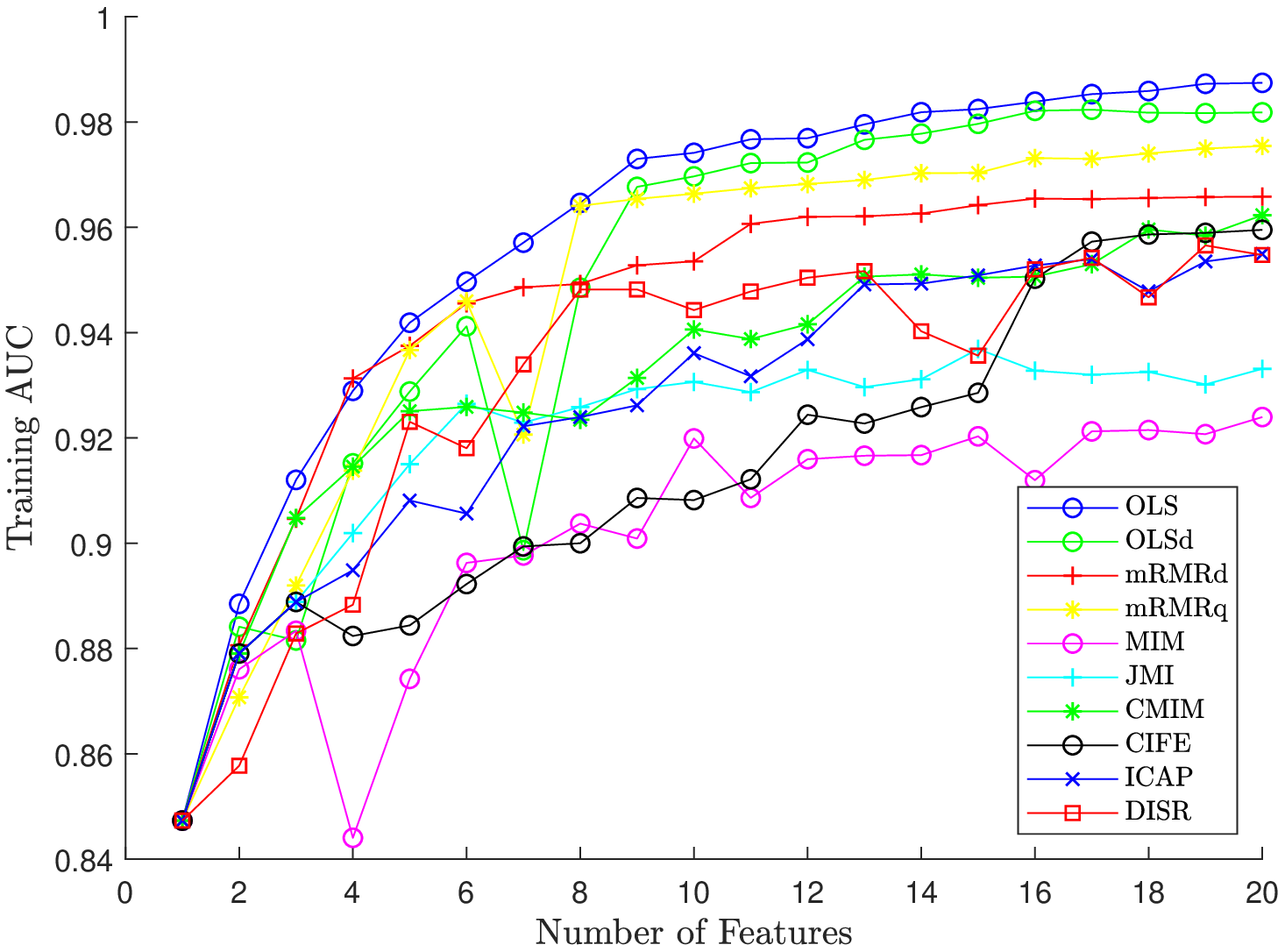}
\caption{}
\label{fig:gisette_a}
\end{subfigure}
\hspace{1cm}
\begin{subfigure}[b]{0.4\textwidth}
\centering
\includegraphics[width=1\linewidth]{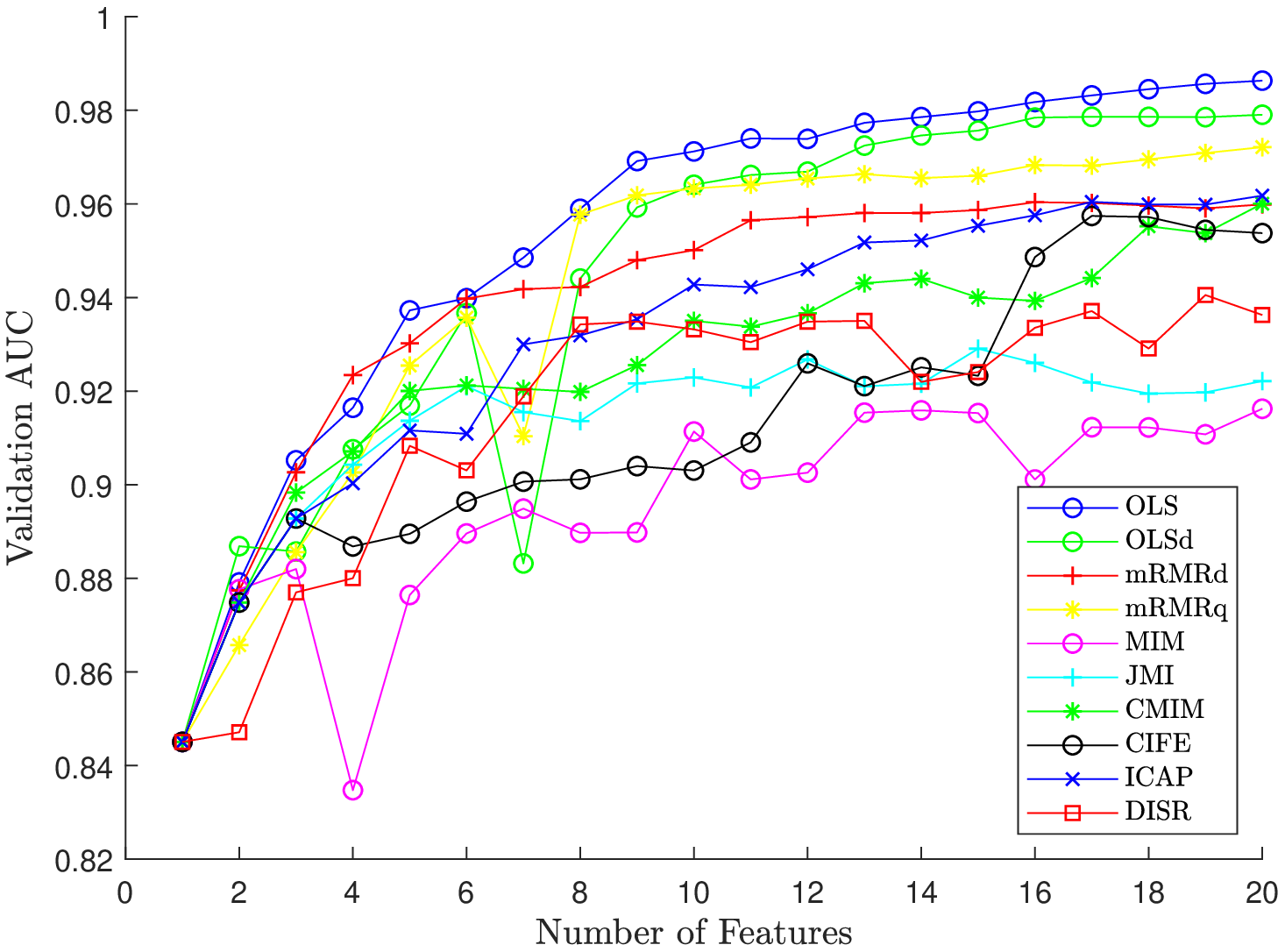}
\caption{}
\label{fig:gisette_b}
\end{subfigure}
\caption{AUC results of the feature selection methods on (a) training and (b) validation Gisette dataset.}
\label{fig:gisette}
\end{figure}

The experiment is implemented in the following steps.
First, the optimal features are selected by different methods using the training data.
Then, given the selected optimal features, a linear Support Vector Machine (SVM) is trained with the training data.
Finally, the prediction results are generated by the SVM model on the training, validation, and test data, respectively.
The classification performance is evaluated by the Area Under the Curve (AUC) of the Receiver Operating Characteristic (ROC) curve.
Each method selects 20 optimal features.
The AUC results on the training and validation data are shown in Fig. \ref{fig:dexter} and Fig. \ref{fig:gisette} for the Dexter and Gisette datasets, respectively.
Generally, OLS which uses the continuous features gives the best classification performance.
In Dexter dataset, OLS shows strikingly better results than other methods.
The results on test data are given in Table \ref{tbl:test}.
Although OLS method selects 1 probe in Dexter dataset, the rest of 19 real features (especially the first 8 features according to Fig. \ref{fig:dexter_b}) selected by OLS are more informative for classification than 20 real features selected by other methods, showing that OLS method can achieve the best AUC results.

\begin{table}[ht]
\caption{Results on the NIPS feature selection challenge test data.}
  \centering
  \scalebox{0.8}{
    \begin{tabular}{ c c c c c c c c c c c c }
    \toprule
    && OLS & OLSd & mRMRd & mRMRq & MIM & JMI & CMIM & CIFE & ICAP & DISR \\
    \hline
    \multirow{2}{*}{Dexter} & AUC & \textbf{0.9551} & 0.8413 & 0.9246 & 0.9355 & 0.8774 & 0.8917 & 0.9444 & 0.8367 & 0.8848 & 0.8908 \\
    & Probe & 1 & 4 & 0 & 1 & 0 & 0 & 1 & 14 & 2 & 0 \\
    \multirow{2}{*}{Gisette} & AUC & \textbf{0.9873} & 0.9824 & 0.9662 & 0.9776 & 0.9324 & 0.9352 & 0.9667 & 0.9605 & 0.9580 & 0.9490 \\
    & Probe & 0 & 0 & 0 & 0 & 0 & 0 & 0 & 0 & 0 & 0 \\
    \bottomrule
    \end{tabular}}
  \label{tbl:test}
\end{table}

\subsection{Application to the MNIST dataset of handwritten digit}
The MNIST dataset \footnote{\url{http://yann.lecun.com/exdb/mnist/}} has a training set of 60000 instances and a test set of 10000 instances, which are categorised into 10 classes (i.e. 0 to 9 digits).
The data is balanced and each class has around 6000 instances for training set and 1000 for test set.
The instance is a $28 \times 28$ pixel box containing grey levels ranging from 0 to 255.
The 784 pixels are features for the classification.
The aim of the feature selection is to choose 50 optimal features for a LDA classifier.
The performance of the feature selection methods is evaluated by the classification accuracy (ACC).

\begin{figure}[ht]
\centering
\begin{subfigure}[b]{0.4\textwidth}
\centering
\includegraphics[width=1\linewidth]{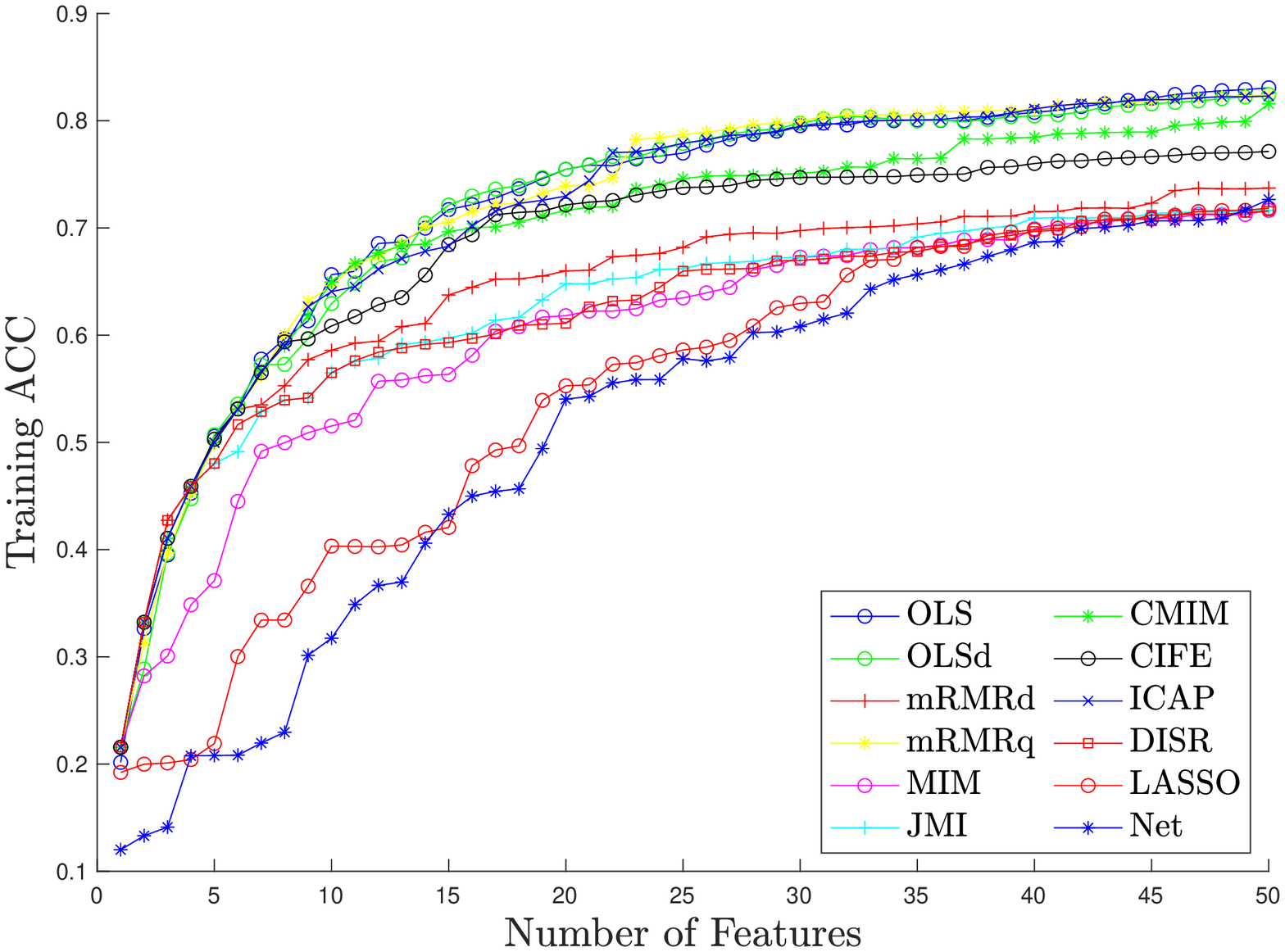}
\caption{}
\label{fig:mnist_a}
\end{subfigure}
\hspace{1cm}
\begin{subfigure}[b]{0.4\textwidth}
\centering
\includegraphics[width=1\linewidth]{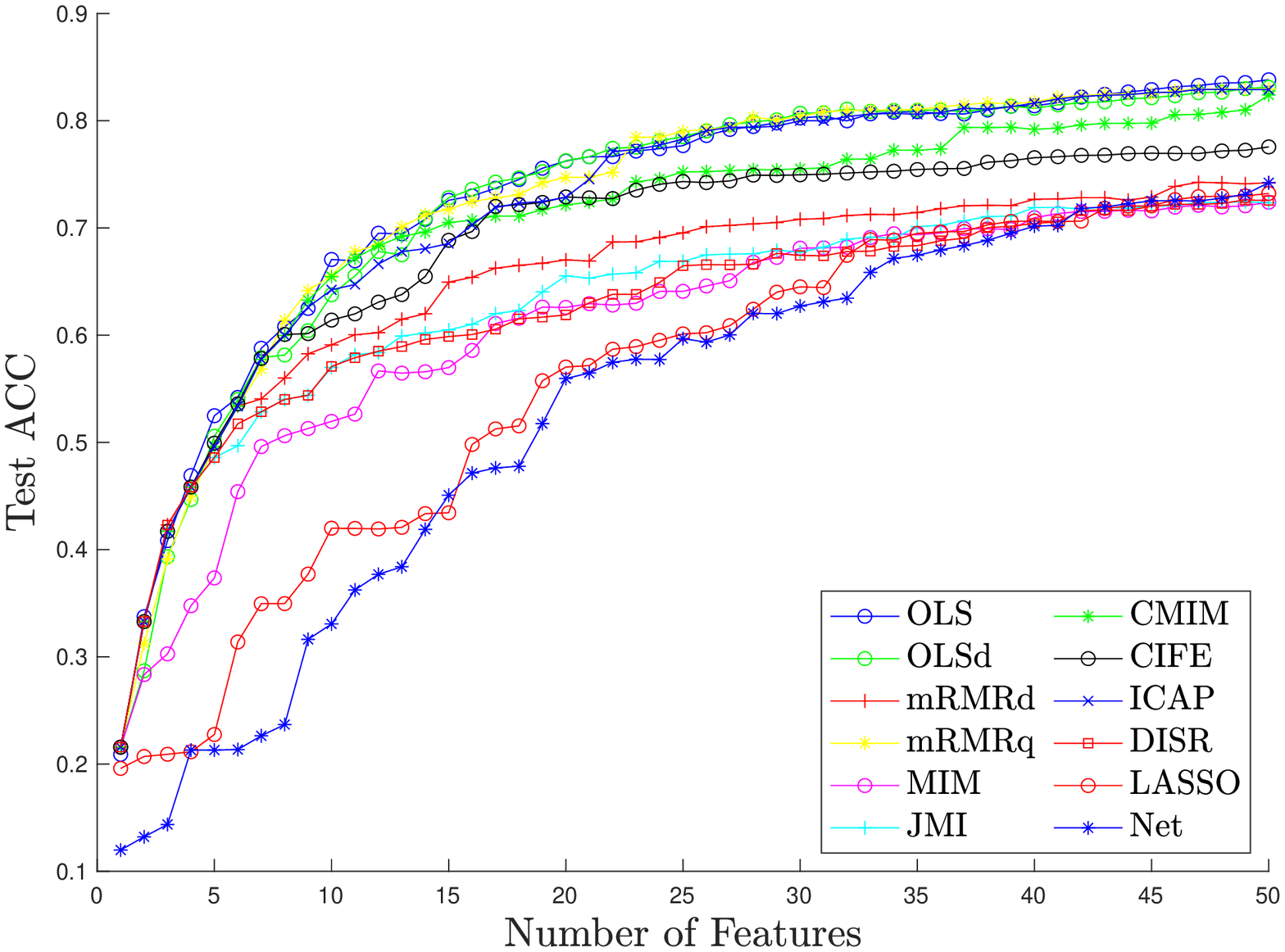}
\caption{}
\label{fig:mnist_b}
\end{subfigure}
\caption{ACC results of the feature selection methods on (a) training and (b) test MNIST dataset.}
\label{fig:mnist}
\end{figure}

For the mutual information based methods and the OLSd method, the continuous grey level features are discretised into 0 and 1 using the method given in \cite{salakhutdinov2008quantitative}.
The OLS method uses the continuous feature directly.
The hyperparameters of the embedded methods are optimised by the grid search.
The feature selection and LDA classifier training are carried out in the training datasets, and tested in the test dataset.
The results of the ACC for the training dataset and the test dataset are shown in Fig. \ref{fig:mnist}.
The performance of the feature selection methods is close at the beginning, where only a few features are selected.
The ACC of training and test datasets starts divided into two groups after more than 14 features are selected.
The OLS, OLSd, mRMRq, CMIM, ICAP, and CIFE methods are in the first tier and the rest of the methods are in the second tier.
When 50 optimal features are selected, the OLS method is in the first place and the OLSd method is in the second place for both the training and test sets.

\begin{figure}[ht]
\centering
\includegraphics[width=1\linewidth]{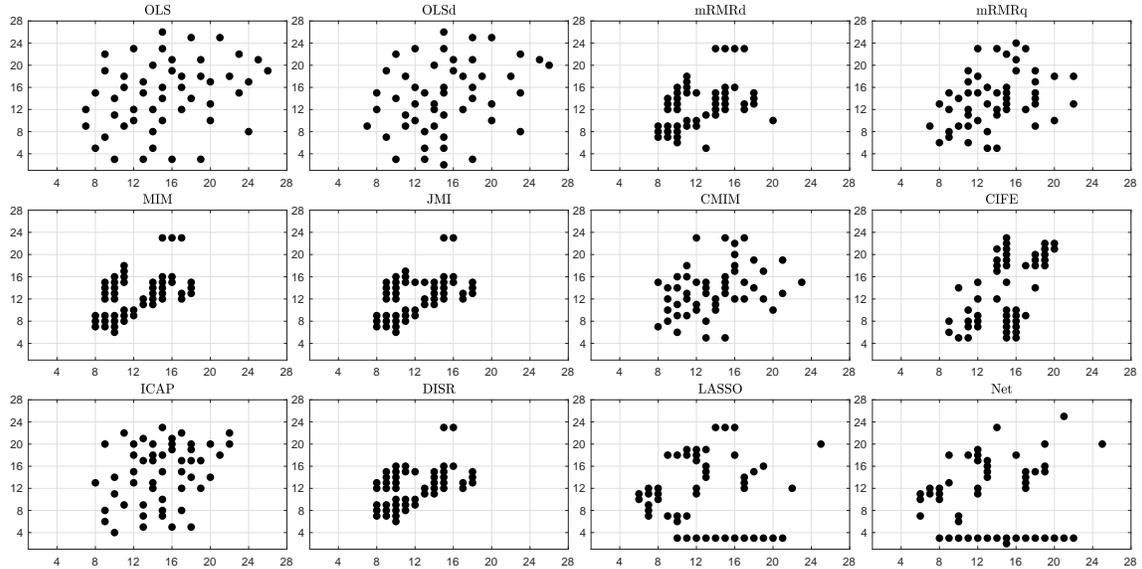}
\caption{The comparison of the selected pixels when using different feature ranking criteria.}
\label{fig:pixel}
\end{figure}

The 50 pixels selected by each feature selection method are compared in Fig. \ref{fig:pixel}.
All the feature selection criteria can indicate the pixels in the centre area is relevant to the digit classification, while the pixels in the corners are irrelevant.
However, it is found that some mutual information based methods and the embedded methods tend to select the blocks of neighbouring pixels, while the proposed methods tend to select the pixels more spread over the entire picture.
As the neighbouring pixels normally contain the redundant information, the criteria without redundancy control, e.g. MIM \cite{brown2012conditional}, will improperly select the relevant but redundant pixels, which are as shown in Fig. \ref{fig:pixel}.
Therefore, the results imply that, in this case, the proposed OLS methods provides best redundancy control.

\subsection{Application to the UCI datasets}
The four real world datasets are from the UCI machine learning repository \footnote{\url{https://archive.ics.uci.edu/ml/datasets.php}} \cite{Dua:2019}.
The detailed information of the datasets are shown in Table \ref{tbl:uci}.

\begin{table}[ht]
\caption{Summary of the UCI datasets.}
  \centering
  \scalebox{0.9}{
    \begin{tabular}{ c c c c c }
    \toprule
    Name & Mfeat & Breast & CNAE & Lymph \\
    \hline
    Data Type  & Continuous & Continuous & Discrete & Categorical/Discrete  \\
    No. of instances & 2000  & 569 & 1080 & 142 \\
    No. of Features & 649 & 30 & 856 & 18 \\
    No. of Classes & 10 & 2 & 9 & 2 \\
    \bottomrule
    \end{tabular}}
  \label{tbl:uci}
\end{table}

The multi-feature digit (Mfeat) dataset \cite{van1998handwritten} consists of handwritten numerals (0 to 9) extracted from a collection of Dutch utility maps.
Each class has 200 instances, and there are 2000 instances in total.
The instances are originally scanned to produce the images of 8 bits greyscale.
649 continuous features are extracted from the raw images, including 76 Fourier coefficients, 216 profile correlations, 64 Karhunen-Lo\`eve coefficients, 240 pixel averages in $2 \times 3$ windows, 47 Zernike moments, and 6 morphological features.

The Wisconsin diagnostic breast cancer (Breast) dataset \cite{street1993nuclear} has 569 instances are classified into malignant (212) or benign (357) breast cancer.
The original data are from 10 measurements, including the cell nucleus' radius, texture, perimeter, etc.
The 30 features are constructed by taking the mean, standard error, and largest (mean of the three largest values) of these measurements.

CNAE-9 (CNAE) dataset \cite{ciarelli2009agglomeration} contains 1080 documents of business descriptions of Brazilian companies, which are categorised into 9 economic activities.
The dataset is balanced, and each activity has 120 documents.
The frequencies of the 856 words in the document are used as features.
The feature matrix are highly sparse and 99.22\% of the matrix is filled with zeros.

Lymphography (Lymph) dataset \cite{michalski1986multi} contains 148 instances in 4 classes, i.e. normal find (2), metastases (81), malign lymph (61), and fibrosis (4).
To make the dataset balanced, only the instances belong to metastases and malign lymph are used.
The 18 medical diagnostic attributes form categorical and discrete features.

\begin{figure}[ht]
\centering
\begin{subfigure}[b]{0.4\textwidth}
\centering
\includegraphics[width=1\linewidth]{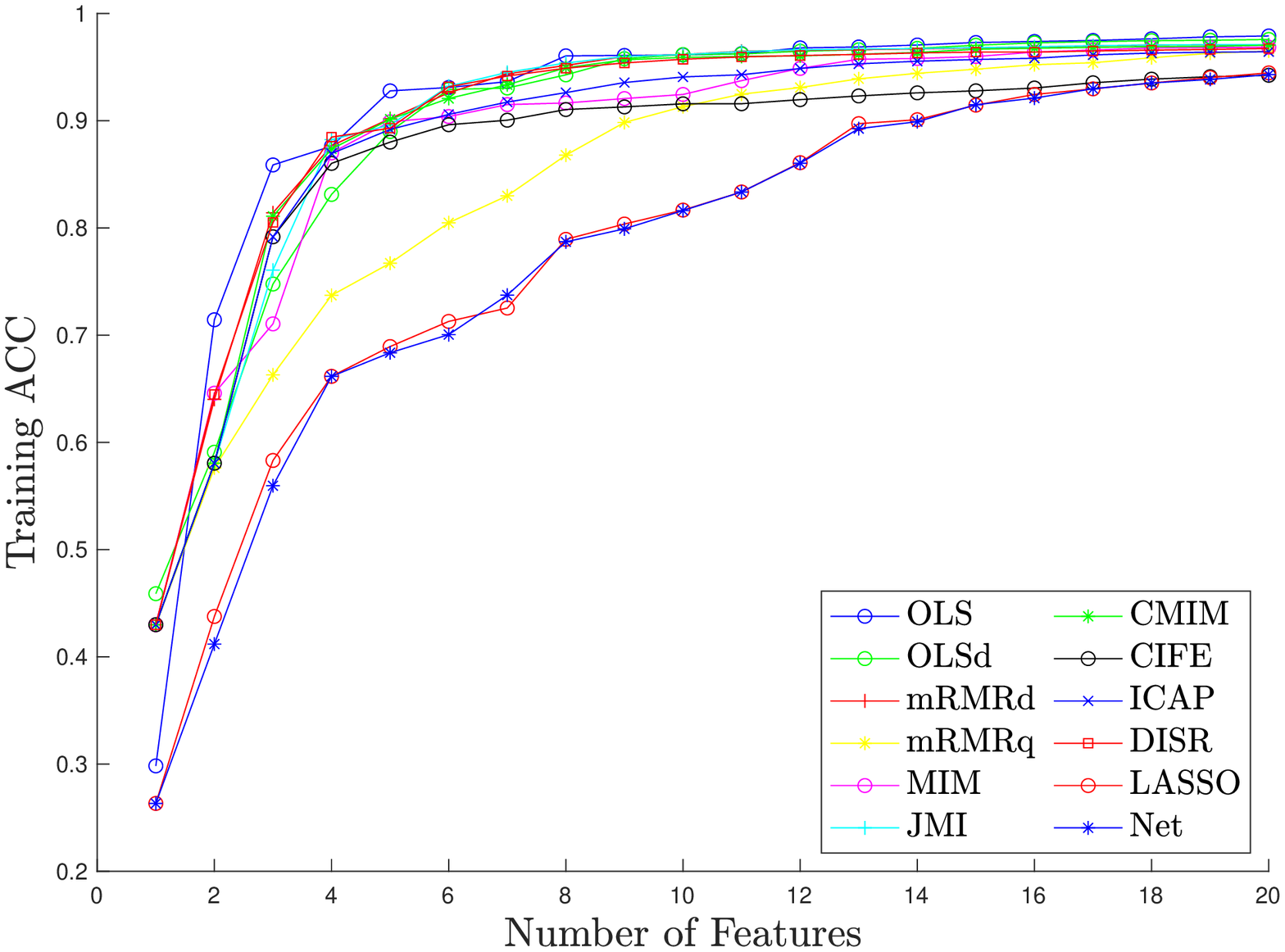}
\caption{}
\label{fig:mfeat_a}
\end{subfigure}
\hspace{1cm}
\begin{subfigure}[b]{0.4\textwidth}
\centering
\includegraphics[width=1\linewidth]{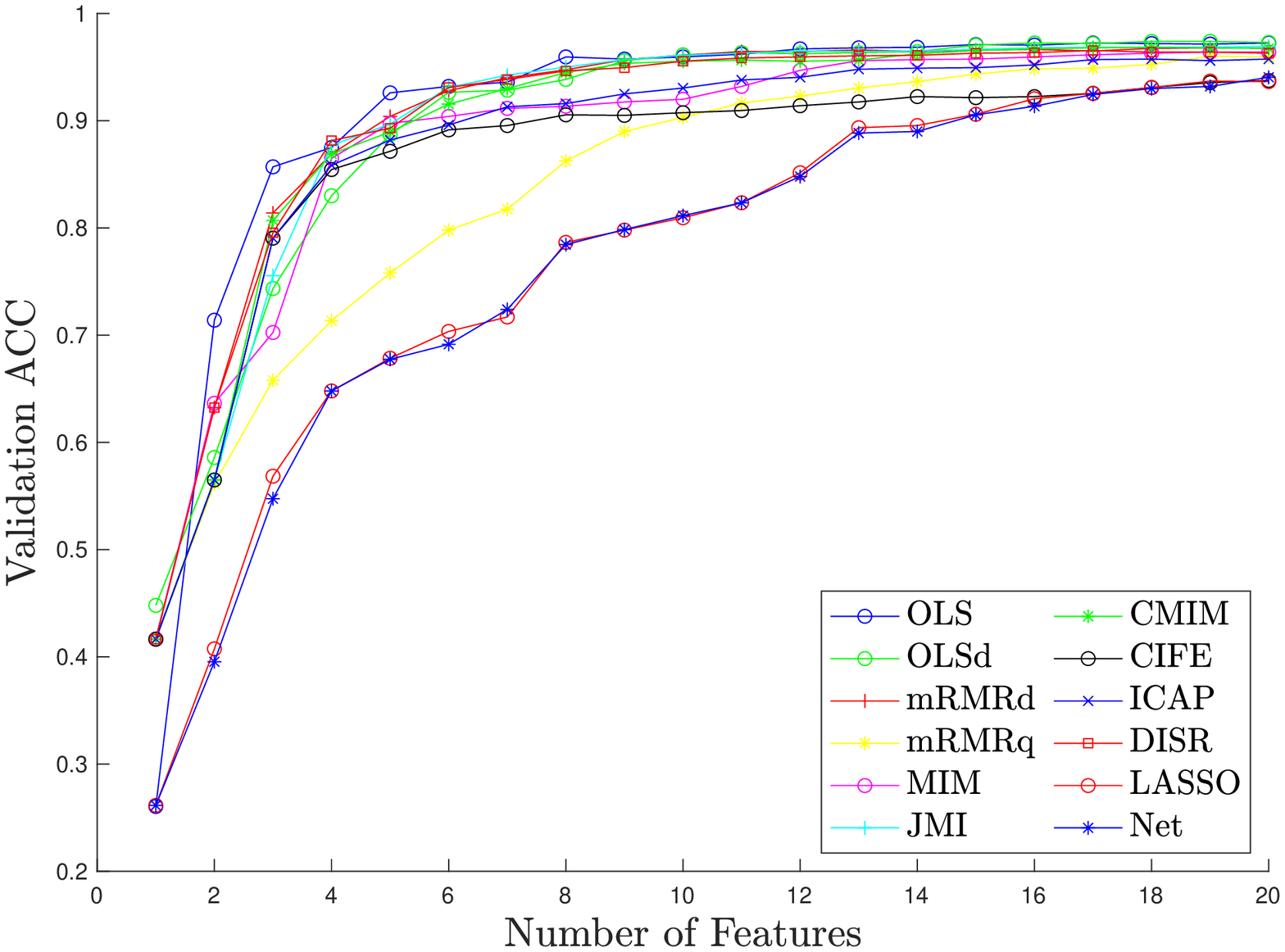}
\caption{}
\label{fig:mfeat_b}
\end{subfigure}
\caption{The average ACC results of the 10-fold cross validation for the feature selection methods on (a) training and (b) validation Mfeat dataset.}
\label{fig:mfeat}
\end{figure}

\begin{figure}[ht]
\centering
\begin{subfigure}[b]{0.4\textwidth}
\centering
\includegraphics[width=1\linewidth]{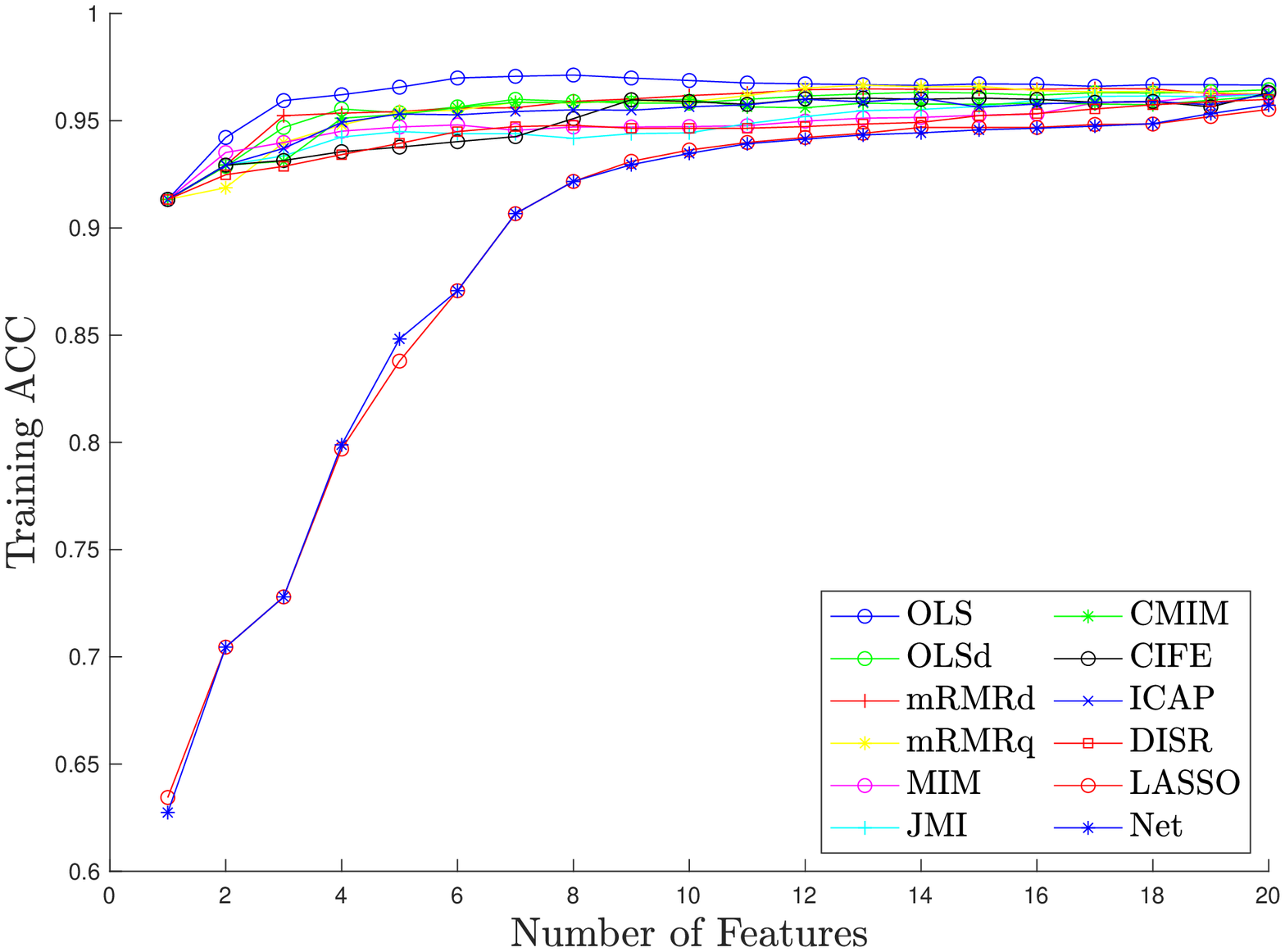}
\caption{}
\label{fig:breast_a}
\end{subfigure}
\hspace{1cm}
\begin{subfigure}[b]{0.4\textwidth}
\centering
\includegraphics[width=1\linewidth]{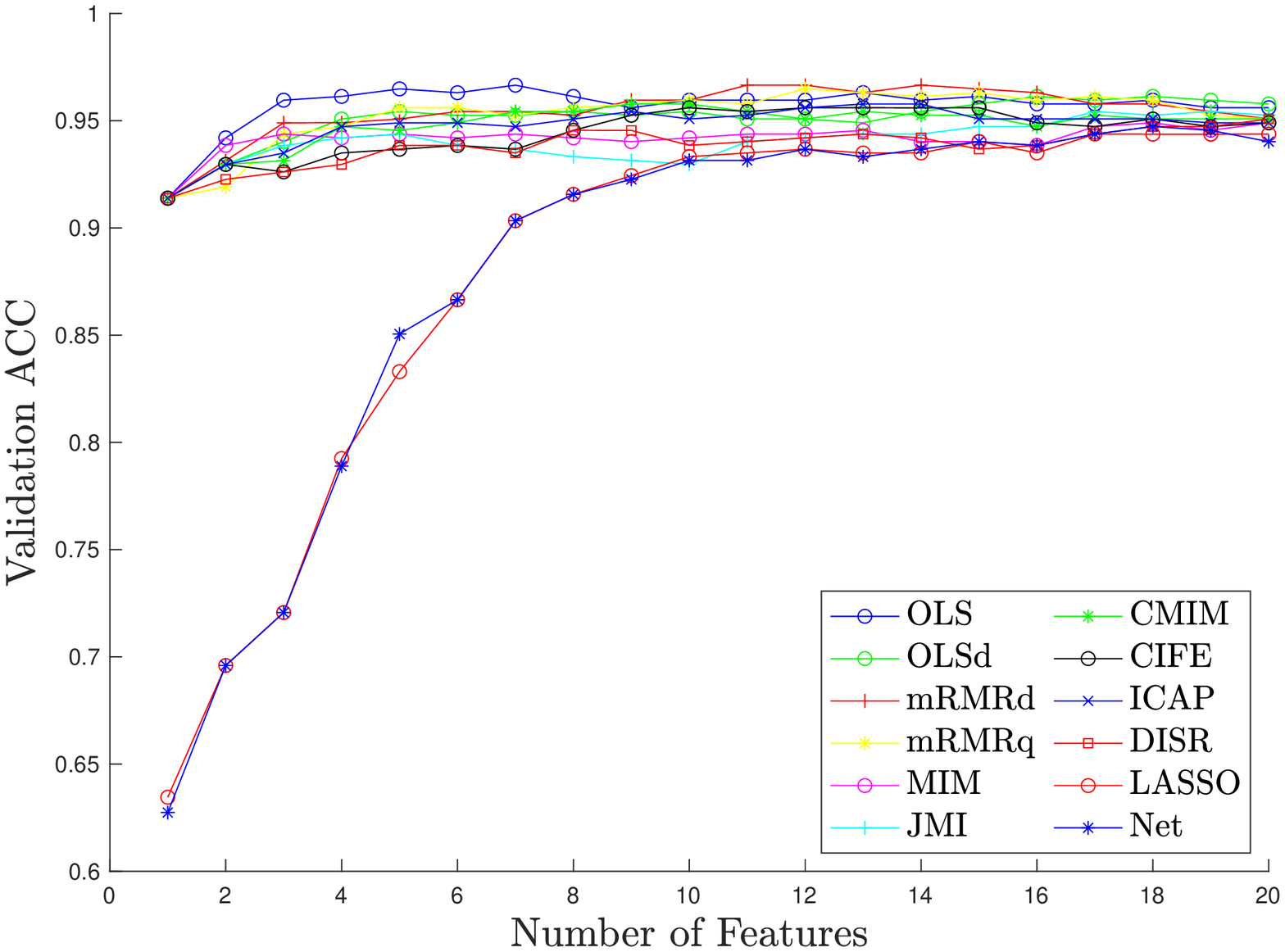}
\caption{}
\label{fig:breast_b}
\end{subfigure}
\caption{The average ACC results of the 10-fold cross validation for the feature selection methods on (a) training and (b) validation Breast dataset.}
\label{fig:breast}
\end{figure}

\begin{figure}[ht]
\centering
\begin{subfigure}[b]{0.4\textwidth}
\centering
\includegraphics[width=1\linewidth]{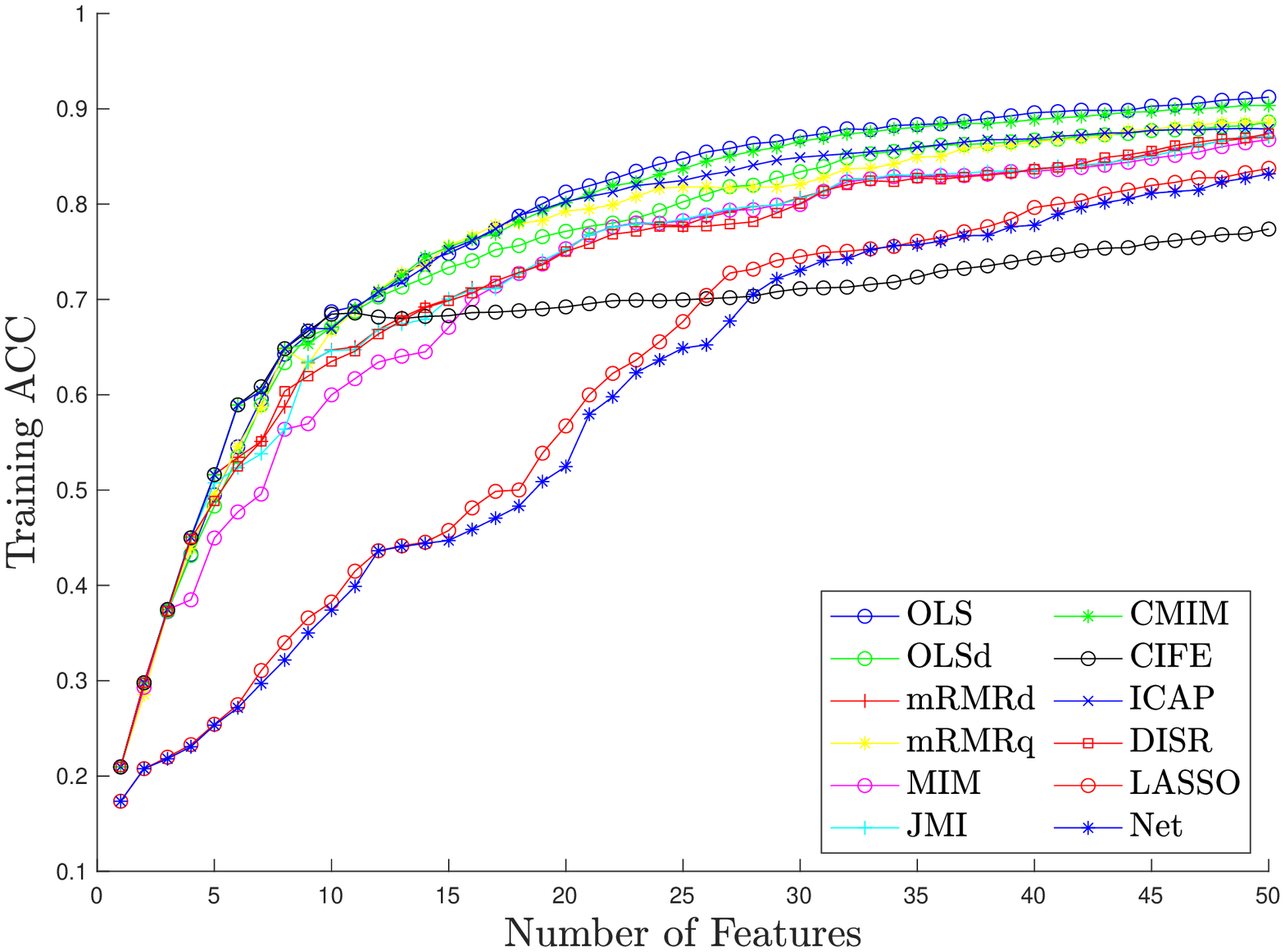}
\caption{}
\label{fig:CNAE_a}
\end{subfigure}
\hspace{1cm}
\begin{subfigure}[b]{0.4\textwidth}
\centering
\includegraphics[width=1\linewidth]{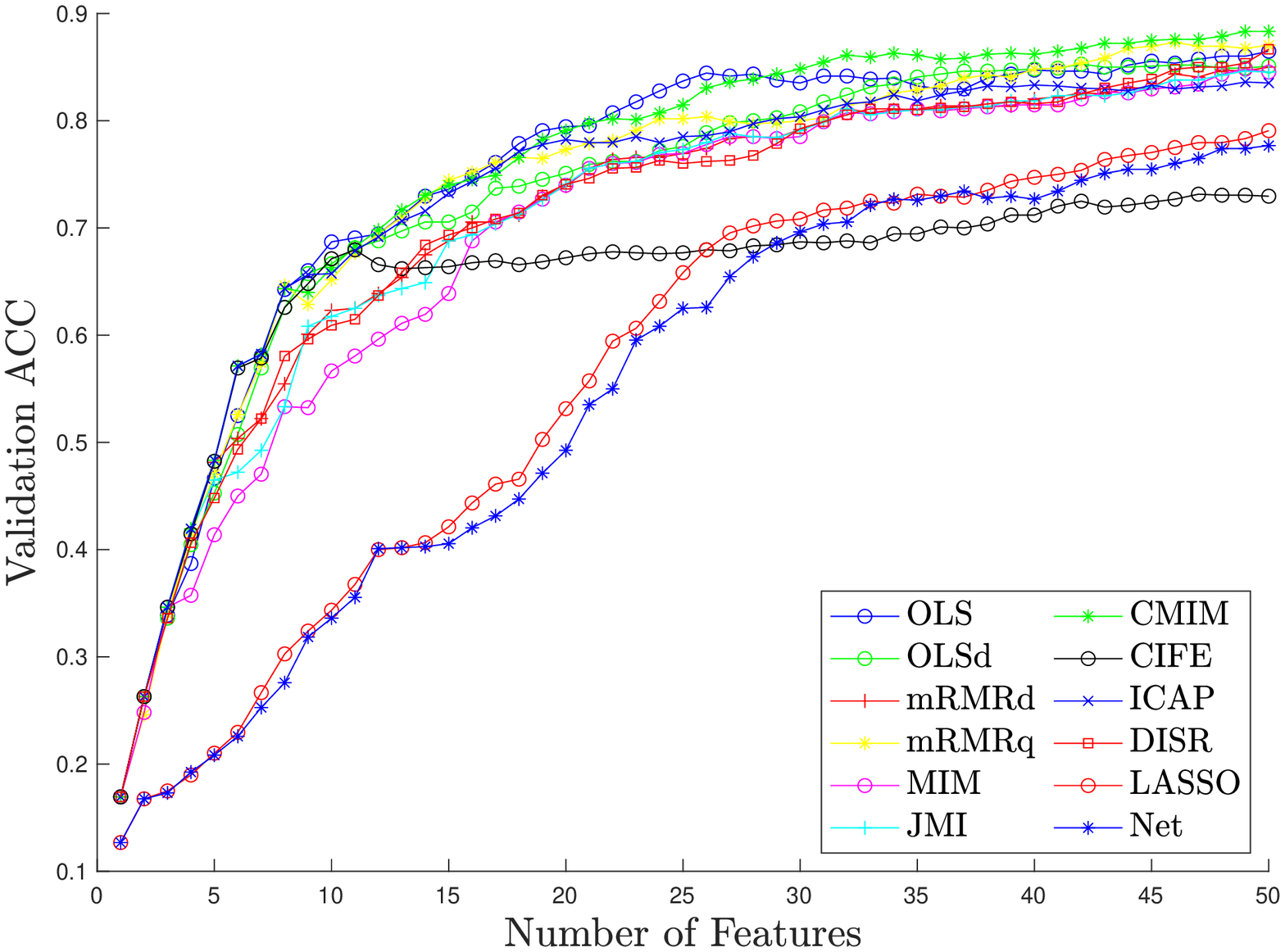}
\caption{}
\label{fig:CNAE_b}
\end{subfigure}
\caption{The average ACC results of the 10-fold cross validation for the feature selection methods on (a) training and (b) validation CNAE dataset.}
\label{fig:CNAE}
\end{figure}

\begin{figure}[ht]
\centering
\begin{subfigure}[b]{0.4\textwidth}
\centering
\includegraphics[width=1\linewidth]{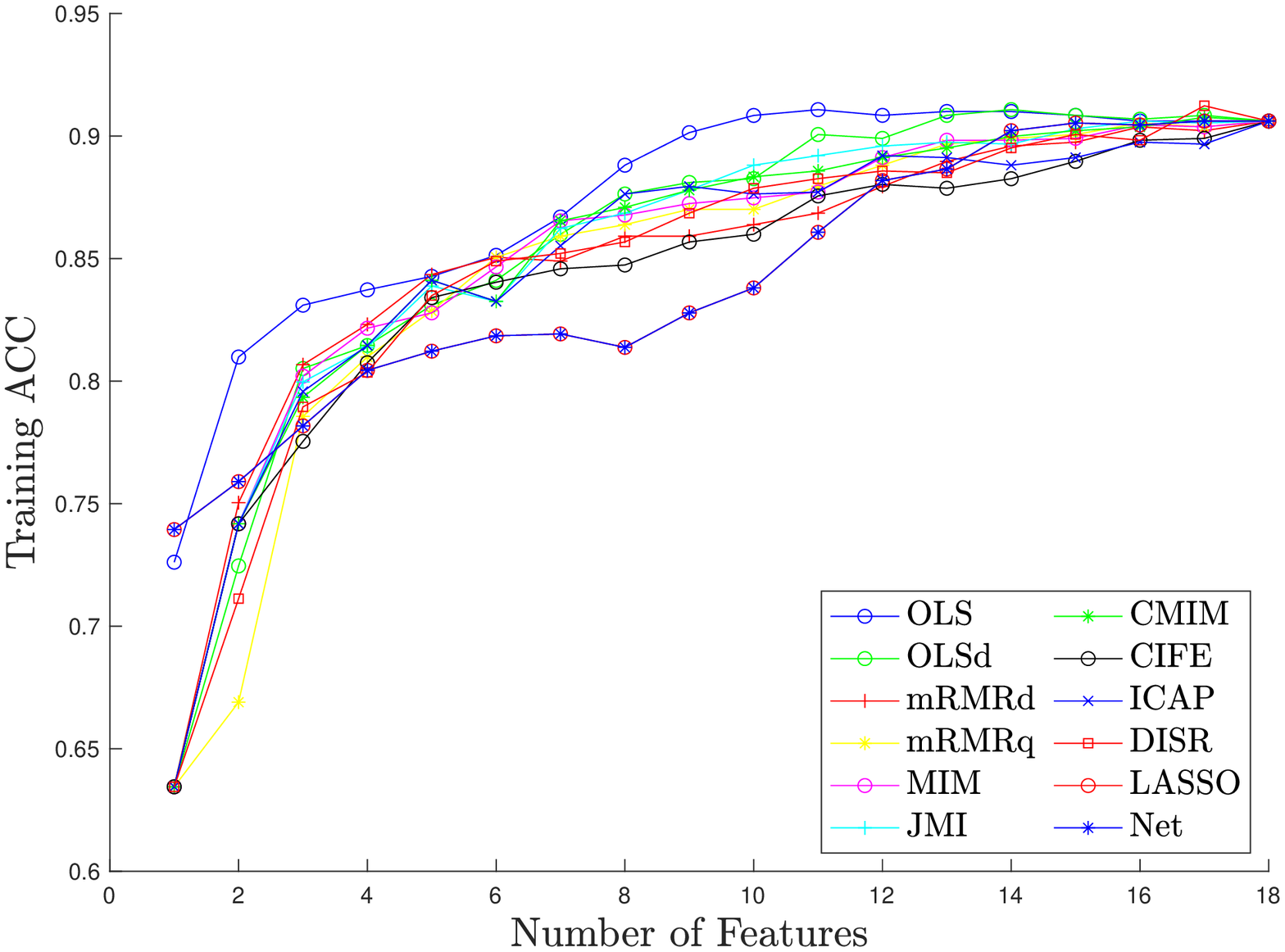}
\caption{}
\label{fig:lymph_a}
\end{subfigure}
\hspace{1cm}
\begin{subfigure}[b]{0.4\textwidth}
\centering
\includegraphics[width=1\linewidth]{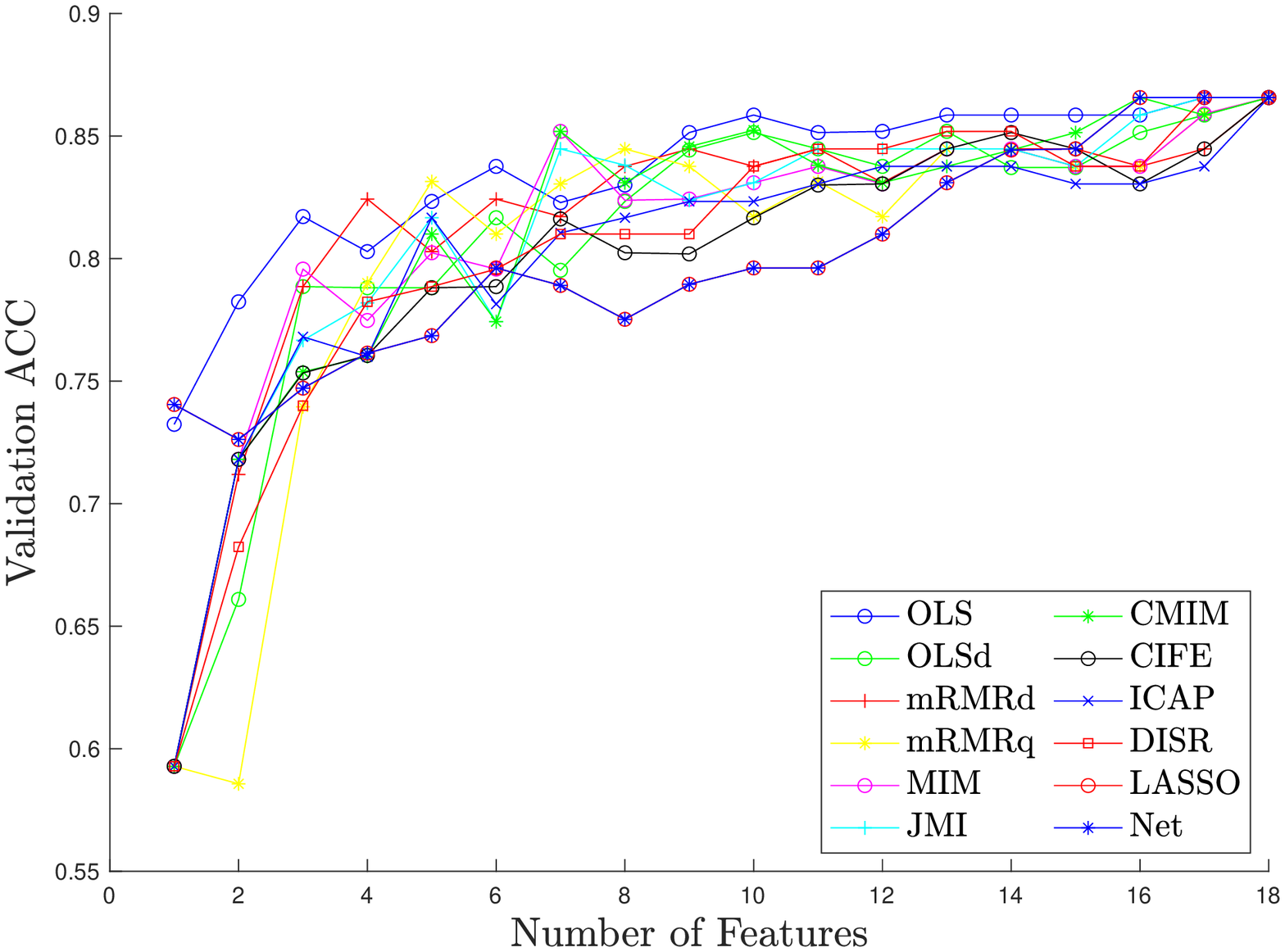}
\caption{}
\label{fig:lymph_b}
\end{subfigure}
\caption{The average ACC results of the 10-fold cross validation for the feature selection methods on (a) training and (b) validation Lymph dataset.}
\label{fig:lymph}
\end{figure}

For the continuous features, the mutual information based feature selection methods use the discretised features which are categorised into 3 categories by the mean values $\pm$ the standard deviations.
The OLSd method uses the dummy features which encode the discretised features into matrices via $c-1$ dummy encoding, while the OLS method uses the continuous features directly.
For the categorical/discrete features, the mutual information based methods can directly use them, and OLSd use $c-1$ dummy encoding features.
The OLS method uses the discrete feature directly, and adopts the categorical features which are encoded into vectors by the ordinal encoding.
The hyperparameters of the LASSO and the elastic net are optimised by the grid search.
The 10-fold cross validation for the ACC of the LDA classifier is applied.
The mean and standard deviation are extracted from the training data for the continuous feature discretisation.
The feature selection and the LDA classifier training are implemented on the 10 training datasets.
The feature selection performance is evaluated by the average ACC on the training and validation datasets, which are shown in Fig. \ref{fig:mfeat} to Fig. \ref{fig:lymph}.

For the Mfeat dataset, except that the mRMRq method gives a significantly worse ACC when the $3^\text{rd}$ to $8^\text{th}$ features are selected, the other methods gives similar results.
When 20 features are selected, the OLS method achieves the highest ACC in the training set and the second highest ACC in the validation set, while the OLSd method achieves the second highest ACC in the training set and the highest ACC in the validation set.

For the Breast dataset, the ACC of the feature selection methods varies in the small range between 0.91 and 0.97.
The OLS gives the best ACC in the training set from the $2^\text{nd}$ to $20^\text{th}$ feature, and the validation set from the $2^\text{nd}$ to $8^\text{th}$ features.

For the CNAE dataset, except that the CIFE method gives a significantly worse ACC, the other mutual information based methods give similar results.
The OLS gives the best ACC in the training set from the $19^\text{th}$ to $50^\text{th}$ features.
When 50 features are selected, the top 5 methods in the test set are CMIM, mRMRq, DISR, OLS, OLSd.

For the Lymph dataset, except that OLS gives a slightly better ACC in the training set when the first 4 features are selected and the $8^\text{th}$ to $12^\text{th}$ features are selected, no method is evidently stronger or weaker than other methods.

It can be seen that no feature selection method ranks first in all four datasets.
However, the proposed OLS and OLSd are generally in the top five methods, and the methods do not show a significant weakness in any of the datasets.
In addition, as no discretisation is required, the proposed OLS method is more convenient than the mutual information based methods when the features are continuous.
The embedded methods generally give worse ACC comparing with the filter methods in the first few selected features.
The reason can be that the filter methods select the most important features first, while the embedded methods select the feature together without ranking them and the features are merely sorted by the absolute values of the features' fitted coefficients.

\subsection{Comparison with the most recent methods}
The two most recent methods are compared with the proposed OLS method.
The ranking criteria of the first method is OBCC \cite{solares2019novel}, which has been briefly described in the introduction.
The second is the Shapley value based feature selection method \cite{lundberg2017unified,fryer2021shapley}.
The Shapley value is firstly introduced by Lloyd S. Shapley as a concept of coalitional game theory \cite{shapley1953value}.
Recently, it has been developed as a unified approach to the feature importance analysis for machine learning models \cite{lundberg2017unified}, which can also be used for feature selection \cite{fryer2021shapley}.
One reason of the popularity of the Shapley value is it has some desirable properties, such as additivity and uniqueness \cite{lundberg2017unified}.
In this paper, the ranking criterion of the Shapley value based features selection method, denoted as SHAP, is the sum of absolute Shapley values for all the training data and classes with a LDA model, which we use \textsc{Matlab} function \texttt{shapley} to compute.
As the computation speed of SHAP is slow for the large dataset, only the small datasets Breast and Lymph in the last subsection and two new small datasets of Flag and Parkinsons are used for this comparison.

The Flag dataset \cite{john1994irrelevant} contains details of 194 nations and their flags, which are classified into 8 religions including Catholic, other Christian, Muslim, Buddhist, Hindu, Ethnic, Marxist, and others.
The information forms 28 categorical and discrete features including population of the nation, number of different colours in the flag, number of vertical bars in the flag, etc, where the categorical features are ordinal encoded.

The Parkinsons dataset \cite{4636708} is composed of 195 biomedical voice measurements from people, where 147 measurements are from people with Parkinson's disease and 48 measurements are from healthy people.
The 22 continuous features including average, maximum and minimum vocal fundamental frequencies, etc.

\begin{figure}[ht]
\centering
\begin{subfigure}[b]{0.4\textwidth}
\centering
\includegraphics[width=1\linewidth]{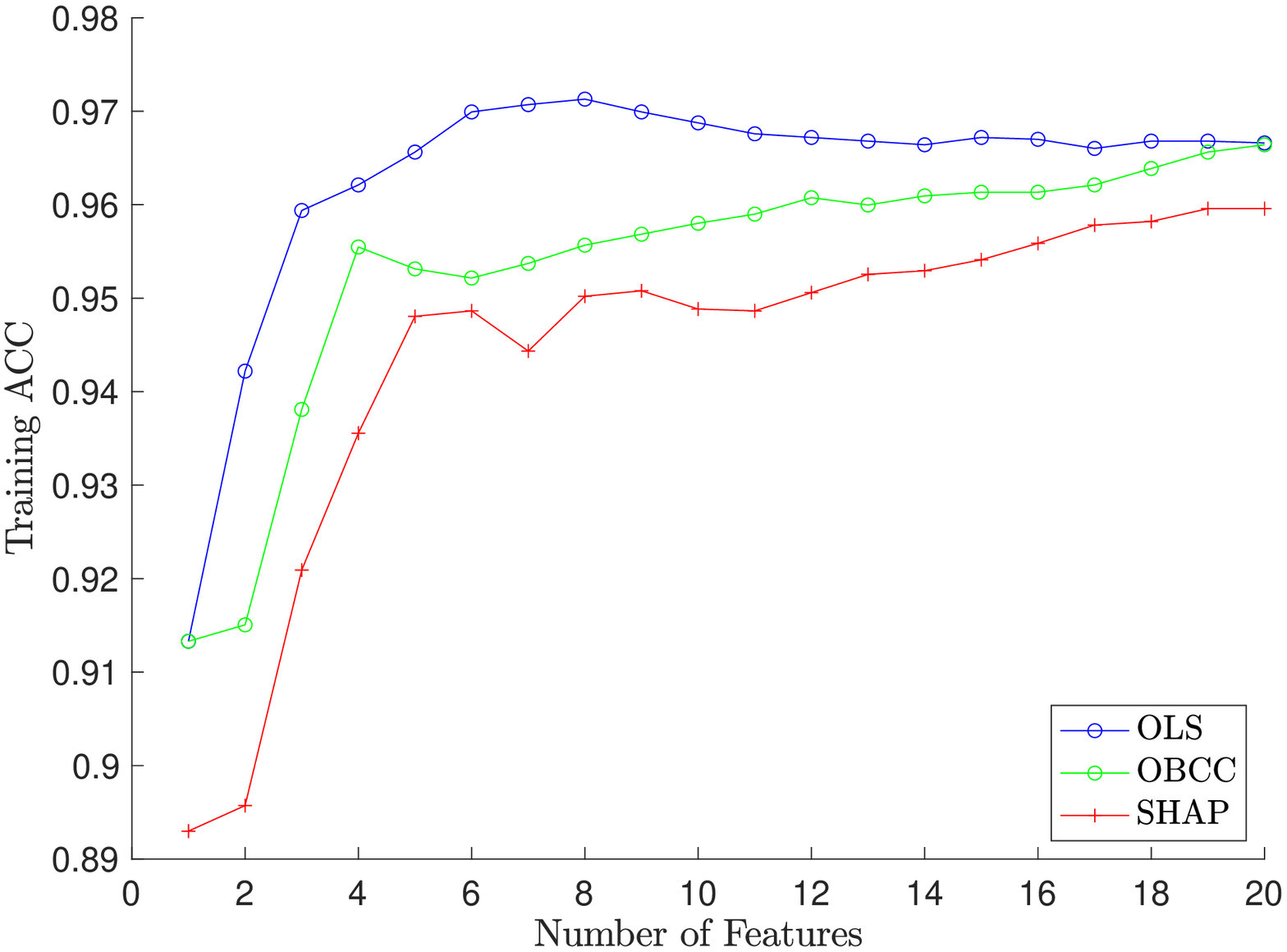}
\caption{}
\label{fig:breast_shap_a}
\end{subfigure}
\hspace{1cm}
\begin{subfigure}[b]{0.4\textwidth}
\centering
\includegraphics[width=1\linewidth]{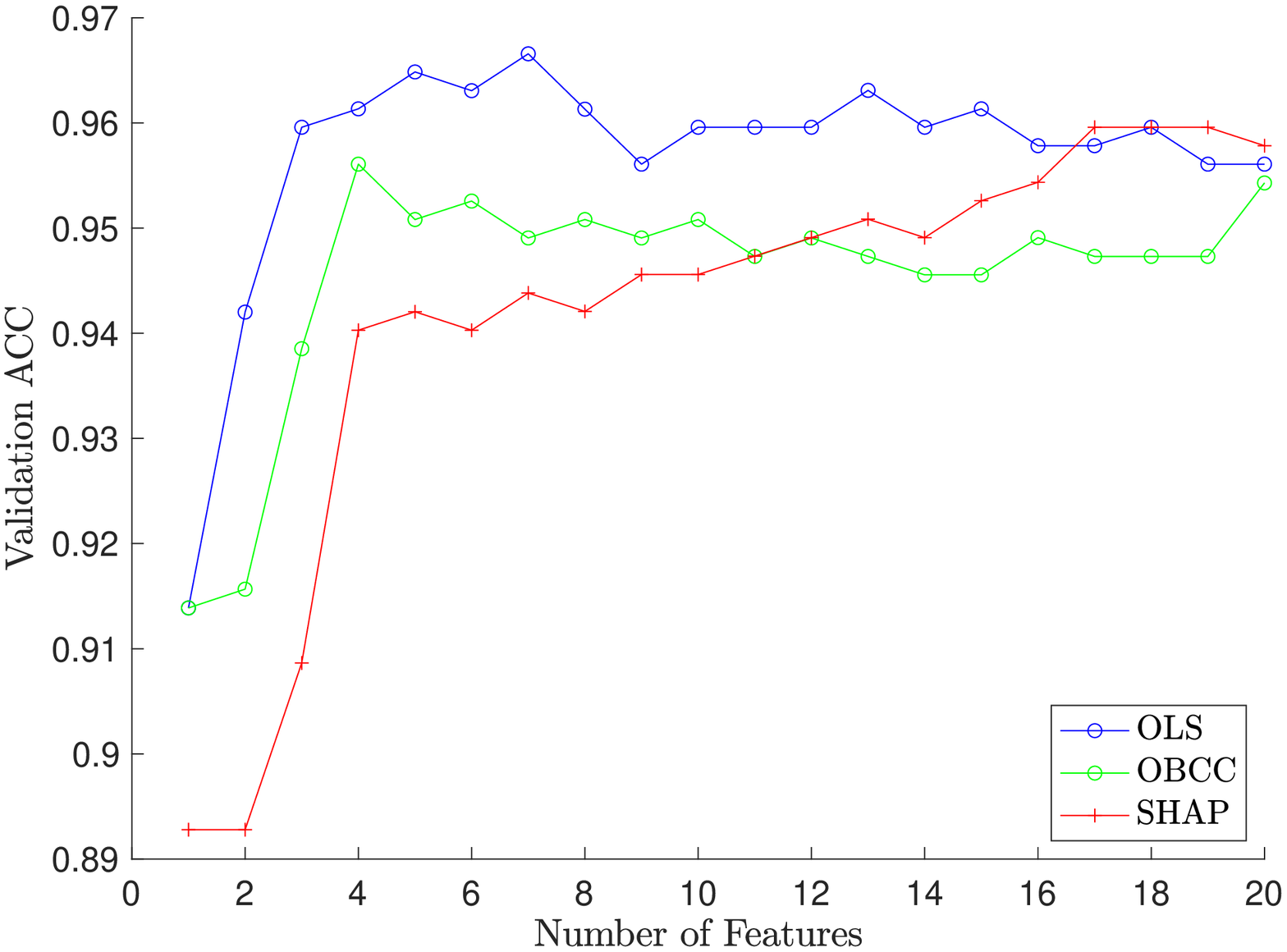}
\caption{}
\label{fig:breast_shap_b}
\end{subfigure}
\caption{The average ACC results of the 10-fold cross validation for the OLS, OBCC and SHAP methods on (a) training and (b) validation Breast dataset.}
\label{fig:breast_shap}
\end{figure}

\begin{figure}[ht]
\centering
\begin{subfigure}[b]{0.4\textwidth}
\centering
\includegraphics[width=1\linewidth]{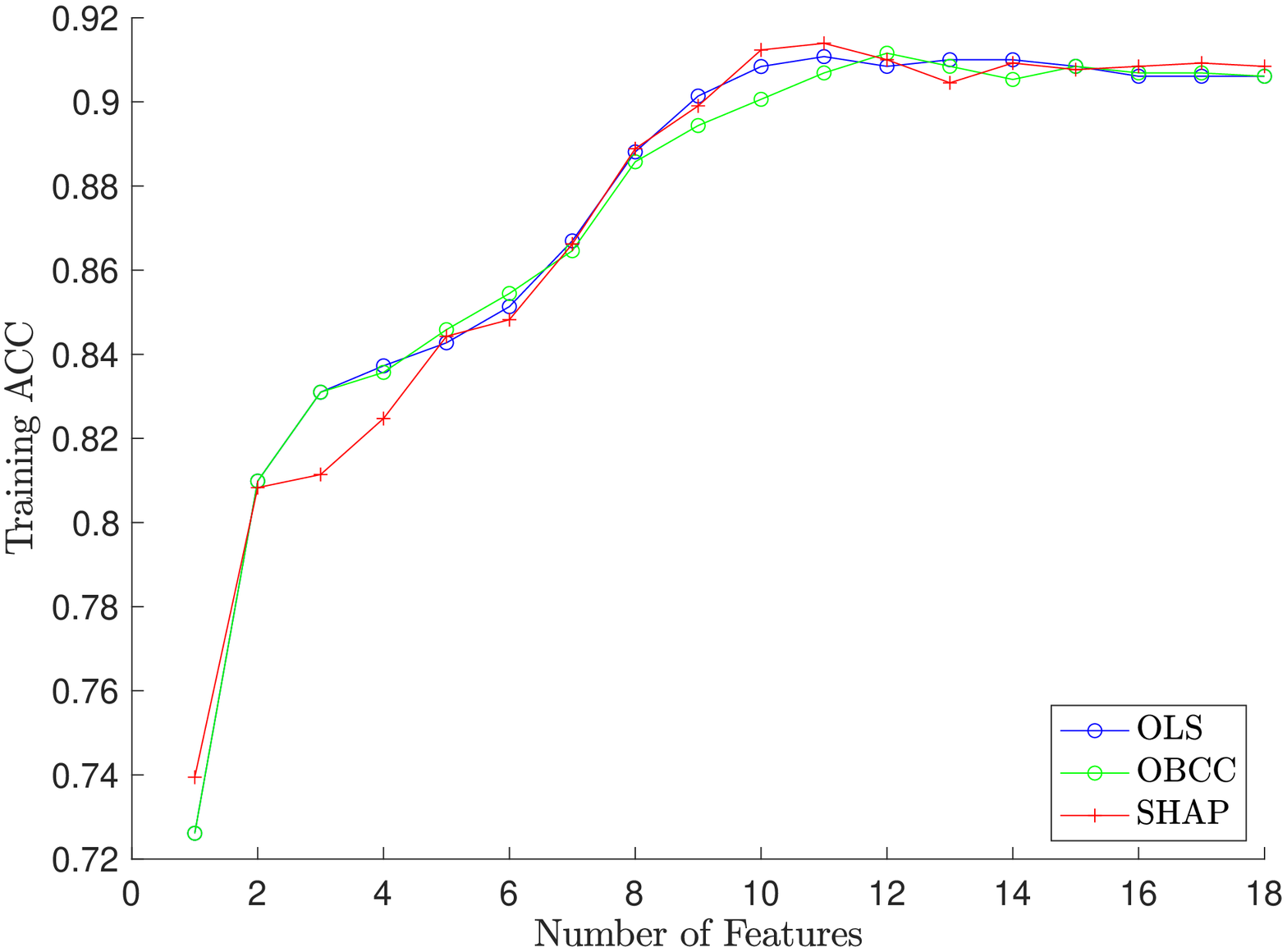}
\caption{}
\label{fig:lymp_shap_a}
\end{subfigure}
\hspace{1cm}
\begin{subfigure}[b]{0.4\textwidth}
\centering
\includegraphics[width=1\linewidth]{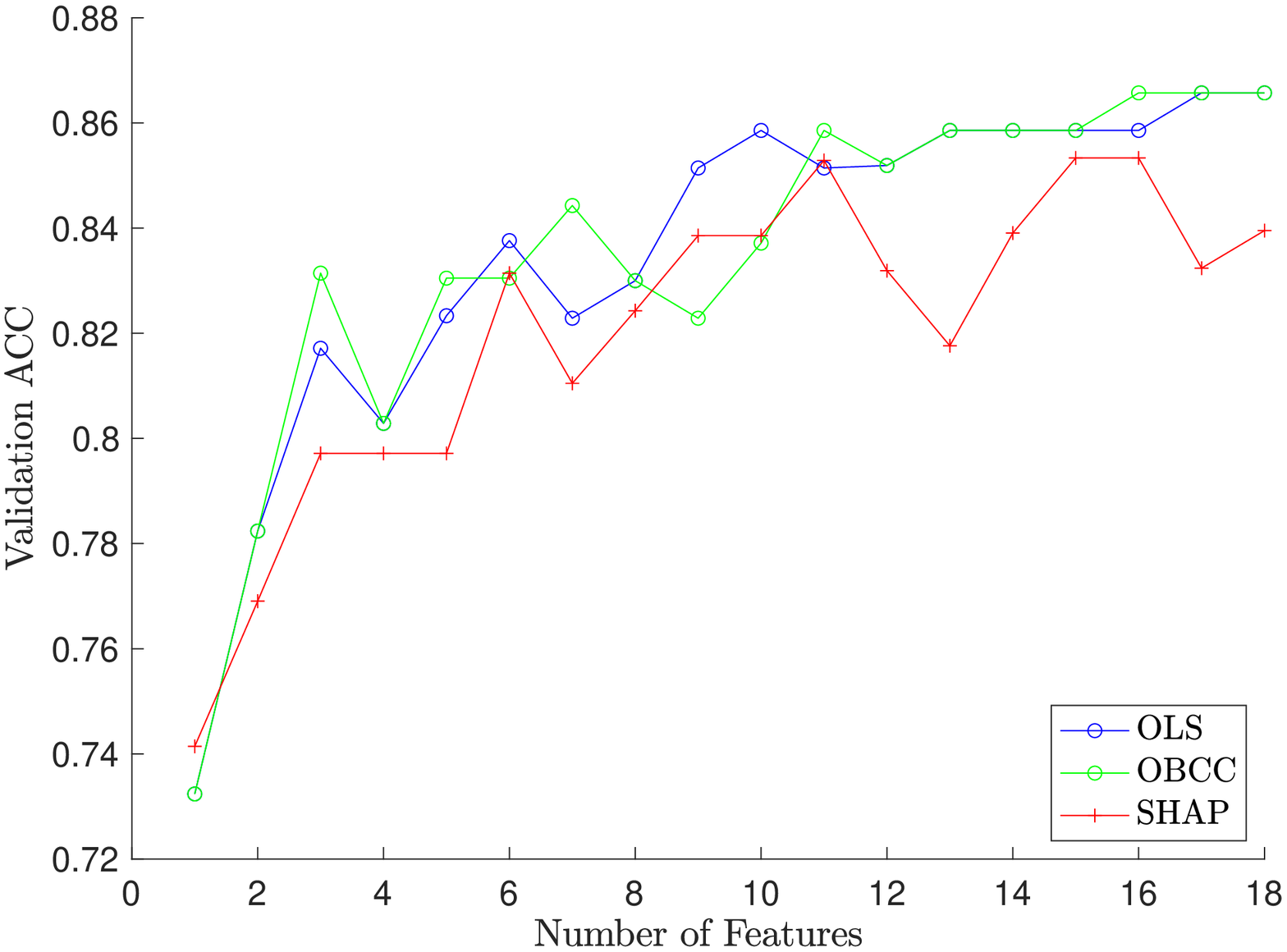}
\caption{}
\label{fig:lymp_shap_b}
\end{subfigure}
\caption{The average ACC results of the 10-fold cross validation for the OLS, OBCC and SHAP methods on (a) training and (b) validation Lymph dataset.}
\label{fig:lymp_shap}
\end{figure}

\begin{figure}[ht]
\centering
\begin{subfigure}[b]{0.4\textwidth}
\centering
\includegraphics[width=1\linewidth]{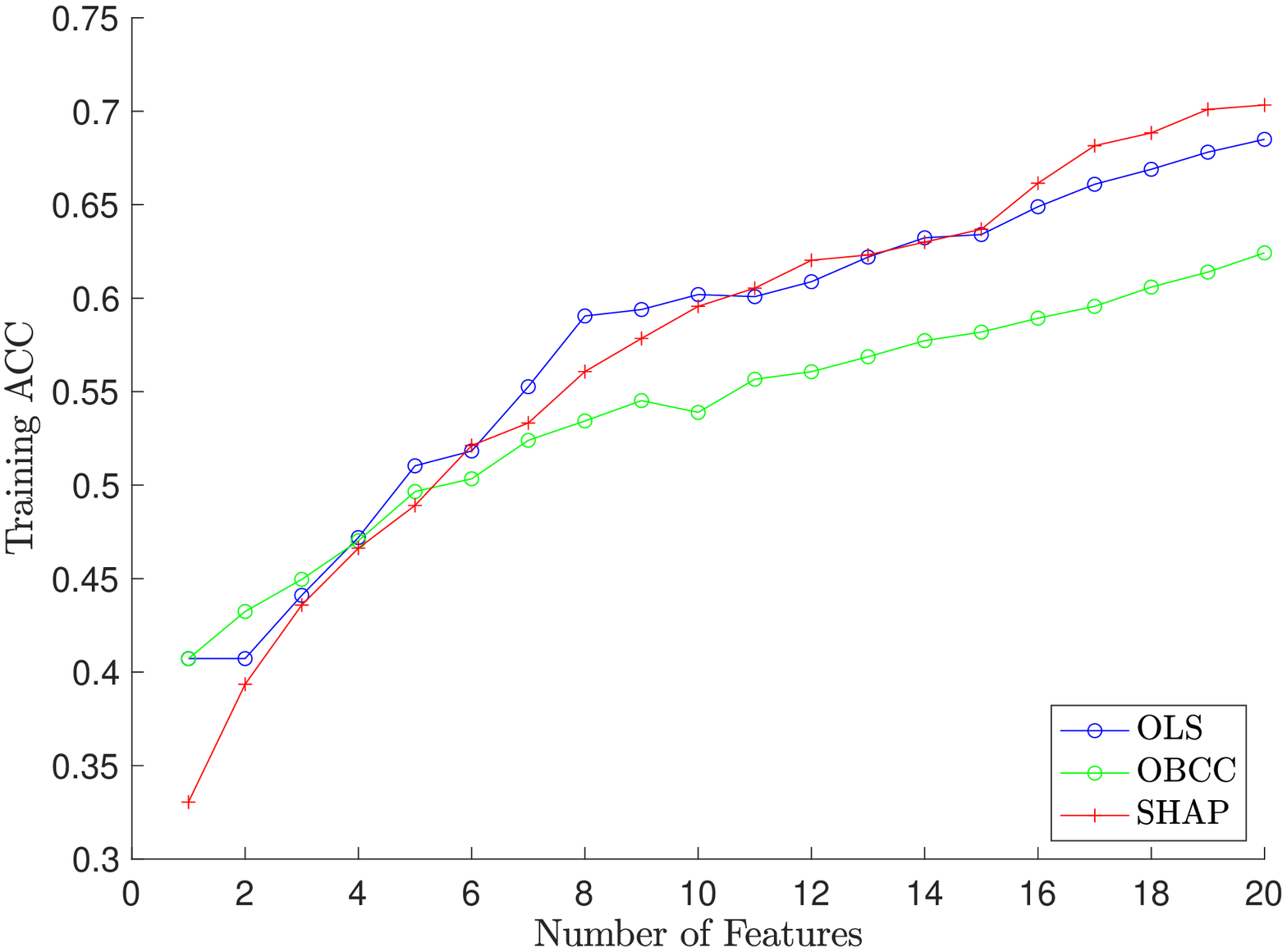}
\caption{}
\label{fig:flag_shap_a}
\end{subfigure}
\hspace{1cm}
\begin{subfigure}[b]{0.4\textwidth}
\centering
\includegraphics[width=1\linewidth]{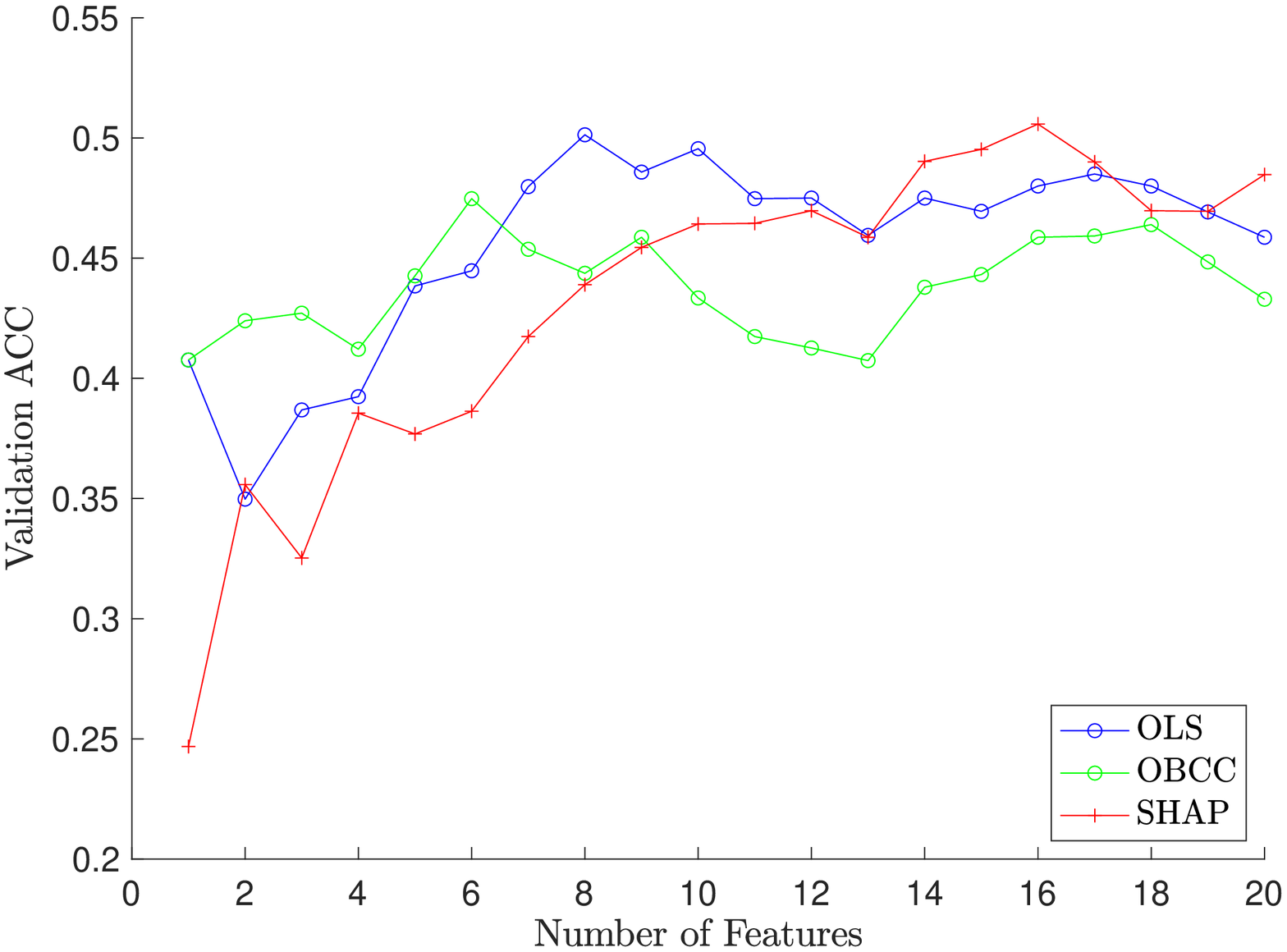}
\caption{}
\label{fig:flag_shap_b}
\end{subfigure}
\caption{The average ACC results of the 10-fold cross validation for the OLS, OBCC and SHAP methods on (a) training and (b) validation Flag dataset.}
\label{fig:flag_shap}
\end{figure}

\begin{figure}[ht]
\centering
\begin{subfigure}[b]{0.4\textwidth}
\centering
\includegraphics[width=1\linewidth]{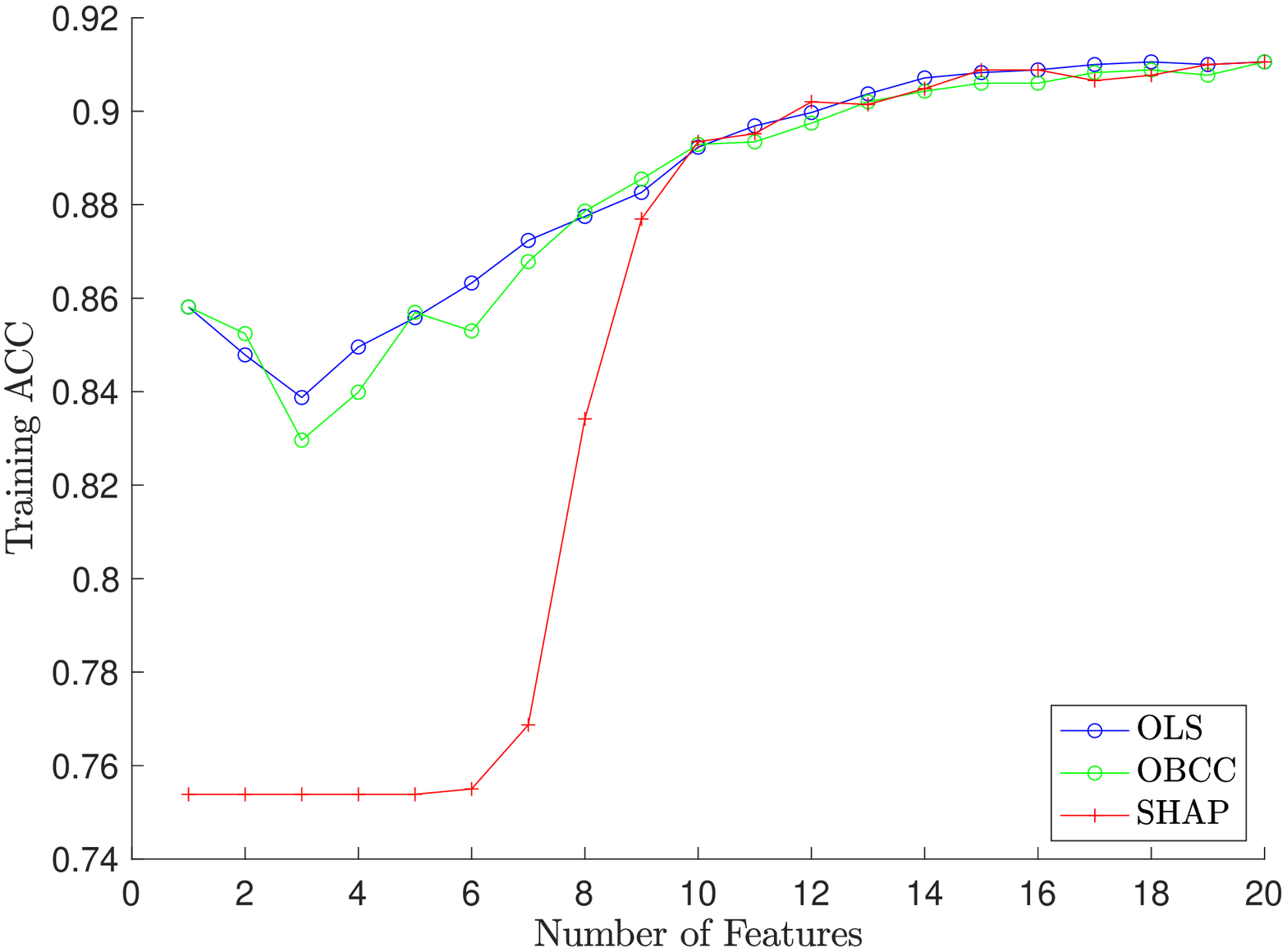}
\caption{}
\label{fig:park_shap_a}
\end{subfigure}
\hspace{1cm}
\begin{subfigure}[b]{0.4\textwidth}
\centering
\includegraphics[width=1\linewidth]{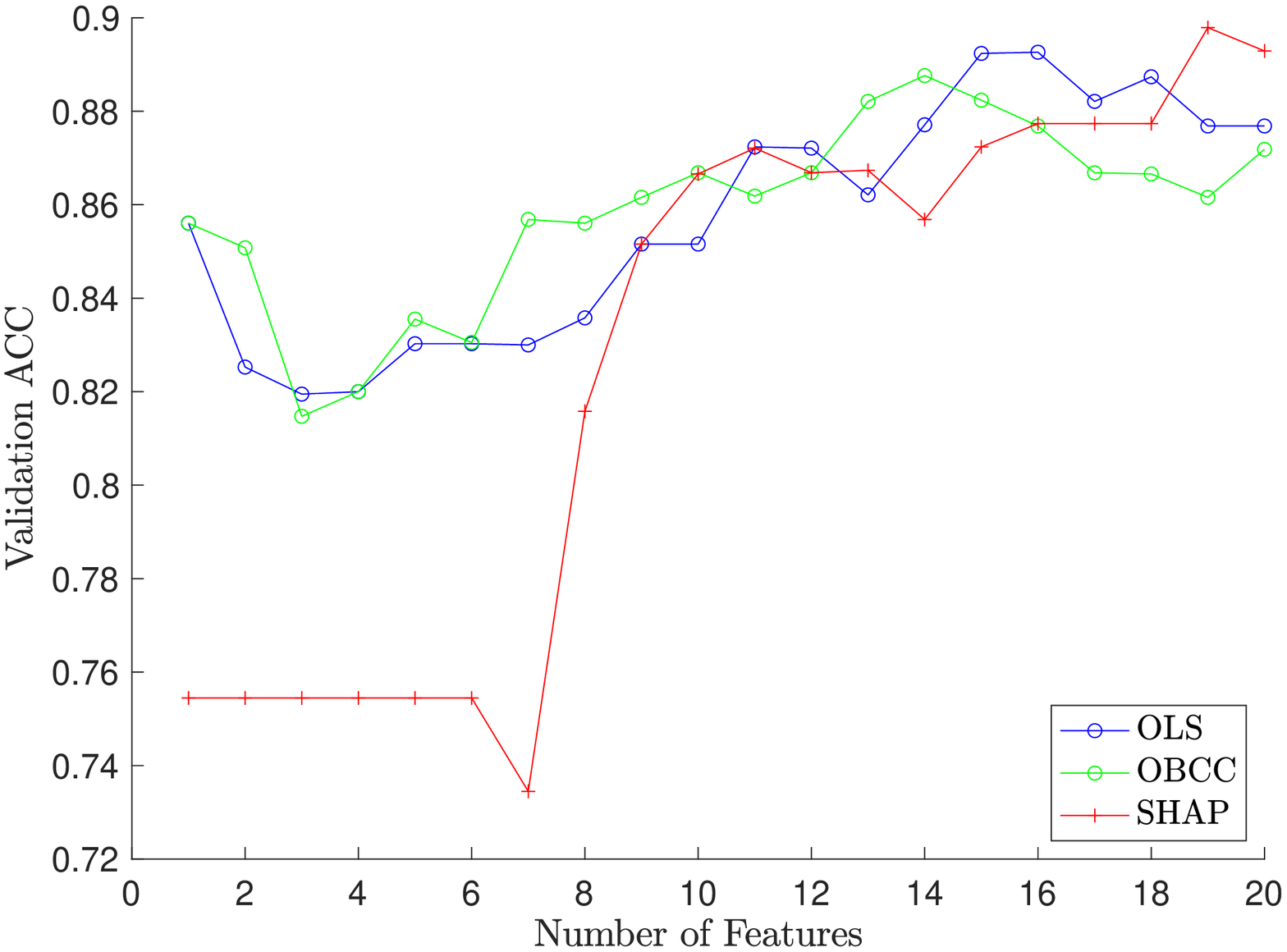}
\caption{}
\label{fig:park_shap_b}
\end{subfigure}
\caption{The average ACC results of the 10-fold cross validation for the OLS, OBCC and SHAP methods on (a) training and (b) validation Parkinsons dataset.}
\label{fig:park_shap}
\end{figure}

All three methods, i.e. OLS, OBCC and SHAP, can be used in both categorical and numerical features.
The 10-fold cross validation for the ACC of the LDA classifier is applied.
The feature selection and the LDA classifier training are implemented on the 10 training datasets.
The feature selection performance is evaluated by the average ACC on the training and validation datasets, which are shown in Fig. \ref{fig:breast_shap} to Fig. \ref{fig:park_shap}.

Theoretically, the SOCC and OBCC are identical when selecting the first feature.
Therefore, the two methods always select the same first feature and give the same ACC results in Fig. \ref{fig:breast_shap} to Fig. \ref{fig:park_shap}.
For the Flag dataset, as the OBCC adopts the point-biserial correlation coefficient, which can only be used in binomial classification issues, we extend it to the more general Pearson correlation coefficient for this multinomial classification case.
The performance of the OBCC method in the training datasets is significantly worse than other two methods, which is shown in Fig. \ref{fig:flag_shap_a}.
For the Parkinsons dataset, the first 6 features selected by the SHAP method is not helpful to the LDA classifiers.
The proposed OLS method achieves the best ACC results in the Breast dataset and comparable results in the rest datasets.
After the comprehensive comparison,  the proposed OLS method does not show the obvious deficiency in any datasets, which implies the great robustness of this method across the different datasets.

\section{Conclusions} \label{sec:conc}
This paper proposes a novel OLS based feature selection method for classification.
The method is based on the newly proposed concept of SOCCs which, for the first time, reveals an important relationship of the OLS based solution to a least-squares problem with the multiple correlation coefficient and the canonical correlation coefficient.
Utilising the relationships, the OLS based feature selection method is developed where either the multiple correlation coefficient (for binomial classification) or the canonical correlation coefficient (for multinomial classification) is used as the feature ranking criterion.
The relationship between CCA and LDA is analysed to demonstrate the statistical implication of the canonical correlation coefficient hence the proposed SOCCs in LDA based classification problem.
The speed advantage of the OLS based feature selection method in greedy search has been analysed.
In empirical studies, a simple example has been used to illustrate the procedure of the OLS based feature selection method, and to demonstrate the relationship of the SOCCs with canonical correlation coefficient and Fisher's criterion.
The synthetic and real world datasets have been used to compare the mutual information based methods and the embedded methods with the new OLS based method, showing that the OLS based method is always in the top 5 among 12 candidate methods.
In addition, as the mutual information estimation normally requires the discretisation on the continuous features, the OLS based methods, which can deal with both numerical (including continuous and discrete) and categorical features, is more convenient to use than the mutual information based methods in addressing continuous feature based classification problems.

\section*{Acknowledgements}
The authors would like to acknowledge that this work was supported by the UK Engineering and Physical Science Research Council Grant {EP/R018480/1}.

\bibliography{pr.bib}

\appendix
\section{The proof of the relationship between SOCCs and multiple correlation coefficient} \label{s:appendix1}
In the ordinary least-squares problem, the linear regression model with $N$ instances is given by
\begin{equation}\label{eq:lm_it}
    \mathbf{y} = \left({\mathbf{1}, \mathbf{X}}\right)\begin{pmatrix}
    \beta_0 \\
    \bm{\upbeta}
    \end{pmatrix} + \mathbf{e}\text{.}
\end{equation}
It can be shown that the least-squares estimation of $\beta _0$ and $\bm{\upbeta}$, denoted as $\hat\beta _0$ and $\bm{\hat\upbeta}$, satisfies the equation (see \ref{s:appendix} for proof)
\begin{equation}\label{eq:beta0}
\begin{split}
    \hat\beta _0 &= \bar{y} - \bar{\mathbf{x}}^\top\bm{\hat\upbeta}\\
    &= \bar{y} - \left({\bar{x}_1,\ldots,\bar{x}_n}\right)\bm{\hat\upbeta}\text{,}
\end{split}
\end{equation}
where $\bar{y}$ is the sample mean of $\mathbf{y}$, and $\bar{x}_i$ is the sample mean of $\mathbf{x}_i$.
Substituting \eqref{eq:beta0} into \eqref{eq:lm_it}, the linear model \eqref{eq:lm_it} is simplified to
\begin{equation} \label{eq:lm_c}
    \mathbf{y}_\text{C} = \mathbf{X}_\text{C}\bm{\hat\upbeta} + \hat{\mathbf{e}}\text{,}
\end{equation}
where $\hat{\mathbf{e}}$ is an estimation for the error term $\mathbf{e}$, $\mathbf{y}_\text{C}$ is the centred response variable given by 
\begin{equation}
\begin{split}
\mathbf{y}_\text{C} &= \begin{pmatrix}
y_1 - \bar{y} \\
\vdots\\
y_N - \bar{y} \\
\end{pmatrix}\text{,}
\end{split}
\end{equation}
and $\mathbf{X}_\text{C}$ is the matrix of the centred independent variables given by
\begin{equation} \label{eq:Xc}
\begin{split}
    \mathbf{X}_\text{C} &= \begin{pmatrix}
    \mathbf{x}_{\text{C}1},\ldots,\mathbf{x}_{\text{C}n}
    \end{pmatrix} \\
    &= \begin{pmatrix}
    x_{1,1} - \bar{x}_1 & \hdots & x_{1,n} - \bar{x}_n \\
    \vdots & \ddots & \vdots \\
    x_{N,1} - \bar{x}_1 & \hdots & x_{N,n} - \bar{x}_n
    \end{pmatrix}\text{.}
\end{split}
\end{equation}
Equation \eqref{eq:lm_c} implies that $\bm{\hat\upbeta}$ satisfies the normal equation (see \ref{s:appendix} for proof)
\begin{equation} \label{eq:beta1}
    \left( {\mathbf{X}_\text{C}^{\top}\mathbf{X}_\text{C}} \right)\bm{\hat\upbeta} = \mathbf{X}_\text{C}^{\top}\mathbf{y}_\text{C}\text{,} 
\end{equation}
transforming the least-squares problem with the intercept into the least-squares problem without the intercept.

When $\mathbf{X}_\text{C}$ has full column rank, the unnormalised reduced QR decomposition is performed on $\mathbf{X}_\text{C}$ as
\begin{equation} \label{eq:qrc}
    \mathbf{X}_\text{C} = \mathbf{W}_\text{C}\mathbf{A}\text{,}
\end{equation}
where $\mathbf{A}$ is a $n \times n$ invertible upper triangular matrix and $\mathbf{W}_\text{C}$ is a $N \times n$ matrix with the orthogonal columns ${\mathbf{w}_{\text{C}1}}, \ldots ,{\mathbf{w}_{\text{C}n}}$.
As $\mathbf{W}_{\text{C}} = \mathbf{X}_{\text{C}}\mathbf{A}^{-1}$, it can be seen that ${{\mathbf{w}}_{\text{C}i}}$, which is the linear transformation of $\mathbf{x}_{\text{C}1},\ldots,\mathbf{x}_{\text{C}n}$, has zero sample mean.
Substituting \eqref{eq:qrc} into \eqref{eq:lm_c} yields
\begin{equation} \label{eq:cols}
    {\mathbf{y}_\text{C}} = {\mathbf{W}_\text{C}\mathbf{\hat{g}}} + {\hat{\mathbf{e}}}\text{,}
\end{equation}
where $\mathbf{\hat{g}} = \mathbf{A}\bm{\hat\upbeta} = \left({\hat{g}_1,\ldots,\hat{g}_n}\right)^\top$.
The parameter vector $\mathbf{\hat{g}}$ obviously satisfies the normal equation
\begin{equation} \label{eq:ols_c}
    \mathbf{W}_\text{C}^{\top}\mathbf{W}_\text{C}\mathbf{\hat{g}} = \mathbf{W}_\text{C}^{\top}\mathbf{y}_\text{C}\text{.}
\end{equation}
Thus, the ordinary least-squares problem \eqref{eq:beta1} about $\mathbf{X}_\text{C}$ and $\mathbf{y}_\text{C}$ is transformed into the OLS problem \eqref{eq:ols_c} about $\mathbf{W}_\text{C}$ and $\mathbf{y}_\text{C}$.

The residual sum of squares for OLS is given by
\begin{equation}
\begin{split}
    \hat{\mathbf{e}}^\top\hat{\mathbf{e}} &= \left(\mathbf{y}_\text{C}-\mathbf{W}_\text{C}\mathbf{\hat{g}}\right)^\top\left(\mathbf{y}_\text{C}-\mathbf{W}_\text{C}\mathbf{\hat{g}}\right)\\
    &= \mathbf{y}_\text{C}^\top\mathbf{y}_\text{C} - 2\mathbf{\hat{g}}^{\top}\mathbf{W}_\text{C}^{\top}\mathbf{y}_\text{C} + \mathbf{\hat{g}}^{\top}\mathbf{W}_\text{C}^{\top}\mathbf{W}_\text{C}\mathbf{\hat{g}}\text{.}
\end{split}
\end{equation}
Because of \eqref{eq:ols_c}, this equation becomes
\begin{equation} \label{eq:sse2c}
    \hat{\mathbf{e}}^\top\hat{\mathbf{e}} = \mathbf{y}_\text{C}^\top\mathbf{y}_\text{C} - \mathbf{\hat{g}}^{\top}\mathbf{W}_\text{C}^{\top}\mathbf{W}_\text{C}\mathbf{\hat{g}}\text{.}
\end{equation}
As $\mathbf{W}_\text{C}$ is orthogonal, the inner product $\mathbf{W}_\text{C}^{\top}\mathbf{W}_\text{C}$ is the diagonal matrix $\diag{({\mathbf{w}_{\text{C}1}^{\top}\mathbf{w}_{\text{C}1},\ldots,\mathbf{w}_{\text{C}n}^{\top}\mathbf{w}_{\text{C}n}})}$.
Thus, \eqref{eq:sse2c} can be rewritten to
\begin{equation} \label{eq:olsc_square}
    {{\hat{\mathbf{e}}}^{\top}}{\hat{\mathbf{e}}} = {{\mathbf{y}_\text{C}}^{\top}}{\mathbf{y}_\text{C}} - \sum\limits_{i = 0}^{n} {\hat{g}_i^{2}\mathbf{w}_{\text{C}i}^{\top}{{\mathbf{w}}_{\text{C}i}}}\text{.}
\end{equation}
Both sides of \eqref{eq:olsc_square} are divided by ${{\mathbf{y}_\text{C}}^{\top}}{\mathbf{y}_\text{C}}$, that is
\begin{equation}\label{eq:cerr}
\frac{{{{\hat{\mathbf{e}}}^{\top}}{\hat{\mathbf{e}}}}}{{{{\mathbf{y}_\text{C}}^{\top}}{\mathbf{y}_\text{C}}}} = 1 - \sum_{i = 1}^n\frac{{\hat{g}_i^{2}{\mathbf{w}_{\text{C}i}^{\top}{\mathbf{w}_{\text{C}i}}} }}{{{{\mathbf{y}_\text{C}}^{\top}}{\mathbf{y}_\text{C}}}}\text{.}
\end{equation}
Due to the orthogonality of $\mathbf{W}_\text{C}$, the computation of the parameter vector $\mathbf{\hat{g}}$ can be simplified as
\begin{equation} \label{eq:g}
    \hat{g}_i = \frac{\mathbf{w}^{\top}_{\text{C}i}\mathbf{y}_\text{C}}{\mathbf{w}^{\top}_{\text{C}i}\mathbf{w}_{\text{C}i}}\text{.}
\end{equation}
Substituting \eqref{eq:g} into \eqref{eq:cerr},
\begin{equation} \label{eq:socc}
\begin{split}
    \frac{{{{\hat{\mathbf{e}}}^{\top}}{\hat{\mathbf{e}}}}}{{{{\mathbf{y}_\text{C}}^{\top}}{\mathbf{y}_\text{C}}}} &= 1 - \sum_{i = 1}^n\frac{\mathbf{y}_\text{C}^\top{\mathbf{w}_{\text{C}i}{\mathbf{w}^{\top}_{\text{C}i}}}\mathbf{y}_\text{C}}{{\mathbf{w}_{\text{C}i}^{\top}{\mathbf{w}_{\text{C}i}}}\mathbf{y}_\text{C}^\top\mathbf{y}_\text{C}} \\
     &= 1- \sum_{i = 1}^n h_i\text{,}
\end{split}
\end{equation}
where $h_i$ is the SOCC.

The definition of the multiple correlation coefficient is based on the centred linear regression model \eqref{eq:lm_c}.
The squared multiple correlation coefficient (or called coefficient of determination) has the following relationship with the total sum of squares $SST$ and the residual sum of squares $SSR$ of the model \eqref{eq:lm_c}
\begin{equation} \label{eq:r2}
    R^2({\mathbf{X},\mathbf{y}}) = 1 - \frac{SSR({\mathbf{X}_\text{C},\mathbf{y}_\text{C}})}{SST({\mathbf{X}_\text{C},\mathbf{y}_\text{C}})} \text{,}
\end{equation}
where
\begin{equation}
\begin{split}
    SST({\mathbf{X}_\text{C},\mathbf{y}_\text{C}}) &= \mathbf{y}_\text{C}^\top\mathbf{y}_\text{C}\\
    SSR({\mathbf{X}_\text{C},\mathbf{y}_\text{C}}) &= \left({\mathbf{y}_\text{C} - \hat{\mathbf{y}}_\text{C}}\right)^\top \left({\mathbf{y}_\text{C} - \hat{\mathbf{y}}_\text{C}}\right) = \hat{\mathbf{e}}^\top\hat{\mathbf{e}}\text{.}
\end{split}
\end{equation}
Comparing \eqref{eq:socc} and \eqref{eq:r2}, it is found
\begin{equation}
    R^2({\mathbf{X},\mathbf{y}}) = \sum_{i = 1}^n h_i\text{.}
\end{equation}

\section{The proof of the relationship between SOCCs and canonical correlation coefficients} \label{s:appendix2}
According to \eqref{eq:ccca}, the sum of the squared canonical correlation coefficients, i.e. the sum of the eigenvalues, are given by
\begin{equation} \label{eq:tr}
    \sum_{k = 1}^{n \wedge m}R_k^2({\mathbf{X},\mathbf{Y}}) = \tr{\left(\mathbf{R}_{\mathbf{X},\mathbf{X}}^{-1} \mathbf{R}_{\mathbf{X},\mathbf{Y}} \mathbf{R}_{\mathbf{Y},\mathbf{Y}}^{-1} \mathbf{R}_{\mathbf{Y},\mathbf{X}}\right)}\text{,}
\end{equation}
where the operator $\tr$ denotes the trace of the matrix.
If the columns of $\mathbf{X}$ are zero sample mean and orthogonal, the correlation matrix of $\mathbf{X}$ is identity matrix, so $\mathbf{R}_{\mathbf{X},\mathbf{X}}^{-1} = \mathbf{I}$.
It is known that the multiple correlation between $\mathbf{Y}$ and each $\mathbf{x}_i$ can be evaluated by \cite[p.~174]{cooley1971multivariate}
\begin{equation} \label{eq:diag}
    \begin{pmatrix}
    R^2({\mathbf{x}_1,\mathbf{Y}})\\
    \vdots\\
    R^2({\mathbf{x}_n,\mathbf{Y}})
    \end{pmatrix} = \diag{\left(\mathbf{R}_{\mathbf{X},\mathbf{Y}}\mathbf{R}_{\mathbf{Y},\mathbf{Y}}^{-1}\mathbf{R}_{\mathbf{Y},\mathbf{X}}\right)}\text{,}
\end{equation}
where the operator $\diag$ obtains the main diagonal of the matrix.
Therefore, according to \eqref{eq:tr} and \eqref{eq:diag}, the following equation holds when the columns of $\mathbf{X}$ are centred and orthogonal.
\begin{equation} \label{eq:mcc2ccc}
    \sum_{k = 1}^{n \wedge m}R_k^2({\mathbf{X},\mathbf{Y}}) = \sum_{i = 1}^{n}R^2({\mathbf{x}_i,\mathbf{Y}})\text{.}
\end{equation}

Through the unnormalised reduced QR decomposition,
\begin{equation}
\begin{split}
    \mathbf{X}_\text{C} &= \mathbf{W}_{\text{C}}\mathbf{A} \\
    \mathbf{Y}_\text{C} &= \mathbf{V}_{\text{C}}\mathbf{B}
\end{split}
\end{equation}
where $\mathbf{W}_{\text{C}}$ is a $N \times n$ matrix with the centred orthogonal columns given by
\begin{equation}
    \mathbf{W}_{\text{C}} = (\mathbf{w}_{\text{C}1},\ldots,\mathbf{w}_{\text{C}n})\text{,}
\end{equation}
$\mathbf{V}_{\text{C}}$ is a $N \times m$ matrix with the centred orthogonal columns given by
\begin{equation}
    \mathbf{V}_{\text{C}} = (\mathbf{v}_{\text{C}1},\ldots,\mathbf{v}_{\text{C}m})\text{,}
\end{equation}
$\mathbf{A}$ is a $n \times n$ invertible upper triangular matrix, and $\mathbf{B}$ is a $m \times m$ invertible upper triangular matrix.
It is noticed that the transformation from $\mathbf{X}$ (or $\mathbf{Y}$) to $\mathbf{W}_{\text{C}}$ (or $\mathbf{V}_{\text{C}}$) is affine.
As the canonical correlation coefficient is invariant under affine transformations,
\begin{equation} \label{eq:affine}
    R_k({\mathbf{X},\mathbf{Y}}) = R_k({\mathbf{W}_{\text{C}},\mathbf{V}_{\text{C}}})\quad\quad k = 1,\ldots,n \wedge m\text{.}
\end{equation}
As the columns of $\mathbf{W}_\text{C}$ are centred and orthogonal, the following equation holds according to \eqref{eq:mcc2ccc} and \eqref{eq:affine}.
\begin{equation}\label{eq:sccc}
    \sum_{k = 1}^{n \wedge m}R_k^2({\mathbf{X},\mathbf{Y}}) = \sum_{k = 1}^{n \wedge m}R_k^2({\mathbf{W}_{\text{C}},\mathbf{V}_{\text{C}}}) = \sum_{i = 1}^{n} R^2({\mathbf{w}_{\text{C}i},\mathbf{V}_{\text{C}}})\text{.}
\end{equation}

Define squared orthogonal correlation matrix as
\begin{equation}
    \mathbf{H} = \begin{pmatrix}
    h_{1,1} & \dots & h_{1,m} \\
    \vdots & \ddots & \vdots \\
    h_{n,1} & \dots & h_{n,m}
    \end{pmatrix}\text{,}
\end{equation}
where $h_{i,j}$ is the SOCC given by
\begin{equation}
    h_{i,j} = \frac{\mathbf{v}_{\text{C}j}^\top{\mathbf{w}_{\text{C}i}{\mathbf{w}_{\text{C}i}^\top}}\mathbf{v}_{\text{C}j}}
    {{\mathbf{w}_{\text{C}i}^{\top}{\mathbf{w}_{\text{C}i}}}\mathbf{v}_{\text{C}j}^\top\mathbf{v}_{\text{C}j}}\text{.}
\end{equation}
Due to \eqref{eq:scerr}, the multiple correlation coefficient between $\mathbf{V}_\text{C}$ and each $\mathbf{w}_{\text{C}}$ can be evaluated by
\begin{equation} \label{eq:smcc}
\begin{split}
    R^2({\mathbf{w}_{\text{C}1},\mathbf{V}_\text{C}}) &= \sum_{j = 1}^{m} h_{1,j} \\
        &\vdots \\
        R^2({\mathbf{w}_{\text{C}n},\mathbf{V}_\text{C}}) &= \sum_{j = 1}^{m} h_{n,j}\text{.}
\end{split}
\end{equation}
Substituting \eqref{eq:smcc} into \eqref{eq:sccc}, it is found that the sum of the squared canonical correlation coefficients between $\mathbf{X}$ and $\mathbf{Y}$ is equal to the sum of all entries of the squared orthogonal correlation matrix $\mathbf{H}$, that is
\begin{equation} 
    \sum_{k = 1}^{n \wedge m}R_k^2({\mathbf{X},\mathbf{Y}}) = \sum_{i = 1}^{n}\sum_{j = 1}^{m}h_{i,j}\text{.}
\end{equation}

\section{Preliminary knowledge of linear regression} \label{s:appendix}
The ordinary least-squares estimation of a linear model with intercept is to find the optimal parameters $\hat\beta_0 \in \mathbb{R}$ and $\bm{\hat\upbeta} \in \mathbb{R}^n$ to minimise the squared residual given by
\begin{equation}
    L = (\mathbf{y} - \hat\beta_0\mathbf{1} - \mathbf{X}\bm{\hat\upbeta})^\top(\mathbf{y} - \hat\beta_0\mathbf{1} - \mathbf{X}\bm{\hat\upbeta})\text{,}
\end{equation}
where $\mathbf{y} \in \mathbb{R}^N$ is the response vector, $\mathbf{X} \in \mathbb{R}^{N \times n}$ is the design matrix, and $\mathbf{1}$ is the $N \times 1$ vector of ones.
The squared residual is minimised when the derivative of $L$ with respect to the parameters is zero, that is
\begin{equation} \label{eq:appendix_pd}
\begin{split}
    \frac{\partial L}{\partial \bm{\hat\upbeta}_{\mathbf{1}}} &= \frac{\partial(\mathbf{y}^\top\mathbf{y} - \mathbf{y}^\top\mathbf{X}_\mathbf{1}\bm{\hat\upbeta}_{\mathbf{1}} - \bm{\hat\upbeta}_{\mathbf{1}}^\top\mathbf{X}_\mathbf{1}^\top\mathbf{y} + \bm{\hat\upbeta}_{\mathbf{1}}^\top\mathbf{X}_\mathbf{1}^\top\mathbf{X}_\mathbf{1}\bm{\hat\upbeta}_{\mathbf{1}})}{\partial \bm{\hat\upbeta}_{\mathbf{1}}} \\
    &= -2\mathbf{X}_\mathbf{1}^\top\mathbf{y} + 2\mathbf{X}_\mathbf{1}^\top\mathbf{X}_\mathbf{1}\bm{\hat\upbeta}_{\mathbf{1}} \\
    &= 0\text{,}
\end{split}
\end{equation}
where $\bm{\hat\upbeta}_\mathbf{1} = (\hat\beta_0, \bm{\hat\upbeta})^\top$ and $\mathbf{X}_\mathbf{1} = ({\mathbf{1}, \mathbf{X}})$.
According to \eqref{eq:appendix_pd}, it is known the optimal parameters satisfy the equation given by
\begin{equation} \label{eq:appendix_norm}
\begin{split}
    \mathbf{X}_\mathbf{1}^\top\mathbf{y} &= \mathbf{X}_\mathbf{1}^\top\mathbf{X}_\mathbf{1}\bm{\hat\upbeta}_\mathbf{1} \\
    \begin{pmatrix}
    \mathbf{1}^\top \\
    \mathbf{X}^\top
    \end{pmatrix}\mathbf{y} &= \begin{pmatrix}
    \mathbf{1}^\top \\
    \mathbf{X}^\top
    \end{pmatrix}({\mathbf{1}, \mathbf{X}})\begin{pmatrix}
    \hat\beta_0 \\
    \bm{\hat\upbeta}
    \end{pmatrix} \\
    \begin{pmatrix}
    N\bar{y} \\
    \mathbf{X}^\top\mathbf{y}
    \end{pmatrix} &= \begin{pmatrix}
    N & N\bar{\mathbf{x}}^\top \\
    N\bar{\mathbf{x}} & \mathbf{X}^\top\mathbf{X}
    \end{pmatrix}\begin{pmatrix}
    \hat\beta_0 \\
    \bm{\hat\upbeta}
    \end{pmatrix}\text{,}
\end{split}
\end{equation}
where $\bar{y} \in \mathbb{R}$ is the sample mean of $\mathbf{y}$ and $\bar{\mathbf{x}} \in \mathbb{R}^n$ is the column vector composed of the sample mean of each column of $\mathbf{X}$.
Then, the two important equations can be found in \eqref{eq:appendix_norm}.
The first one is 
\begin{equation} \label{eq:appendix_intercept}
\begin{split}
    N\bar{y} &= N\hat\beta_0 + N\bar{\mathbf{x}}^\top\bm{\hat\upbeta} \\
    \bar{y} &= \hat\beta_0 + \bar{\mathbf{x}}^\top\bm{\hat\upbeta} \text{.}
\end{split}
\end{equation}
The second one is
\begin{equation} \label{eq:appendix_centre}
\begin{split}
    \mathbf{X}^\top\mathbf{y} &= N\bar{\mathbf{x}}\hat\beta_0 + \mathbf{X}^\top\mathbf{X}\bm{\hat\upbeta} \\
    (\mathbf{X}_\text{C} + \mathbf{1}\bar{\mathbf{x}}^\top)^\top(\mathbf{y}_\text{C} + \bar{y}\mathbf{1}) &= N\bar{\mathbf{x}}\hat\beta_0 + (\mathbf{X}_\text{C} + \mathbf{1}\bar{\mathbf{x}}^\top)^\top(\mathbf{X}_\text{C} + \mathbf{1}\bar{\mathbf{x}}^\top)\bm{\hat\upbeta} \\
    \mathbf{X}_\text{C}^\top\mathbf{y}_\text{C} + \bar{y}\mathbf{X}_\text{C}^\top\mathbf{1} +  \bar{\mathbf{x}}\mathbf{1}^\top\mathbf{y}_\text{C} + \bar{y}\bar{\mathbf{x}}\mathbf{1}^\top\mathbf{1} &= N\bar{\mathbf{x}}\hat\beta_0 + (\mathbf{X}_\text{C}^\top\mathbf{X}_\text{C} + \mathbf{X}_\text{C}^\top\mathbf{1}\bar{\mathbf{x}}^\top + \bar{\mathbf{x}}\mathbf{1}^\top\mathbf{X}_\text{C} + \bar{\mathbf{x}}\mathbf{1}^\top\mathbf{1}\bar{\mathbf{x}}^\top)\bm{\hat\upbeta} \\
    \mathbf{X}_\text{C}^\top\mathbf{y}_\text{C} + \bar{y}\mathbf{X}_\text{C}^\top\mathbf{1} +  \bar{\mathbf{x}}\mathbf{1}^\top\mathbf{y}_\text{C} + N\bar{y}\bar{\mathbf{x}} &= N\bar{\mathbf{x}}\hat\beta_0 + (\mathbf{X}_\text{C}^\top\mathbf{X}_\text{C} + \mathbf{X}_\text{C}^\top\mathbf{1}\bar{\mathbf{x}}^\top + \bar{\mathbf{x}}\mathbf{1}^\top\mathbf{X}_\text{C} + N\bar{\mathbf{x}}\bar{\mathbf{x}}^\top)\bm{\hat\upbeta} \text{,}
\end{split}
\end{equation}
where $\mathbf{X}_\text{C}$ is the column centred matrix of $\mathbf{X}$ by its sample mean $\bar{\mathbf{x}}$ and $\mathbf{y}_\text{C}$ is the centred vector of $\mathbf{y}$ by its sample mean $\bar{y}$.
As $\mathbf{X}_\text{C}$ and $\mathbf{y}_\text{C}$ have zero sample means, it is known that $\mathbf{X}_\text{C}^\top\mathbf{1}$ and $\mathbf{1}^\top\mathbf{y}_\text{C}$ are zeros.
Due to this and \eqref{eq:appendix_intercept}, the equation \eqref{eq:appendix_centre} can be rewritten as
\begin{equation}
\begin{split}
    \mathbf{X}_\text{C}^\top\mathbf{y}_\text{C} + N\bar{y}\bar{\mathbf{x}} &= N\bar{\mathbf{x}}\hat\beta_0 + (\mathbf{X}_\text{C}^\top\mathbf{X}_\text{C} + N\bar{\mathbf{x}}\bar{\mathbf{x}}^\top)\bm{\hat\upbeta} \\
    \mathbf{X}_\text{C}^\top\mathbf{y}_\text{C} + N\bar{y}\bar{\mathbf{x}} &= N\bar{\mathbf{x}}(\bar{y}-\bar{\mathbf{x}}^\top\bm{\hat\upbeta}) + (\mathbf{X}_\text{C}^\top\mathbf{X}_\text{C} + N\bar{\mathbf{x}}\bar{\mathbf{x}}^\top)\bm{\hat\upbeta} \\
    \mathbf{X}_\text{C}^\top\mathbf{y}_\text{C} &= \mathbf{X}_\text{C}^\top\mathbf{X}_\text{C}\bm{\hat\upbeta} 
\end{split}
\end{equation}

\end{document}